\documentclass[11pt]{article}

\usepackage[final]{acl}

\usepackage{times}
\usepackage{latexsym}

\usepackage[T1]{fontenc}
\usepackage[utf8]{inputenc}

\usepackage{microtype}

\usepackage{inconsolata}

\usepackage{graphicx}

\usepackage{subcaption}
\usepackage{booktabs}
\usepackage{hyperref}
\usepackage{amsmath}
\usepackage{amssymb}
\usepackage{mathtools}
\usepackage{amsthm}
\usepackage[capitalize,noabbrev]{cleveref}
\crefname{appendix}{Appendix}{Appendices}
\Crefname{appendix}{Appendix}{Appendices}
\usepackage{listings}
\usepackage{rotating}
\usepackage{placeins}
\usepackage{needspace}
\title{RetroReasoner: A Reasoning LLM for Strategic Retrosynthesis Prediction}

\author{
 \textbf{Hanbum Ko\textsuperscript{1}},
 \textbf{Chanhui Lee\textsuperscript{1}},
 \textbf{Ye Rin Kim\textsuperscript{1}},
 \textbf{Rodrigo Hormazabal\textsuperscript{3}},
\\
 \textbf{Sehui Han\textsuperscript{3}},
 \textbf{Sungbin Lim\textsuperscript{2}},
 \textbf{Sungwoong Kim\textsuperscript{1}},
\\
\\
 \textsuperscript{1}Department of Artificial Intelligence, Korea University,\\
 \textsuperscript{2}Department of Statistics, Korea University,\\
 \textsuperscript{3}Materials Intelligence Lab, LG AI Research,
\\
 \small{
   \textbf{Correspondence:} Sungbin Lim \href{mailto:sungbin@korea.ac.kr}{sungbin@korea.ac.kr},
   Sungwoong Kim \href{mailto:swkim01@korea.ac.kr}{swkim01@korea.ac.kr}
 }
}

\begin{document}
\maketitle

\begin{abstract}
Retrosynthesis prediction aims to identify reactants that can synthesize a given product molecule.
Although molecular large language models (LLMs) have recently shown promising results, most existing methods either generate reactants directly or provide only generic product-level analysis, without explicitly reasoning about bond-disconnection strategies that justify specific reactant choices.
This paper proposes RetroReasoner, a retrosynthetic reasoning model that captures chemists' strategic disconnection-based thinking.
RetroReasoner is trained with supervised fine-tuning and reinforcement learning.
For supervised fine-tuning, SyntheticRetro generates structured disconnection rationales paired with reactant predictions.
For reinforcement learning, a round-trip reward evaluates predicted reactants by passing them through a forward synthesis model and rewarding predictions that reconstruct the original product.
RetroReasoner can also be applied to multi-step retrosynthetic planning by incorporating it into a parallelized Monte Carlo tree search framework, reducing search time while increasing the number and diversity of valid synthetic pathways.
Experimental results show that RetroReasoner outperforms prior baselines, including not only molecular LLMs but also retrosynthesis-specific expert models, and generates a broader range of feasible reactant proposals, especially for challenging reaction instances.
The code is available at \url{https://github.com/KU-AGI/RetroReasoner}.
\end{abstract}
\section{Introduction}
\label{sec:intro}

\begin{figure*}[!t]
  % \vskip 0.2in
  \begin{center}
    % \centerline{
    \includegraphics[width=0.57\linewidth]{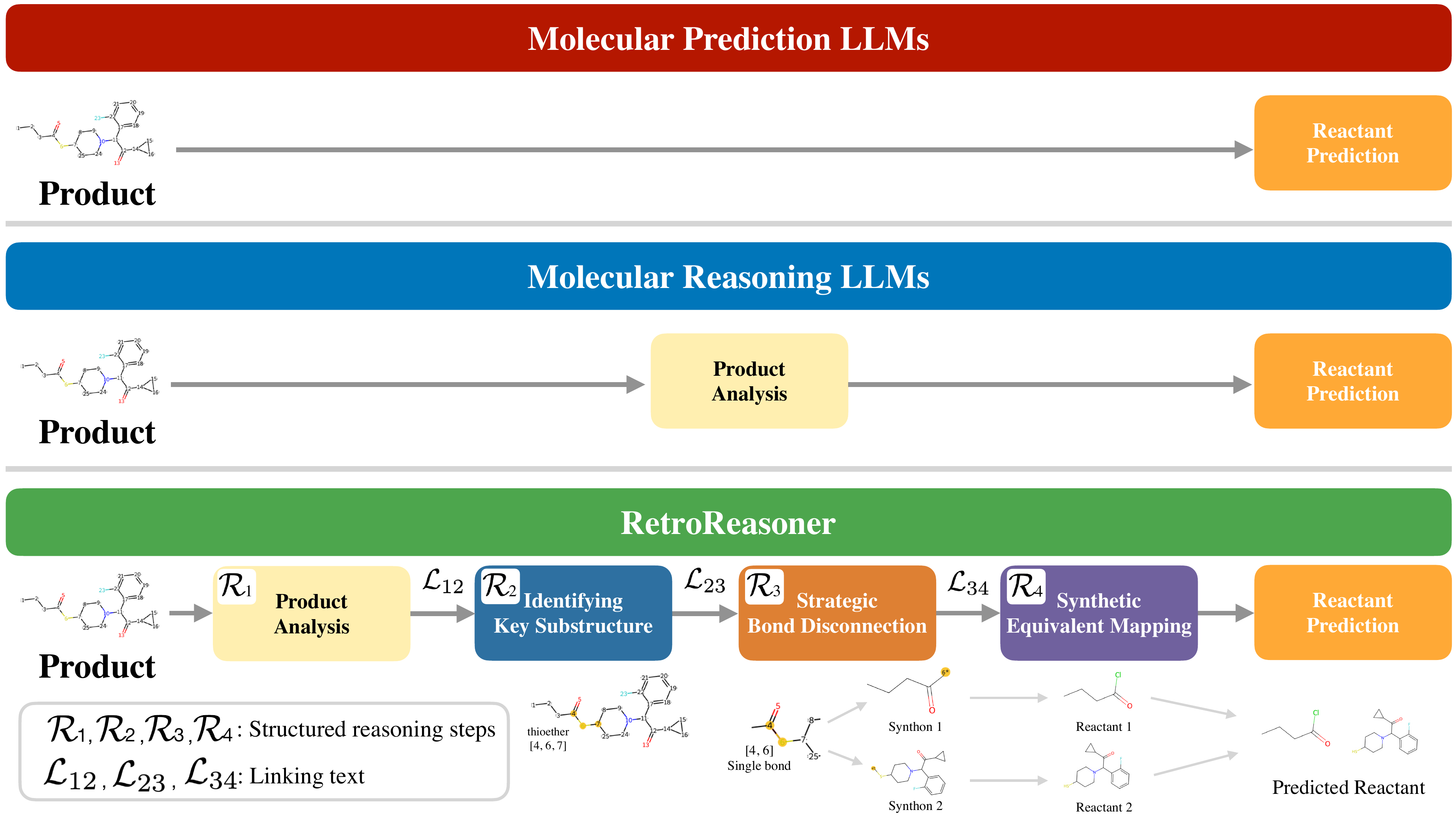}
    \includegraphics[width=0.41\linewidth]{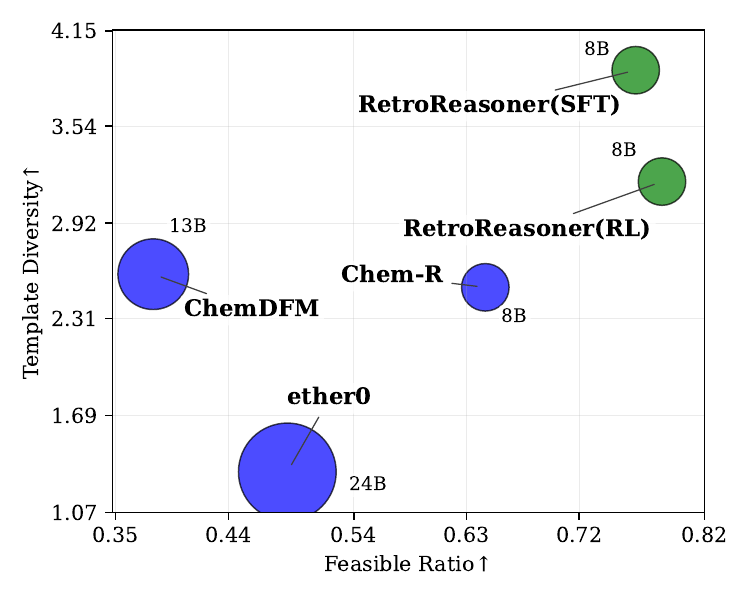}
    \caption{
        (Left) Comparison of reasoning processes among Molecular Reasoning LLMs, Molecular Prediction LLMs, and RetroReasoner(Ours). RetroReasoner suggest valid reactant given product by explicit strategic disconnection steps. (Right) Comparison of molecular reasoning LLMs. The x-axis represents the feasible ratio of reactant proposals, and the y-axis represents the diversity of proposals. Model sizes are distinguished by the size of the circle.
    }
    \label{fig:RetroReasoner}
  \end{center}
\end{figure*}

Retrosynthesis prediction is an important task in organic synthesis that aims to identify reactants for a desired target molecule.
A representative chemist strategy is to infer likely bond formations, disconnect the corresponding bonds into hypothetical reactant fragments called synthons~\cite{corey1967general,corey1988robert,corey1995logic}, and then assign standalone synthetic equivalents to each synthon.
Although effective, this process requires expert knowledge and substantial validation time.

Recent advances in large language models (LLMs)~\cite{yu2024llasmol,fang2023mol,pei2024biot5+,lee2025mol} have led to molecular LLMs that process molecular representations such as SMILES~\cite{weininger1988} and SELFIES~\cite{krenn2020self}, enabling their application to reactant prediction.
However, most molecular LLMs either predict reactants without explicit reasoning or provide reasoning that remains a generic product-level analysis, such as functional group enumeration and reaction mechanism explanation~\cite{wang2025chem,zhao2024chemdfm,zhao2025chemdfm-r,narayanan2025training,zhang2025reasoning,li2025retro}.
Such reasoning omits intermediate disconnection steps that are essential for retrosynthesis, causing a logical gap between product analysis and the selection of specific reactants.
An illustrative example is shown in~\Cref{fig:RetroReasoner}.

To address this limitation, RetroReasoner reasons through four stages: product analysis, key-substructure identification, strategic bond disconnection, and synthetic-equivalent mapping.
It is trained by supervised fine-tuning (SFT) on rationales generated by SyntheticRetro, followed by Group Relative Policy Optimization (GRPO)~\cite{shao2024deepseekmath} with a round-trip reward~\cite{schwaller2020predicting}.
SyntheticRetro uses reaction-derived supporting information to generate accurate intermediate reasoning between the structured stages.
By checking whether predicted reactants regenerate the product through a forward synthesis model, the round-trip reward supports accurate, round-trip-consistent alternatives beyond the single reference.

Experimental results show that RetroReasoner outperforms models without strategic reasoning, improving reactant prediction accuracy and round-trip consistency.
It also achieves strong performance on challenging evaluation datasets, including those focused on rare reaction types and those dominated by rare atoms or rare n-gram tokens.
Furthermore, RetroReasoner is applied to multi-step retrosynthetic planning through a parallelized Monte Carlo tree search (MCTS) framework.
This addresses a practical bottleneck of LLM-based single-step models, whose slow proposal speed can restrict repeated tree expansion.
Through this, RetroReasoner is used efficiently in multi-step planning in terms of runtime and diversity of synthetic paths.

In summary, the main contributions are as follows:

\begin{itemize}
    \item RetroReasoner, chemist-inspired stepwise reasoning model leveraging strategic bond disconnection, trained with SFT and further optimized with RL using a round-trip reward, improving prediction accuracy and round-trip consistency while achieving a favorable feasibility-diversity trade-off.
    \item SyntheticRetro, a data generation framework that produces the stepwise reasoning text by using various supporting information to generate accurate intermediate reasoning.
    \item Empirical validation across challenging single-step retrosynthesis benchmarks and multi-step planning experiments, where simple parallelized Monte Carlo tree search reduces solving time while increasing synthesis routes diversity.
\end{itemize}

% \begin{itemize}
%     \item A chemist-inspired stepwise reasoning process for retrosynthesis prediction and SyntheticRetro, a data generation framework that produces structured disconnection rationales for supervised fine-tuning.
%     \item RetroReasoner, a retrosynthetic reasoning model trained with SFT and optimized with RL using a round-trip reward, leading to more accurate, diverse, and feasible reactant proposals.
%     \item Empirical validation across challenging single-step retrosynthesis benchmarks and multi-step planning experiments, where LLM inference optimization combined with simple parallelized Monte Carlo tree search reduces solving time while increasing route diversity.
% \end{itemize}

A detailed discussion of related work is provided in~\Cref{sec:related_works}.

% However, many molecular LLMs either predict reactants without an explicit reasoning process or provide reasoning that remains a generic analysis of the product~\cite{wang2025chem,zhao2024chemdfm,zhao2025chemdfm-r,narayanan2025training,zhang2025reasoning,li2025retro}.
% As a result, step by step reasoning for reactant selection is missing, and logical jumps occur.
% An illustrative example of this difference is shown in~\Cref{fig:RetroReasoner}.
% \input{contents/02_related}
% \input{contents/03_preliminary}
\newpage
\section{RetroReasoner: A Reasoning LLM for Strategic Retrosynthesis Prediction}
\label{sec:RetroReasoner}
\begingroup
\setlength{\abovedisplayskip}{0pt plus 1pt}
\setlength{\belowdisplayskip}{0pt plus 1pt}
\setlength{\abovedisplayshortskip}{0pt plus 1pt}
\setlength{\belowdisplayshortskip}{0pt plus 1pt}

Retrosynthesis prediction aims to infer reactants that can synthesize a given target product, where molecules and reactions are represented as SMILES~\cite{weininger1988} or reaction (RXN) SMILES~\cite{rxnsmiles}.
Chemists approach this task through backward reasoning: identifying plausible bond disconnections, deriving hypothetical fragments called synthons, and assigning practically accessible synthetic equivalents.
Since multiple disconnections and synthon-to-equivalent mappings can be valid, a single product may correspond to multiple feasible reactant sets.
Multi-step retrosynthetic planning extends this process into a search problem by repeatedly applying single-step retrosynthesis to intermediate molecules until the terminal precursors are available building blocks.
Thus, single-step retrosynthesis acts as the expansion operation for multi-step planning.

\subsection{Strategic Reasoning Design}
The reasoning process of RetroReasoner consists of four structured reasoning steps as shown in~\Cref{fig:RetroReasoner}: product analysis $\mathcal{R}_1$, identifying key substructures for bond formation $\mathcal{R}_2$, strategic bond disconnection $\mathcal{R}_3$, and mapping synthons to synthetic equivalents $\mathcal{R}_4$.
These steps are linked by natural-language linking texts $(\mathcal{L}_{12}, \mathcal{L}_{23}, \mathcal{L}_{34})$, which provide logical connections between the structured reasoning steps.
In $\mathcal{R}_1$, basic product information is listed that is used throughout the subsequent steps and content.
Specifically, it provides (1) atom-mapped SMILES (atoms indexed from 1 to $n$), (2) functional group information for the product (name, corresponding SMILES, and corresponding positions), and (3) statistics for the product SMILES string itself (number of rings, number of carbons, and the number of characters representing stereochemistry).
The linking text $\mathcal{L}_{12}$ transitions from this initial product information to the next step by inferring a reaction mechanism and the substructure formed during the reaction.
In the next step, $\mathcal{R}_2$, a candidate substructure is identified, narrowing the set of candidates for strategic bond disconnection.
The linking text $\mathcal{L}_{23}$ infers which bond from the candidate set should be selected.
In $\mathcal{R}_3$, a bond is selected from the candidates, and the corresponding synthons are generated.
The linking text $\mathcal{L}_{34}$ explains where the synthon can typically be obtained.
Finally, in $\mathcal{R}_4$, each synthon is mapped to its synthetic equivalent.
Each of these four structured steps is encapsulated in the \texttt{<PRODUCT\_INFO>},
\texttt{<CANDIDATE\_STRUCTURE>}, \texttt{<STRATEGIC\_BOND\_}\allowbreak\texttt{DISCONNECTION>}, and \texttt{<SYNTHETIC\_}\allowbreak\texttt{EQUIVALENT>} tags, as shown in~\Cref{fig:rationale_example} of~\Cref{appendix:sec:details_of_SyntheticRetro}.

\subsection{Supervised Fine-tuning with Strategic Reasoning}

\subsubsection{SyntheticRetro: Synthetic Retrosynthesis Rationale Generator}
\begin{figure*}[!t]
  % \vskip 0.2in
  \begin{center}
    % \centerline{
    \includegraphics[width=0.95\linewidth]{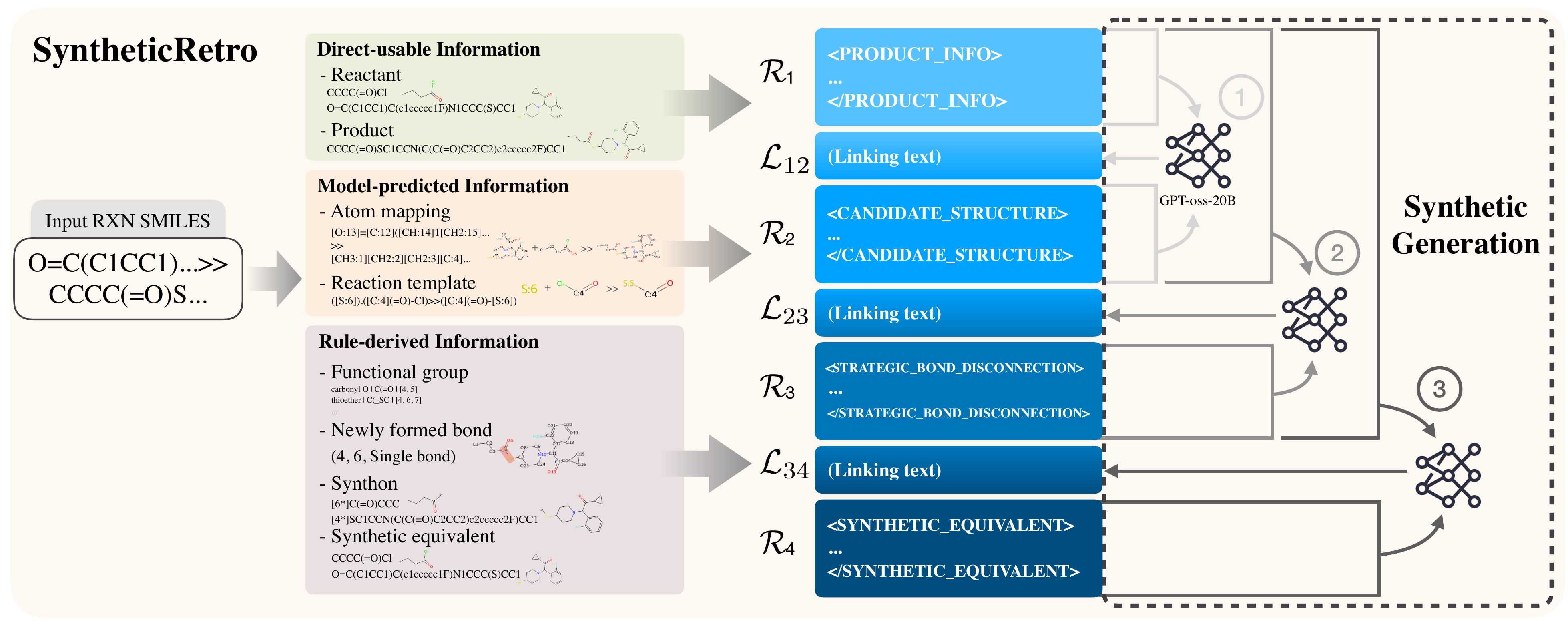}
    \caption{
      A schematic diagram of the generation process of SyntheticRetro, a chemist's strategy-based reasoning data generation process.
    }
    \label{fig:SyntheticRetro}
  \end{center}
\end{figure*}

SyntheticRetro generates a single-step retrosynthesis prediction rationale from the RXN SMILES.
SyntheticRetro first extracts three types of supporting information: direct-usable information, model-predicted information, and rule-derived information.
Direct-usable information is information that can be obtained directly from the RXN SMILES.
Model-predicted information is information obtained from deep learning-based models.
% LocalMapper~\cite{chen2024precise} is used to obtain atom mapping between reactants and product, along with the corresponding reaction template.
Rule-derived information is information obtained through rule-based post-processing.
% It includes functional groups, newly formed bond, synthons, and synthetic equivalents.
Using these three types of supporting information, SyntheticRetro leverages a general-purpose LLM (GPT-oss-20B)~\cite{agarwal2025gpt} to restructure them into a reasoning text.
An illustrative example is shown in~\Cref{fig:SyntheticRetro} and details of SyntheticRetro are described in~\Cref{appendix:sec:details_of_SyntheticRetro} with an example.

\paragraph{Synthetic Generation}
SyntheticRetro leverages GPT-oss-20B to generate the linking texts $\mathcal{L}_{12}, \mathcal{L}_{23}, \mathcal{L}_{34}$.
SyntheticRetro provides the general LLM with all supporting information and two consecutive structured reasoning steps $\mathcal{R}_i, \mathcal{R}_{i+1}$, instructs it to connect the content between the two steps, and then inserts the contents between the two steps.
First, it fills in the contents between $\mathcal{R}_1$ and $\mathcal{R}_2$, then it provides all contents up to $\mathcal{R}_2$ and $\mathcal{R}_3$ to the general LLM, instructs it to generate the contents connecting $\mathcal{R}_2$ and $\mathcal{R}_3$, and inserts the contents between $\mathcal{R}_2$ and $\mathcal{R}_3$.
Finally, it uses the general LLM to fill in all contents up to $\mathcal{R}_3$ and the contents between $\mathcal{R}_3$ and $\mathcal{R}_4$, completing the generation.
This generation process is repeated $n=15$ times for one reaction instance to obtain various paths to the next structured reasoning steps.
The generated data statistics are summarized in~\Cref{tab:dataset_statistics}, and the generation prompt is described in~\Cref{appendix:subsec:SyntheticRetro_generation_prompt}.

\subsubsection{Training Objective}
Let $\pi_\theta$ denote an LLM policy, $\mathbf{x}$ a product, and $\mathbf{y}=(\mathbf{y}^\text{reasoning}, \mathbf{y}^\text{reactant})=(y_1, y_2, \ldots, y_T)$ the corresponding ground-truth token sequence, which consists of reasoning text tokens $\mathbf{y}^\text{reasoning}$ and reactant SMILES tokens $\mathbf{y}^\text{reactant}$.
SFT minimizes the standard cross-entropy loss, defined as
\begin{equation}
\mathcal{L}_\text{SFT}(\theta) = -\frac{1}{T}\sum_{t=1}^{T}\log \pi_\theta (y_t|\mathbf{x}, y_{<t}).
\end{equation}

\subsection{Reinforcement Learning with Round-trip Accuracy}
The training adopts the reinforcement learning with verifiable rewards (RLVR) paradigm, in which an automatic verifier assigns rewards to policy-generated outputs that serve as the learning signal for policy updates.
For each training instance, a group of $G$ output sequences $\{\mathbf{y}_i\}_{i=1}^{G}$ is sampled from a behavior policy $\pi_{\theta_\text{old}}$, where $\mathbf{y}_i=(y_{(i, 1)}, y_{(i, 2)}, \dots)$ denotes the $i$-th sampled sequence.
Each sequence is assigned a scalar reward $R_i$, and token-level advantages are computed by normalizing rewards across the group: $\hat{A}_{i, t}=\frac{R_i-\text{mean}(\{R_i\}_{i=1}^G)}{\text{std}(\{R_i\}_{i=1}^G)}$, shared across all token positions $t$ within the same output sequence.
The RL objective is defined as
\begin{equation}
\label{eq:rl_objective}
\mathcal{J}_{\mathrm{RL}}(\theta)
=
\mathbb{E}_{\{\mathbf{y}_i\}\sim\pi_{\theta_{\text{old}}}}
\left[
\frac{1}{G}
\sum_{i=1}^{G}
\sum_{t=1}^{|\mathbf{y}_i|}
\ell_{i,t}(\theta)
\right],
\end{equation}
where the per-token loss follows a PPO~\cite{schulman2017proximal}-style clipped objective $\ell_{i,t}(\theta)
= \min\!\big(
r_{i,t}(\theta)\hat{A}_{i,t},\, \mathrm{clip}(r_{i,t}(\theta),1-\epsilon_\text{low},1+\epsilon_\text{high})
\hat{A}_{i,t}
\big)$, with probability ratio $r_{i,t}(\theta)=\frac{\pi_\theta(y_{i, t} | q, y_{i, <t})}{\pi_{\theta_\text{old}}(y_{i, t} | q, y_{i, <t})}$ and $q$ representing a user query.
In addition to the clipped term in~\Cref{eq:rl_objective}, training applies a KL-divergence penalty toward the reference policy and an entropy bonus to maintain exploration.
The KL-loss and entropy coefficients are both set to $0.05$, as reported in~\Cref{tab:hyperparameters_RL}.
The objective in~\Cref{eq:rl_objective} is optimized using round-trip accuracy as the reward.
Specifically, the policy generates a sequence of reactant SMILES tokens, $\hat{\mathbf{y}}^{\text{reactant}}$.
A separately trained round-trip model $f_{\phi}$ then takes $\hat{\mathbf{y}}^{\text{reactant}}$ as input and predicts the corresponding product SMILES.
% The round-trip reward $R^{\text{round-trip}}$ is computed using an identity function $\mathbb{I}(\cdot, \cdot)$ that checks whether the predicted product SMILES is identical to the ground-truth product SMILES $\mathbf{x}$.
To account for equivalent SMILES representations, molecular identity is determined after converting SMILES to standard InChI~\cite{heller2013inchi}.
Let $h(\cdot)$ denote the standard InChI conversion and define $\mathbb{I}(a, b)=\mathbf{1}[h(a) = h(b)]$.
The round-trip reward is computed as
% Formally,
\begin{equation}
\label{equation:round-trip}
R^{\text{round-trip}}=\mathbb{I}\Big(\mathbf{x}, f_{\phi}(\hat{\mathbf{y}}^{\text{reactant}})\Big).
\end{equation}
Training details, hyperparameters, and the cached round-trip verifier are described in~\Cref{appendix:sec:details_of_RetroReasoner}; details of the forward model $f_\phi$ are provided in~\Cref{appendix:sec:details_of_roundtrip}.

\subsection{Parallelized Multi-step Retrosynthesis Prediction}
\label{subsec:parallel_mcts}
Given a target molecule $m_{\text{target}}$ and a set of commercially available stock molecules $\mathcal{B}$, multi-step retrosynthesis prediction aims to find a synthesis route whose leaf molecules all belong to $\mathcal{B}$.
The search is conducted over an AND-OR tree, where OR nodes denote molecules and AND nodes denote reactions, and is performed via MCTS consisting of selection, expansion, evaluation, and backpropagation stages.
During selection, the tree is traversed from the root using a UCB score that balances exploitation and exploration over reaction nodes, and the expansion stage queries an LLM-based single-step retrosynthesis model $\pi_\theta$ at the leaf molecule to generate candidate reactant sets.
However, LLM-based single-step retrosynthesis models are computationally expensive in MCTS-based multi-step planning, since the proposal model \(\pi_\theta\) must be repeatedly queried during tree expansion.
To mitigate this expansion bottleneck, multiple expansion requests are generated in parallel and served as batched LLM queries, leveraging inference optimizations such as continuous batching~\cite{yu2022orca}, paged key-value caching~\cite{kwon2023efficient}, and optimized attention kernels~\cite{dao2022flashattention}.
Parallel expansion follows the core idea of WU-UCT~\cite{liu2018watch}, in which each node maintains an unobserved count \(O\) representing the number of in-flight simulations assigned to that node. The standard UCB score is augmented with these unobserved counts as

\begin{equation}
\text{UCB}(r)_\text{WU} =
\frac{V_r}{N_r}
+
c \cdot p_r \cdot
\frac{\sqrt{N_{\text{parent}} + O_{\text{parent}}}}
{1 + N_r + O_r},
\end{equation}

where $V_r$ denotes the cumulative value backed up through reaction node $r$, $N_r$ its visit count, $N_{\text{parent}}$ the visit count of the parent molecule node, $p_r$ the reaction prior, and $c$ the exploration constant.
The term \(O_r\) lowers the exploration bonus of branches already being expanded, encouraging different workers to explore different tree regions.
As a result, parallel MCTS reduces redundant LLM calls, better utilizes batched inference, and improves route diversity.
See~\Cref{appendix:sec:details_of_mcts} for details.
\endgroup

\section{Experiments}
\label{sec:experiments}

\subsection{Baseline Models}
RetroReasoner is evaluated against four categories of baselines: \emph{Molecular Prediction LLMs}, \emph{Molecular Reasoning LLMs}, \emph{General Purpose LLMs}, and \emph{Expert Models}.
\emph{Molecular Prediction LLMs} directly generate reactants without explicit reasoning and include LlaSMol~\cite{yu2024llasmol}, Mol-Instructions~\cite{fang2023mol}, BioT5+~\cite{pei2024biot5+}, PRESTO~\cite{cao2024presto}, and Mol-LLM~\cite{lee2025mol}.
\emph{Molecular Reasoning LLMs} incorporate reasoning into retrosynthesis prediction, including Chem-R~\cite{wang2025chem}, ChemDFM~\cite{zhao2024chemdfm}, and ether0~\cite{narayanan2025training}.
\emph{General Purpose LLMs} are prompted foundation models, including OpenAI-o3~\cite{openaio3}, GPT-5-mini~\cite{singh2025openai}, GPT-oss-120B~\cite{agarwal2025gpt}, Qwen3-8B, and Qwen3-235B-A22B~\cite{yang2025qwen3}.
For OpenAI-o3, only greedy metrics are reported due to the high cost of sampling 100 outputs per instance (\texttt{>}500\$), and GPT-5-mini results are used for sampling-based comparison~\cite{singh2025openai}.
\emph{Expert Models} are non-LLM retrosynthesis models, including RetroSynFlow~\cite{yadav2026retro} and G2S-HCVAE~\cite{wang2026enhancing}, re-trained on the same training data as RetroReasoner.
A Prediction-Only variant is also included as a controlled baseline; it shares the same architecture, data, and optimization procedure as RetroReasoner but removes explicit reasoning generation.
Prompts for \emph{General Purpose LLMs} are provided in~\Cref{appendix:sec:prompts}.
Training details for the expert models are provided in~\Cref{appendix:sec:experiment_details}.

For multi-step retrosynthesis, RetroInText~\cite{kang2025retrointext} is additionally included as a reproducible planning baseline, using MolT5~\cite{edwards2022translation} for single-step retrosynthesis and MCTS with a fusion-based value function.
Evaluation is conducted on the same fixed PaRoutes~\cite{genheden2022paroutes} set-n1 and set-n5 target subsets used for RetroReasoner.
Retro-R1~\cite{liu2026retro} is excluded because its trained search-policy weights are not publicly available, preventing faithful reproduction of the complete method.

\subsection{Evaluation Metrics}
\paragraph{Single-step Retrosynthesis}
Single-step retrosynthesis is evaluated under both greedy decoding and sampling.
Greedy metrics assess the top-1 proposal, while sampling metrics evaluate whether valid and diverse solutions can be found among 100 candidates.
\textsf{Exact@1↑} and \textsf{Exact@100↑} measure matches to the labeled reactants based on the standard InChI identity function $\mathbb{I}$ defined in~\Cref{equation:round-trip}, for the top-1 prediction and among 100 samples, respectively.
\textsf{Round-trip@1↑} and \textsf{Round-trip@100↑} measure whether the predicted reactants are mapped back to the input product by a forward reaction model \(f_\phi\).
\textsf{Feasible Ratio↑} is a forward-model-based proxy for reaction feasibility, measuring the fraction of sampled reactants that reconstruct the target product under the forward verifier.
\textsf{Template Diversity↑} counts the number of distinct canonical reaction templates among feasible candidates, reflecting the diversity of valid disconnection patterns.
All metrics are averaged over evaluation instances, with detailed calculations provided in~\Cref{appendix:subsec:metric_calculation_details}.

\paragraph{Multi-step Retrosynthesis}
Multi-step retrosynthesis is evaluated using three PaRoutes metrics~\cite{genheden2022paroutes}.
\textsf{Solved Targets↑} counts targets for which at least one fully purchasable route is found.
\textsf{Routes Extracted↑} counts all complete routes extracted from the search trees.
\textsf{Number of Clusters↑} measures route diversity by clustering extracted routes based on pairwise route distances and summing the number of distinct clusters.
Further details are provided in~\Cref{appendix:subsec:metric_calculation_details}.

\subsection{Evaluation Dataset}
We adopt ORDerly~\cite{wigh2024orderly}, a curated collection of high-quality reaction instances, as the base dataset.
To mitigate evaluation bias arising from the multi-label nature of retrosynthesis, instances in which a single product corresponds to multiple distinct valid reactant sets are excluded from the validation and test sets.
Instances used to train the baseline models are also excluded from the test set.
The main evaluation dataset is constructed by randomly selecting 500 instances from the remaining filtered test set.
To assess generalization ability for challenging cases, two additional hard evaluation subsets are also constructed, consisting of rare reaction template instances and rare atom or token instances.
For the rare reaction template subset, canonical reaction template frequencies are computed over all ORDerly train and test instances, and 100 instances are sampled in total, with 50 drawn from cases whose canonical reaction template frequency lies in the range one to three and 50 drawn from those in the range four to six.
For the rare atom or token subset, each product SMILES is assigned a rarity score based on the extent to which it is composed of rare $n$-gram tokens, with higher scores given to sequences containing a larger proportion of rare $n$-grams.
Instances are then sorted by this score, and 100 instances are selected by taking the top 50 under the rare 2-gram scoring and the top 50 under the rare 3-gram scoring.
For multi-step retrosynthesis evaluation, 500 target molecules are sampled from each of the PaRoutes set-n1 and set-n5 benchmarks~\cite{genheden2022paroutes}.
A fixed subset is used to ensure a controlled and repeatable comparison across methods under the same MCTS search budget.
The details of the evaluation dataset creation are described in~\Cref{appendix:subsec:evaluation_dataset_generation}.

\subsection{Hyperparameters}
BFloat16~\cite{google2019bfloat16} weight precision is used with full parameter tuning on 8 NVIDIA H100 GPUs.
Identical training hyperparameters are applied across all ablation studies and the Prediction-Only baseline.
For greedy metrics, greedy decoding is used to obtain a single top-1 reactant suggestion per instance.
For sampling-based metrics, 100 reactant suggestions are sampled per instance with temperature 1.2, matching the setting used by RetroReasoner, unless the corresponding open evaluation code specifies different sampling parameters.
For all MCTS-based comparisons, including RetroInText, MCTS is run with a maximum of 300 iterations, a beam width of 10, and a maximum search depth of 8.
Training hyperparameters are provided in~\Cref{appendix:sec:details_of_RetroReasoner,appendix:sec:details_of_roundtrip}, MCTS settings in~\Cref{appendix:sec:details_of_mcts}, and evaluation and expert-model settings in~\Cref{appendix:sec:experiment_details}.

\subsection{Results}
This section presents the main evaluation, rare template and rare atom/token evaluations, multi-step retrosynthesis results, the round-trip reward ablation in~\Cref{subsubsec:roundtrip_reward}, and the human expert evaluation in~\Cref{subsubsec:human_expert_evaluation}.
\Cref{appendix:sec:additional_experimental_results} provides supplementary reasoning, linking-text, model-size, and sampling ablations, as well as forward-model, inference-speed, confidence-interval, benchmark, qualitative, intermediate-step, and LLM-judge analyses.
It supplements the round-trip reward results with forward-model error and external-verifier analyses, and the human evaluation with detailed procedures and qualitative expert feedback.

\subsubsection{Main Evaluation}
\label{subsubsec:main_evaluation}
\begin{table*}[!t]
  \begin{center}
    \begin{small}
        \resizebox{0.96\linewidth}{!}{%
\begin{tabular}{@{}lllllll@{}}
\toprule
                                    & \multicolumn{2}{c}{Greedy Metrics} & \multicolumn{4}{c}{Sampling Metrics}   \\ \midrule
 &
  \textsf{Exact@1↑} &
  \textsf{Round-trip@1↑} &
  \textsf{Exact@100↑} &
  \textsf{Round-trip@100↑} &
  \textsf{Feasible Ratio↑} &
  \textsf{Template Diversity↑} \\ \midrule
\emph{Molecular Prediction LLMs} &
  \multicolumn{1}{c}{} &
  \multicolumn{1}{c}{} &
  \multicolumn{1}{c}{} &
  \multicolumn{1}{c}{} &
  \multicolumn{1}{c}{} &
  \multicolumn{1}{c}{} \\
LlaSMol~\cite{yu2024llasmol} &
  \textbf{0.254} &
  \textbf{0.668} &
  \textbf{0.584} &
  \textbf{0.926} &
  \textbf{0.437} &
  \textbf{7.814} \\
Mol-Instructions~\cite{fang2023mol} & 0.008                & 0.036       & 0.038 & 0.188 & 0.007 & 0.290          \\
BioT5+~\cite{pei2024biot5+}         & 0.098                & 0.388       & 0.166 & 0.660 & 0.366 & 1.198          \\
PRESTO~\cite{cao2024presto}         & 0.006                & 0.186       & 0.012 & 0.458 & 0.434 & 0.496          \\
Mol-LLM~\cite{lee2025mol}           & 0.016                & 0.036       & 0.066 & 0.154 & 0.015 & 0.192          \\ \midrule
\emph{Molecular Reasoning LLMs}     &                      &             &       &       &       &                \\
Chem-R~\cite{wang2025chem} &
  \textbf{0.214} &
  \textbf{0.624} &
  \textbf{0.382} &
  \textbf{0.902} &
  \textbf{0.645} &
  2.510 \\
ChemDFM~\cite{zhao2024chemdfm}      & 0.140                & 0.360          & 0.338 & 0.864 & 0.380 & \textbf{2.594} \\
ether0~\cite{narayanan2025training} & 0.034                & 0.488          & 0.128 & 0.758 & 0.487 & 1.330          \\ \midrule
% RetroDFM-R~\cite{zhang2025reasoning}& \textbf{0.488}       & \textbf{0.834} & \textbf{0.646} & \textbf{0.922} & \textbf{0.824} & 0.958          \\ \midrule
\emph{General Purpose LLMs}         &                      &                &       &       &       &                \\
OpenAI-o3~\cite{openaio3}           & 0.010                & 0.022          & -     & -     & -     & -              \\
GPT-5-mini~\cite{singh2025openai} &
  0.010 &
  \textbf{0.038} &
  \textbf{0.152} &
  \textbf{0.498} &
  \textbf{0.031} &
  \textbf{0.794} \\
GPT-oss-120B~\cite{agarwal2025gpt}  & \textbf{0.014}       & 0.028       & 0.070 & 0.164 & 0.004 & 0.190          \\
Qwen3-8B~\cite{yang2025qwen3}       & 0.000                & 0.008       & 0.006 & 0.096 & 0.006 & 0.074          \\
Qwen3-235B-A22B~\cite{yang2025qwen3} & 0.006                & 0.024       & 0.002 & 0.006 & 0.000 & 0.004          \\ \midrule
\emph{Expert Models} &
   &
   &
   &
   &
   &
   \\
RetroSynFlow~\cite{yadav2026retro} &
  0.424 &
  0.528 &
  0.692 &
  0.702 &
  \textbf{0.533} &
  3.358 \\
G2S-HCVAE~\cite{wang2026enhancing} &
  \textbf{0.492} &
  \textbf{0.814} &
  \textbf{0.704} &
  \textbf{0.944} &
  0.482 &
  \textbf{3.872} \\ \midrule
Prediction-Only (SFT)               & 0.482                & 0.784       & 0.678 & 0.950 & 0.774 & 2.562          \\
Prediction-Only (RL)                & 0.486                & 0.802       & 0.662 & 0.936 & 0.785 & 2.324          \\
RetroReasoner (SFT) &
  0.512 &
  0.812 &
  \textbf{0.734} &
  0.944 &
  0.765 &
  \textbf{3.898} \\
RetroReasoner (RL) &
  \textbf{0.526} &
  \textbf{0.826} &
  0.724 &
  \textbf{0.952} &
  \textbf{0.786} &
  3.186 \\ \bottomrule
\end{tabular}%
        }
    \end{small}
  \end{center}
  \caption{Main evaluation comparison. Performance in each category is shown. The highest value within each category is highlighted in \textbf{bold}. Sampling metrics for RetroReasoner and Prediction-Only use the default decoding setting with temperature 1.2; the feasibility-diversity trade-off across decoding settings is analyzed in~\Cref{fig:feasibility_diversity_tradeoff}.}
  \label{tab:main_ID}
  \vskip -0.05in
\end{table*}
\begin{figure}[!t]
  % \vskip 0.2in
  \begin{center}
    % \centerline{
    \includegraphics[width=0.95\columnwidth]{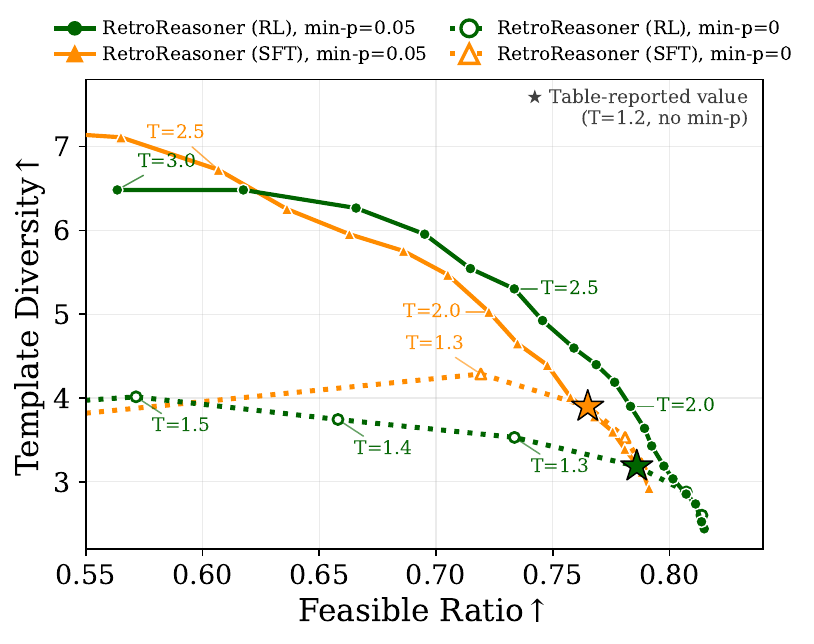}
    \caption{
      Feasibility-diversity trade-off of RetroReasoner (SFT) and RetroReasoner (RL). \textsf{Feasible Ratio↑} and \textsf{Template Diversity↑} are measured while varying the sampling temperature from 1.2 to 3.0 under min-$p$ decoding ($p=0.05$). RetroReasoner (RL) achieves higher \textsf{Template Diversity↑} at matched \textsf{Feasible Ratio↑} levels, forming a more favorable empirical Pareto frontier than SFT.}
    \label{fig:feasibility_diversity_tradeoff}
  \end{center}
\end{figure}
% \paragraph{SFT narrows exploration to the feasible space, and RL further optimizes the model toward higher round-trip reward.}
\paragraph{RL improves accuracy and round-trip consistency while shifting the feasibility-diversity trade-off.}
\Cref{tab:main_ID} shows that RetroReasoner generally outperforms \emph{Molecular Reasoning LLMs} and Prediction-Only, with particularly clear gains in \textsf{Exact@100↑} and \textsf{Template Diversity↑}.
These results indicate that explicit disconnection reasoning broadens the coverage of plausible reactant solutions beyond direct prediction.
RL further improves \textsf{Exact@1↑}, \textsf{Round-trip@1↑}, and \textsf{Feasible Ratio↑}, but decreases \textsf{Template Diversity↑} from 3.898 to 3.186 under the default sampling configuration.
To determine whether this reduction reflects an inherent loss of diversity or a decoding-dependent trade-off, \Cref{fig:feasibility_diversity_tradeoff} compares SFT and RL across a broader range of sampling conditions.
Min-p decoding is applied to suppress low-probability out-of-distribution tokens while allowing the sampling temperature to vary.
At comparable \textsf{Feasible Ratio↑} levels, RetroReasoner (RL) consistently achieves higher \textsf{Template Diversity↑} than RetroReasoner (SFT), forming a more favorable empirical Pareto frontier.
Thus, the lower diversity at the default operating point does not persist when feasibility and diversity are compared under controlled decoding conditions.
Detailed token-entropy and decoding analyses are provided in~\Cref{appendix:subsubsec:effect_of_sampling_hyperparameters}.

% \Cref{tab:main_ID} presents the main evaluation results.
% For both Prediction-Only and RetroReasoner, the SFT trained and RL trained variants are reported separately according to the model name.
% Across metrics, RetroReasoner generally outperforms \emph{Molecular Reasoning LLMs} and Prediction-Only, with particularly clear gains in \textsf{Exact@100↑} and \textsf{Template Diversity↑}.
% This trend indicates that strategic disconnection reasoning broadens the set of feasible solutions and improves coverage of the chemically plausible search space relative to generation without explicit reasoning.
% Comparing SFT and RL variants, RL improves most accuracy-based metrics, while \textsf{Template Diversity↑} decreases.
% This indicates that during the SFT stage, the model is made to explore the feasible reactant space, and in the RL stage, it is directed to explore the more feasible regions within that feasible reactant space.

\subsubsection{Rare Template and Rare Atom/Token}
\begin{table*}[!t]
  \begin{center}
    \begin{small}
        \resizebox{0.96\linewidth}{!}{%
\begin{tabular}{@{}lllllll@{}}
\toprule
                                    & \textsf{Exact@1↑}    & \textsf{Round-trip@1↑} & \textsf{Exact@100↑}  & \textsf{Round-trip@100↑} & \textsf{Feasible Ratio↑} & \textsf{Template Diversity↑} \\ \midrule
\textit{Rare Template Evaluation}   & \multicolumn{1}{c}{} & \multicolumn{1}{c}{}   & \multicolumn{1}{c}{} & \multicolumn{1}{c}{}     & \multicolumn{1}{c}{}     & \multicolumn{1}{c}{}         \\
Prediction-Only (SFT)               & 0.12                 & 0.65                   & 0.24                 & 0.92                     & 0.560                    & 3.20                         \\
Prediction-Only (RL)                & 0.12                 & 0.64                   & 0.22                 & 0.89                     & 0.621                    & 2.66                         \\
RetroReasoner (SFT)                 & \textbf{0.14}        & 0.63                   & \textbf{0.41}        & 0.92                     & 0.587                    & \textbf{5.20}                \\
RetroReasoner (RL)                  & 0.13                 & \textbf{0.70}          & 0.39                 & \textbf{0.95}            & \textbf{0.625}           & 4.25                         \\ \midrule
\textit{Rare Atom/Token Evaluation} &                      &                        &                      &                          &                          &                              \\
Prediction-Only (SFT)               & 0.36                 & 0.49                   & 0.52                 & 0.82                     & 0.458                    & 2.02                         \\
Prediction-Only (RL)                & 0.35                 & 0.49                   & 0.49                 & 0.78                     & 0.478                    & 1.97                         \\
RetroReasoner (SFT)                 & \textbf{0.43}        & \textbf{0.64}          & 0.66                 & 0.91                     & 0.530                    & \textbf{3.43}                \\
RetroReasoner (RL)                  & 0.38                 & 0.60                   & \textbf{0.67}        & \textbf{0.92}            & \textbf{0.557}           & 2.97                         \\ \bottomrule
\end{tabular}%
        }
    \end{small}
  \end{center}
  \caption{Performance comparison of hard instances consisting of rare reaction templates and rare atoms/tokens. The highest metric performance for each evaluation dataset is highlighted in \textbf{bold}.}
  \label{tab:main_hard}
  % \vskip -0.1in
\end{table*}

\paragraph{RetroReasoner is more robust on hard instances.}
\Cref{tab:main_hard} shows evaluation results using the rare reaction template and atom/token instances.
The performance gap between RetroReasoner and Prediction-Only widens in this challenging problem setting.
Notably, in the rare atom/token evaluation, RetroReasoner demonstrates a higher \textsf{Feasible Ratio↑} and \textsf{Template Diversity↑} compared to Prediction-Only.
These results indicate that RetroReasoner is more robust in handling challenging cases, likely due to the inclusion of strategic reasoning in its design.

\subsubsection{Multi-step Retrosynthesis}
\begin{table*}[!t]
  \begin{center}
    \begin{small}
        \resizebox{0.96\linewidth}{!}{%
\begin{tabular}{@{}lccccc@{}}
\toprule
Single-step Model                        & Search Method & Value Function                         & \textsf{Solved Targets↑} & \textsf{Routes Extracted↑} & \textsf{Number of Clusters↑} \\ \midrule
\textit{PaRoutes (500 subset of set-n1)} &               &                                        &                          &                            &                              \\
MolT5 (RetroInText)~\cite{kang2025retrointext}                      & MCTS          & \multicolumn{1}{l}{RetroInText fusion} & \textbf{449}             & 2659                       & 694                          \\
Prediction-Only (SFT)                    & MCTS          & Retro* default                         & \textit{438}             & \textit{3298}              & \textit{874}                 \\
Prediction-Only (RL)                     & MCTS          & Retro* default                         & \textit{414}             & \textit{2659}              & \textit{752}                 \\
RetroReasoner (SFT)                      & MCTS          & Retro* default                         & \textit{415}             & \textit{4432}              & \textit{1130}                \\
RetroReasoner (RL)                       & MCTS          & Retro* default                         & \textit{423}             & \textit{\textbf{4612}}     & \textit{\textbf{1150}}       \\ \midrule
\textit{PaRoutes (500 subset of set-n5)} &               &                                        &                          &                            &                              \\
MolT5 (RetroInText)~\cite{kang2025retrointext}                      & MCTS          & \multicolumn{1}{l}{RetroInText fusion} & \textbf{451}             & 2651                       & 742                          \\
Prediction-Only (SFT)                    & MCTS          & Retro* default                         & 429                      & 3636                       & 888                          \\
Prediction-Only (RL)                     & MCTS          & Retro* default                         & 420                      & 3289                       & 915                          \\
RetroReasoner (SFT)                      & MCTS          & Retro* default                         & 398                      & \textbf{5246}              & 1243                         \\
RetroReasoner (RL)                       & MCTS          & Retro* default                         & 416                      & 5080                       & \textbf{1332}                \\ \bottomrule
\end{tabular}%
        }
    \end{small}
  \end{center}
  \caption{Multi-step performance comparison. The highest metric performance is highlighted in \textbf{bold}.}
  \label{tab:multistep_comparison_n1}
  % \vskip -0.1in
\end{table*}
\begin{figure*}[!t]
  % \vskip -0.2in
  \begin{center}
    % \centerline{
    \begin{minipage}{0.99\textwidth}
        \centering
        \includegraphics[width=0.85\linewidth]{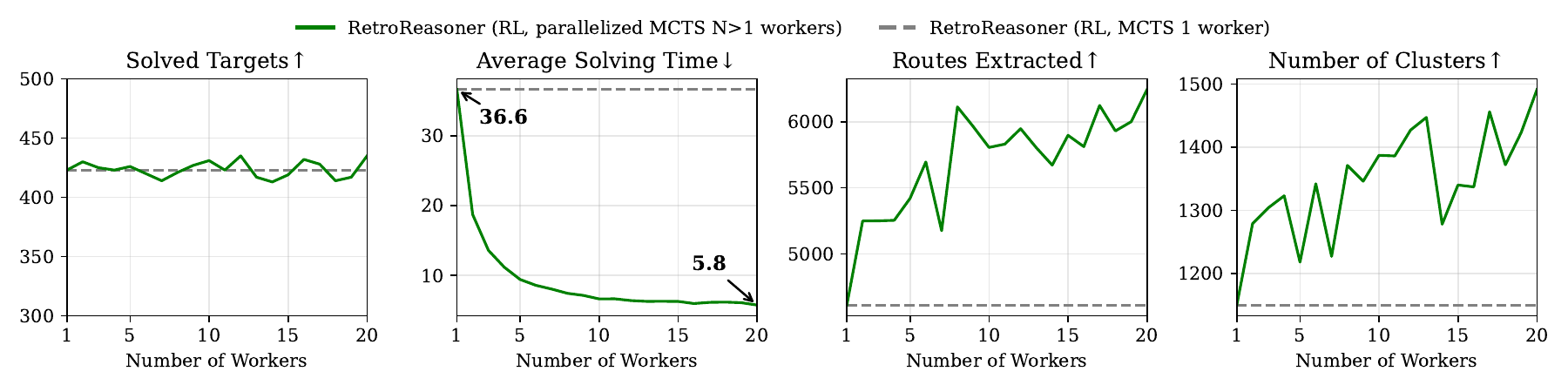}
    \end{minipage} \hfill
    \begin{minipage}{0.99\textwidth}
        \centering
        \includegraphics[width=0.85\linewidth]{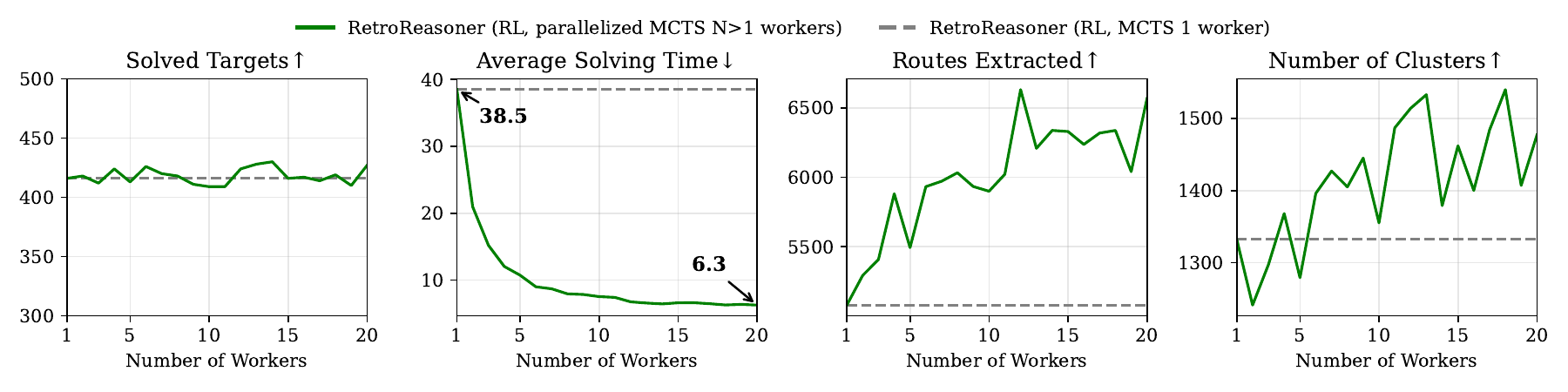}
    \end{minipage}
    \caption{
      Performance comparison across different numbers of MCTS workers on the PaRoutes set-n1 and set-n5 subsets.}
    \label{fig:mcts_comparison_n1}
  \end{center}
  \vskip -0.1in  
\end{figure*}

% \input{figure_tex/mcts_comparison_curve_n5}
% \paragraph{Diversity in single-step predictions also increases diversity in multi-step synthetic route prediction.}
% \Cref{tab:multistep_comparison_n1} reports the results when using Prediction-Only and RetroReasoner as the single-step model.
% The results show that the \textsf{Solved Targets↑} remains comparable across models, whereas \textsf{Routes Extracted↑} and \textsf{Number of Clusters↑} differ substantially.
% This indicates that the stronger diversity observed in single-step predictions carries over to multi-step planning, enabling the search to discover more diverse synthetic routes.
% In addition, the use of the round-trip reward in RL allows the model to identify a broader set of feasible solutions, leading to a slightly higher \textsf{Solved Targets↑}.

\paragraph{The lower single-step diversity under default RL decoding does not translate into lower multi-step route diversity.}

Although RetroReasoner (RL) shows lower \textsf{Template Diversity↑} than its SFT counterpart under the default single-step decoding setting, this reduction does not propagate to final route diversity.
As shown in \Cref{tab:multistep_comparison_n1}, the \textsf{Number of Clusters↑} increases from 1,130 to 1,150 on set-n1 and from 1,243 to 1,332 on set-n5 after RL.
\textsf{Routes Extracted↑} also increases on set-n1 (4,432 to 4,612), although it slightly decreases on set-n5 (5,246 to 5,080).
In addition, RL increases \textsf{Solved Targets↑} relative to SFT on both PaRoutes subsets.
These results indicate that the lower raw single-step \textsf{Template Diversity↑} at one decoding operating point does not imply reduced diversity of the complete routes discovered by search.
RetroInText is additionally included as a reproducible external multi-step planning baseline.
RetroReasoner (RL) solves fewer targets than RetroInText (423 vs.\ 449 on set-n1 and 416 vs.\ 451 on set-n5), but discovers substantially more complete routes (4,612 vs.\ 2,659 and 5,080 vs.\ 2,651) and route clusters (1,150 vs.\ 694 and 1,332 vs.\ 742).
Accordingly, the multi-step advantage of RetroReasoner is primarily reflected in broader route exploration and route diversity rather than uniformly higher target solvability.

\paragraph{Optimizing LLM inference within parallelized tree search improves both search efficiency and synthetic route diversity.}
\Cref{fig:mcts_comparison_n1} shows performance on the PaRoutes set-n1 and set-n5 subsets when using a single MCTS search worker and when applying the parallelization method introduced in~\Cref{subsec:parallel_mcts} with varying numbers of workers.
While the \textsf{Solved Targets↑} is largely maintained as the number of workers increases, the average solving time is reduced by nearly sixfold.
Moreover, by introducing the unobserved count, workers are encouraged to explore branches that are not currently being searched by other workers.
As a result, \textsf{Routes Extracted↑} and \textsf{Number of Clusters↑}, which measure the diversity of synthetic routes, gradually increase.
This suggests that the proposed parallelization strategy is also beneficial in terms of route diversity.

\subsubsection{Effect of Round-trip Reward}
\label{subsubsec:roundtrip_reward}
\begin{table*}[!t]
  \begin{center}
    \begin{small}
        \resizebox{\linewidth}{!}{%
\begin{tabular}{@{}lllllll@{}}
\toprule
 &
  \multicolumn{1}{c}{\textsf{Exact@1↑}} &
  \multicolumn{1}{c}{\textsf{Round-trip@1↑}} &
  \multicolumn{1}{c}{\textsf{Exact@100↑}} &
  \multicolumn{1}{c}{\textsf{Round-trip@100↑}} &
  \multicolumn{1}{c}{\textsf{Feasible Ratio↑}} &
  \multicolumn{1}{c}{\textsf{Template Diversity↑}} \\ \midrule
RetroReasoner (RL) &
  \textbf{0.526} &
  \textbf{0.826} &
  \textbf{0.724} &
  \textbf{0.952} &
  0.786 &
  \textbf{3.186} \\
RetroReasoner (RL, w/ $R^{\text{exact}}$) &
  0.524 &
  0.824 &
  0.696 &
  0.934 &
  \textbf{0.812} &
  2.430 \\ \bottomrule
\end{tabular}%
        }
    \end{small}
  \end{center}
  \caption{Performance comparison of round-trip reward $R^\text{round-trip}$ in RL. When $R^\text{round-trip}$ are not used, the reward is the exact match between the model's predicted reactant and the reactant labeled in the data instance ($R^\text{exact}$). The best performance is highlighted in \textbf{bold}.}
  \label{tab:ablation_roundtrip_reward}
  % \vskip -0.1in
\end{table*}
\begin{table*}[!t]
        \resizebox{\linewidth}{!}{%
\begin{tabular}{@{}lllll@{}}
\toprule
                     & Our round-trip model & MolecularTransformer (MT) & Our round-trip model & MolecularTransformer (MT) \\ \midrule
\textit{}            & \textsf{Round-trip@1↑} & \textsf{Round-trip@1↑}  & \textsf{Round-trip@100↑} & \textsf{Round-trip@100↑}           \\ \midrule
Prediction-Only (RL) & 0.802                & 0.740                     & 0.936                & 0.862                     \\
RetroReasoner (RL)   & \textbf{0.826}       & \textbf{0.748}            & \textbf{0.952}       & \textbf{0.896}            \\ \bottomrule
\end{tabular}%
        }
  \caption{Comparison of round-trip metrics under our internal forward verifier and MolecularTransformer (MT), an independent external verifier not trained on ORDerly.}
  \label{tab:roundtrip_MT_comparison}
  % \vskip -0.1in
\end{table*}
\paragraph{Round-trip reward enhances feasible molecule space exploration.}
\Cref{tab:ablation_roundtrip_reward} compares the results of using the round-trip reward versus the exact match reward for the reactant.
When $R^\text{round-trip}$ is not used (i.e., $R^\text{exact}$ is used), \textsf{Feasible Ratio↑} increases, but \textsf{Exact@100↑}, \textsf{Round-trip@100↑}, and \textsf{Template Diversity↑} decrease significantly compared with the results obtained using $R^\text{round-trip}$.
These findings suggest that the $R^\text{round-trip}$ helps explore a broader feasible reactant space during policy updates.
Without it, the model fails to explore this broader space and instead narrows its focus, ultimately handling a smaller reactant space.
This comparison isolates the effect of the reward function within RL and should be distinguished from the SFT-RL comparison in~\Cref{subsubsec:main_evaluation}: round-trip-based RL preserves substantially more diversity than exact-match-based RL, even though its default decoding point remains less diverse than the SFT model.

\paragraph{The effectiveness of the round-trip reward is consistent under an independent forward verifier.}

Because the round-trip reward and several evaluation metrics rely on a learned forward model, predictions are additionally evaluated with Molecular Transformer (MT)~\cite{schwaller2019molecular}, an independent verifier not trained on ORDerly.
\Cref{tab:roundtrip_MT_comparison} reports \textsf{Round-trip@1↑} and \textsf{Round-trip@100↑} under both validators.
RetroReasoner (RL) maintains the same advantage over Prediction-Only (RL) under MT, improving \textsf{Round-trip@1↑} from 0.740 to 0.748 and \textsf{Round-trip@100↑} from 0.862 to 0.896.
These results suggest that the observed gains are not specific to the internal verifier, while round-trip consistency remains a proxy for laboratory feasibility.
Detailed results are provided in~\Cref{appendix:subsec:effects_of_using_an_external_forward_model}.

% \subsubsection{Failure Analysis}
% \input{table_tex/failure_analysis}
% \paragraph{Most exact-match errors arise from alternative retrosynthetic decisions rather than internally inconsistent reasoning.}
% Complete reasoning trajectories from RetroReasoner (RL) are analyzed on the 500-instance test set under greedy decoding. Internal consistency checks verify the selected disconnection, resulting synthons, synthetic equivalents, and final answer, while comparison with the SyntheticRetro reference identifies the first deviating step.
% As shown in~\Cref{tab:failure_analysis}, among 237 Exact@1 errors, 97.9\% pass all internal consistency checks, with most deviations occurring at alternative bond disconnections (49.4\%) or synthetic-equivalent choices (38.8\%).
% Moreover, 69.2\% (164/237) are round-trip consistent.
% In contrast, only 0.4\% of \textsf{Exact@1↑}-correct predictions fail an internal consistency check, and 97.0\% match the reference reasoning throughout.
% These results indicate that many exact-match failures reflect alternative, internally consistent retrosynthetic decisions rather than malformed reasoning, highlighting the limitation of single-reference evaluation in retrosynthesis.

\subsubsection{Human Expert Evaluation}
\label{subsubsec:human_expert_evaluation}
\begin{table}[!t]
  \begin{center}
    \begin{small}
        \resizebox{\columnwidth}{!}{%
\begin{tabular}{@{}cccc@{}}
\toprule
\textbf{} & \textbf{RetroReasoner win rate} & \textbf{Chem-R win rate} & \textbf{Tie} \\ \midrule
Person A  & 88\%                            & 4\%                      & 8\%          \\
Person B  & 64\%                            & 12\%                     & 24\%         \\
Person C  & 20\%                            & 80\%                     & 0\%          \\
Person D  & 84\%                            & 10\%                     & 6\%          \\
Person E  & 20\%                            & 28\%                     & 52\%         \\ \midrule
Average   & 55.2\%                          & 26.8\%                   & 18\%         \\ \bottomrule
\end{tabular}%
        }
    \end{small}
  \end{center}
  \caption{Comparison of win rates between Chem-R, the best-performing \emph{Molecular Reasoning LLMs}, and RetroReasoner through A/B testing of the reasoning contents.}
  \label{tab:human_evaluation}
  \vskip -0.1in
\end{table}
To further assess whether the improved prediction performance is accompanied by higher-quality chemical reasoning, a human evaluation is performed against Chem-R, the strongest reasoning baseline in the comparison except RetroReasoner.
\Cref{tab:human_evaluation} reports the results of a blind pairwise evaluation conducted by five experts.
The evaluation set includes 33 instances where both methods are correct, 9 instances where only RetroReasoner is correct, and 8 instances where only Chem-R is correct.
In each case, both the reasoning process and the predicted reactants are provided, and the evaluators are asked to choose the response that demonstrates better chemical reasoning.
As shown in~\Cref{tab:human_evaluation}, RetroReasoner is preferred over Chem-R on average, supporting the higher overall quality of its single-step retrosynthetic reasoning and reactant predictions.
Detailed evaluation instructions and qualitative expert feedback are provided in~\Cref{appendix:subsec:details_of_human_expert_evaluation}.
% demonstrating that strategy-guided reasoning improves the reasoning quality itself, beyond its gains in prediction performance.

% \subsection{Analysis}

% \subsubsection{Model Size}
% \paragraph{Large model size are essential for learning the chemist's reasoning strategy.}
% \Cref{fig:model_size_ablation} compares the learning curves of two different model sizes: Qwen3-8B, the base model for RetroReasoner, and Qwen3-1.7B.
% While both models show improvement during training, a clear performance gap emerges toward the end, indicating that a sufficiently large model size is crucial for effectively learning a reasoning model that mimics the chemist's strategy.

% \subsubsection{Qualitative Comparison}
% In progress.
% (Qualitative Comparison)
\section{Conclusion}
\label{sec:conclusion}
% In this study, we develop RetroReasoner, a reasoning model for retrosynthesis prediction in organic synthesis, based on the strategies used by chemists.
% To train RetroReasoner, SyntheticRetro, a framework for generating SFT training data, is proposed and used to generate training data for the SFT.
% Afterward, performance is enhanced through RL using round-trip rewards.
% This approach demonstrates greater effectiveness than direct reactant prediction, particularly when handling rare reaction types and atoms or tokens.
% We believe that this work will be valuable for the development of agent systems addressing future challenges in organic synthesis prediction and planning.
This study presents RetroReasoner, a retrosynthetic reasoning model that follows chemists' disconnection-based strategies for organic synthesis.
SyntheticRetro is introduced to generate structured reasoning data for supervised fine-tuning, and RetroReasoner is further optimized through reinforcement learning with a round-trip reward.
Experimental results show that the proposed strategic reasoning model outperforms direct reactant prediction as well as previous reasoning approaches, particularly on challenging cases involving rare reaction types and rare atoms or tokens.
RetroReasoner also extends to multi-step retrosynthetic planning, where LLM inference optimization combined with parallelized MCTS reduces solving time and improves route diversity.
These results connect strategic single-step reasoning to broader route exploration and highlight RetroReasoner's potential for future agent systems in organic synthesis prediction and planning.
\Needspace{0.38\textheight}
\section*{Limitation}
\label{sec:limitation}
Although RetroReasoner, a reasoning LLM based on chemists' strategy, is developed in this study, several limitations exist.
First, it does not consider the various environmental conditions in actual synthesis, such as temperature, humidity, and pressure.
Second, the model primarily handles organic molecules and small molecules, but is not capable of dealing with polymers or crystals used in the development of new materials.
Lastly, the current reasoning process remains at the level of reaction mechanisms and does not extend to more sophisticated reasoning, such as electron transfer.
If these limitations are addressed, the model could evolve into a more advanced chemical agent system that would be valuable for organic synthesis research and could be effectively applied in an industrial environment.
% Additional discussions can be found in~\Cref{appendix:sec:discussion}.
\section*{Ethical Considerations}
\label{sec:ethical_considerations}
Retrosynthesis prediction with RetroReasoner can suggest feasible organic synthesis processes for a given target molecule.
However, the model may also generate predictions involving hazardous or toxic compounds.
Therefore, RetroReasoner is intended for research use at the current stage, and any future practical deployment should be accompanied by expert review and appropriate safety assessment.
Another ethical consideration is the computational cost of generating data with SyntheticRetro and training LLMs on the generated data.
Such costs may contribute to increased carbon emissions.
This concern could become more pronounced in future development if models larger than the current 8B-scale model or larger training datasets are used.
% \section*{Acknowledgements}
% LG AI Research supported this work. This work was supported by the National Research Foundation of Korea(NRF) grant funded by the Korea government(MSIT) (No. RS-2024-00410082, 33\%), and the Institute of Information \& communications Technology Planning \& Evaluation(IITP) grant funded by the Korea government(MSIT) (No. 2022-0-00612, Geometric and Physical Commonsense Reasoning based Behavior Intelligence for Embodied AI, 33\%), and (IITP-2026-RS-2025-02304828, artificial intelligence star fellowship support program to nurture the best talents, 34\%).
\section*{Acknowledgements}

This work was supported by the National Research Foundation of Korea (NRF) grant funded by the Korean government (MSIT) in 2026 (Grant No. RS-2026-25613293, Project Title: Physics- and Mathematics-Informed Causal AI Models for Robust Analysis and Prediction under Geometry, Boundary-Condition, Parameter, and Multi-Environment Shifts, 20\%) and (Grant No. RS-2026-25473475, Project Title: Functional Analytic Approach to Foundational Theory of Stochastic PDE based Probabilistic Generative Models, 20\%), and was supported by the Institute of Information \& Communications Technology Planning \& Evaluation (IITP) grant funded by the Korean government (MSIT) (IITP-2026-RS-2025-02304828, artificial intelligence star fellowship support program to nurture the best talents, 20\%), (No. RS-2026-25507543, Development of AI Co-Scientist based on Scientific Causal World Model, 20\%), and (No. RS-2022-II220612, Project Title: Geometric and Physical Commonsense Reasoning based Behavior Intelligence for Embodied AI, 20\%), and was also supported by LG AI Research.
% \input{contents/}
% \input{contents/}

% \input{contents/07_limitation}

% \input{contents/97_acknowledgements}

% \bibliography{custom,anthology-overleaf-1,anthology-overleaf-2}

\bibliography{main}

\appendix
\crefalias{section}{appendix}
\crefalias{subsection}{appendix}
\crefalias{subsubsection}{appendix}
\section{Related Works}
\label{sec:related_works}

\begin{table*}[!t]
  \begin{center}
    \begin{small}
        \resizebox{\linewidth}{!}{%
\begin{tabular}{@{}cccccccc@{}}
\toprule
Model Name &
  Data Source &
  Model Size &
  Public Release &
  Training Objective &
  Reward Function &
  Task Scope &
  Supporting Information \\ \midrule
\begin{tabular}[c]{@{}c@{}}Chem-R\\ \cite{wang2025chem}\end{tabular} &
  \begin{tabular}[c]{@{}c@{}}ChemLLMBench\\ (USPTO 50k, mixed)\end{tabular} &
  Llama-3.1-8B &
  O &
  \begin{tabular}[c]{@{}c@{}}GRPO (Task-balanced\\ sampling)\end{tabular} &
  Answer accuracy &
  \begin{tabular}[c]{@{}c@{}}Forward prediction,\\ Retrosynthesis,\\ Reagent selection,\\ Yield prediction\end{tabular} &
  \begin{tabular}[c]{@{}c@{}}SMILES,\\ Functional group\end{tabular} \\ \midrule
\begin{tabular}[c]{@{}c@{}}ChemDFM\\ \cite{zhao2024chemdfm}\end{tabular} &
  \begin{tabular}[c]{@{}c@{}}ChemLLMBench\\ (USPTO 50k, mixed)\end{tabular} &
  Llama3-13B &
  O &
  \begin{tabular}[c]{@{}c@{}}Domain pretraining,\\ Instruction tuning\end{tabular} &
  - &
  \begin{tabular}[c]{@{}c@{}}Forward prediction,\\ Retrosynthesis,\\ Reagent selection,\\ Yield prediction\end{tabular} &
  \begin{tabular}[c]{@{}c@{}}SMILES,\\ Functional group\end{tabular} \\ \midrule
\begin{tabular}[c]{@{}c@{}}ether0\\ \cite{narayanan2025training}\end{tabular} &
  25 of USPTO 50k &
  Mistral-Small-24B &
  O &
  GRPO &
  \begin{tabular}[c]{@{}c@{}}Answer format,\\ Answer accuracy\end{tabular} &
  \begin{tabular}[c]{@{}c@{}}Forward prediction,\\ Retrosynthesis\end{tabular} &
  \begin{tabular}[c]{@{}c@{}}SMILES,\\ Functional group\end{tabular} \\ \midrule
\begin{tabular}[c]{@{}c@{}}Retro-Expert\\ \cite{li2025retro}\end{tabular} &
  USPTO-50k(5k) &
  Qwen2.5-7B &
  X &
  GRPO &
  \begin{tabular}[c]{@{}c@{}}Answer format,\\ Answer accuracy,\\ Stage\end{tabular} &
  Retrosynthesis &
  \begin{tabular}[c]{@{}c@{}}Reaction type,\\ Reaction center,\\ External knowledge\end{tabular} \\ \midrule
\begin{tabular}[c]{@{}c@{}}RetroDFM-R\\ \cite{zhang2025reasoning}\end{tabular} &
  \begin{tabular}[c]{@{}c@{}}USPTO-50k(5k),\\ USPTO-FULL (100k of 1M)\end{tabular} &
  Llama3-8B &
  O &
  DAPO &
  \begin{tabular}[c]{@{}c@{}}Answer format,\\ Answer accuracy\end{tabular} &
  Retrosynthesis &
  \begin{tabular}[c]{@{}c@{}}SMILES,\\ IUPAC name\end{tabular} \\ \midrule
\begin{tabular}[c]{@{}c@{}}ChemDFM-R\\ 
\cite{zhao2025chemdfm-r}\end{tabular} &
  \begin{tabular}[c]{@{}c@{}}USPTO-50k(5k),\\ USPTO-FULL (100k of 1M),\\ USPTO-MIT (40k) \end{tabular} &
  Qwen2.5-14B &
  O &
  DAPO &
  \begin{tabular}[c]{@{}c@{}}Answer format,\\ Answer accuracy\end{tabular} &
  Retrosynthesis &
  \begin{tabular}[c]{@{}c@{}}SMILES,\\ Functional group \end{tabular} \\ \midrule
\textbf{\begin{tabular}[c]{@{}c@{}}RetroReasoner\\ (Ours)\end{tabular}} &
  \textbf{ORDerly} &
  \textbf{Qwen3-8B} &
  \textbf{\begin{tabular}[c]{@{}c@{}}O\\ \end{tabular}} &
  \textbf{GRPO} &
  \textbf{Round-trip accuracy} &
  \textbf{Retrosynthesis} &
  \textbf{\begin{tabular}[c]{@{}c@{}}SMILES,\\ Functional group,\\ Reaction template,\\ Reaction center,\\ Synthon,\\ Synthetic equivalent\end{tabular}} \\ \bottomrule
\end{tabular}%
        }
    \end{small}
  \end{center}
  \caption{Comparison of \emph{Molecular Reasoning LLMs}. The Supporting Information column lists inputs used to generate reasoning. Chem-R, ChemDFM, ether0, RetroDFM-R, and ChemDFM-R rely mainly on SMILES, functional groups, and IUPAC names, so their rationales stay at generic product analysis. Retro-Expert adds reaction type and reaction center, but does not derive synthons or synthetic equivalents from the disconnection. Most models use reactant correctness as the reward, which can bias learning and does not reflect feasibility, since even reactants that can genuinely synthesize the target product may receive no positive reward if they do not match the single reference reactant. RetroReasoner instead uses a forward model based round-trip reward to better target feasible reactants.}
  \label{tab:molecular_reasoning_LLMs_comparison}
  \vskip -0.1in
\end{table*}

\subsection{Disconnection-aware Retrosynthesis Before LLMs}
Prior to LLM-based approaches, semi-template retrosynthesis models that explicitly leverage reaction centers and intermediate structures have been proposed to mimic chemists' retrosynthetic strategies.
For example, GLN~\cite{dai2019retrosynthesis} combines reaction templates with graph neural networks to learn whether a specific transformation rule can be applied to a given product, complementing the chemical validity of template-based retrosynthesis with neural networks.
RetroXpert~\cite{yan2020retroxpert} proposes a two-stage architecture that first identifies potential reaction centers in the product, then generates intermediate synthons based on these centers, and finally predicts the reactants.
Similarly, G2G~\cite{shi2020graph} formulates retrosynthesis as a graph-to-graphs translation problem by decomposing the product graph into multiple synthons based on reaction centers, and then converting each synthon into a reactant graph.
These studies demonstrate the importance of disconnection-aware intermediate representations in retrosynthesis prediction.
However, in existing semi-template or graph-based models, the intermediate states are typically used only as prediction outputs of specific modules and are not trained as human-readable natural language reasoning trajectories.
RetroReasoner is distinguished from these approaches in that it follows such chemist-like decomposition processes while transforming product analysis, candidate substructure identification, strategic bond disconnection, synthon generation, and synthetic-equivalent mapping into a single structured reasoning trace, which then serves as a direct training target for both supervised fine-tuning and verifier-based reinforcement learning.
Specifically, the SyntheticRetro component of RetroReasoner is designed to generate reasoning text composed of $\mathcal{R}_1$ product analysis, $\mathcal{R}_2$ candidate substructure, $\mathcal{R}_3$ strategic bond disconnection, and $\mathcal{R}_4$ synthetic equivalent mapping, along with linking texts $\mathcal{L}_{12}$, $\mathcal{L}_{23}$, and $\mathcal{L}_{34}$ that connect these reasoning stages.

\subsection{Reasoning LLMs for Chemical Reaction Prediction}
Recently, as LLMs have been applied to chemistry and synthesis problems, research has emerged that goes beyond simple SMILES-to-SMILES translation~\cite{fang2023mol,yu2024llasmol,cao2024presto,pei2024biot5+,lee2025mol} to generate natural language reasoning or to improve chemical decision-making through reinforcement learning.
Chem-R~\cite{wang2025chem} is proposed as a general chemical reasoning model that combines chemical foundation training, reasoning protocol distillation, and multi-task GRPO to learn chemist-like reasoning abilities across various molecular and reaction-level tasks, including forward prediction, retrosynthesis, reagent selection, and yield prediction.
ChemDFM-R~\cite{zhao2025chemdfm-r} enhances chemical reasoning capabilities by leveraging atomized chemical knowledge and domain-specific reinforcement learning, while ether0~\cite{narayanan2025training} is a 24B-scale reasoning model post-trained to perform natural language reasoning before outputting chemical structures.
More directly related to RetroReasoner are RetroDFM-R~\cite{zhang2025reasoning} and Retro-Expert~\cite{li2025retro}.
RetroDFM-R is a reasoning-based LLM specialized for retrosynthesis that improves prediction accuracy and explainability through reinforcement learning with chemically verifiable rewards.
Retro-Expert is a collaborative reasoning framework in which a specialized retrosynthesis model constructs the chemical decision space, an LLM performs critical reasoning, and reinforcement learning optimizes an interpretable decision policy.
However, compared to these works, RetroReasoner aligns its reasoning more closely with the logical structure intrinsic to retrosynthesis.
While existing chemical reasoning LLMs generally provide explanations grounded in product functional groups, reaction types, and general chemical knowledge, RetroReasoner explicitly incorporates into its reasoning trace which bonds should be disconnected, which synthons are formed after disconnection, and which synthetic equivalents each synthon should be mapped to.
Thus, the key distinction of RetroReasoner lies not merely in ``generating reasoning,'' but in bridging the logical gap between generic product-level analysis and reactant selection through disconnection-centric reasoning.
The differences from these related works are summarized in~\Cref{tab:molecular_reasoning_LLMs_comparison}.

\subsection{Round-trip Rewards and Multi-step Planning}
In retrosynthesis, multiple reactant combinations are possible for a single product, so using exact match with a single reference as the sole reward signal may unnecessarily penalize chemically valid alternative pathways.
To address this limitation, prior work has employed round-trip accuracy, which verifies whether predicted reactants can reconstruct the target product via a forward prediction model, as an evaluation metric for single-step retrosynthesis~\cite{schwaller2020predicting}.
RetroReasoner extends this concept beyond evaluation by incorporating it as a reward signal during the RL stage.
Specifically, a reward is granted when the predicted reactants successfully reconstruct the original product through the forward model, enabling the model to learn feasible reactant proposals even when they differ from the reference.
In terms of multi-step planning, recent studies have explored directly generating entire routes or coupling LLMs with search algorithms.
DirectMultiStep~\cite{shee2025directmultistep} proposes generating multi-step synthetic routes directly as a single string, AOT*~\cite{song2025aot} integrates LLM-generated synthetic pathways with AND-OR tree search, and Retro-R1~\cite{liu2026retro} trains a multi-step retrosynthesis agent via reinforcement learning.
RetroInText~\cite{kang2025retrointext} combines a MolT5~\cite{edwards2022translation}-based single-step model with MCTS and a fusion-based value function for retrosynthetic planning.
In contrast, RetroReasoner does not generate entire routes end-to-end; rather, it adopts a hybrid approach that first strengthens a disconnection-centric single-step proposal model and then integrates it into a parallelized MCTS.
Accordingly, RetroReasoner is better positioned not as a multi-step planner itself, but as a strategic single-step reasoner that can be plugged into a search framework.

% 역합성에서는 하나의 생성물에 대해 여러 반응물 조합이 가능하므로, 단일 reference와의 exact match만을 보상으로 사용하면 화학적으로 가능한 대안 경로가 불필요하게 penalize될 수 있다. 이를 보완하기 위해 기존 연구에서도 forward prediction model을 통해 예측 반응물이 목표 생성물을 재구성할 수 있는지를 확인하는 round-trip accuracy가 single-step retrosynthesis 평가 지표로 사용되어 왔다. RetroReasoner는 이 개념을 평가 지표에만 머무르게 하지 않고, RL 단계의 reward signal로 사용한다. 즉, 예측된 반응물이 forward model을 통해 원래 생성물을 재구성하면 보상을 부여함으로써, reference와 다르더라도 feasible한 reactant proposal을 학습할 수 있도록 한다.
% Multi-step planning 측면에서는 최근 route 전체를 직접 생성하거나 LLM을 search에 결합하는 방향의 연구들이 등장하고 있다. DirectMultiStep은 multi-step synthetic route를 하나의 문자열로 직접 생성하는 방식을 제안했고, AOT*는 LLM이 생성한 합성 경로를 AND-OR tree search와 결합하며, Retro-R1은 reinforcement learning으로 multi-step retrosynthesis agent를 학습한다. 이들과 달리 RetroReasoner는 route 전체를 end-to-end로 생성하는 것이 아니라, disconnection-centric single-step proposal model을 먼저 강화한 뒤 이를 parallelized MCTS에 연결하는 hybrid approach에 가깝다. 따라서 RetroReasoner는 multi-step planner 자체라기보다, search framework에 삽입 가능한 전략적 single-step reasoner로 위치시키는 것이 적절하다.

\section{Details of SyntheticRetro}
\label{appendix:sec:details_of_SyntheticRetro}
This section describes SyntheticRetro in detail using an illustrative example.

\subsection{Supporting Information}
\label{appendix:subsec:supporting_information}
Supporting information refers to the data required to construct reasoning text. This information is used in both structured reasoning steps and the generation of linking text.

\subsubsection{Direct-usable Information}
Direct-usable information is obtained directly from an RXN SMILES string.
For example, given the RXN SMILES \texttt{O=C(C1CC1)C(c1ccccc1F)N1CCC\\(S)CC1.CCCC(=O)Cl>>CCCC(=O)SC1CCN(C(C(=O)\\C2CC2)c2ccccc2F)CC1}, the reactants are \texttt{O=C(C1\\CC1)C(c1ccccc1F)N1CCC(S)CC1} and \texttt{CCCC(=O)\\Cl}, and the product is \texttt{CCCC(=O)SC1CCN(C(C(=O)\\C2CC2)c2ccccc2F)CC1}.
These SMILES strings are treated as direct-usable information.

\subsubsection{Model-predicted Information}
Model-predicted information is obtained from deep-learning-based atom-mapping models.
LocalMapper~\cite{chen2024precise} is a sequence-to-sequence model that takes an RXN SMILES string as input and outputs an atom-mapped RXN SMILES string.
For the example above, LocalMapper produces
\begin{lstlisting}
[O:13]=[C:12]([CH:14]1[CH2:15][CH2:16]1)[CH:11]([c:17]1[cH:18][cH:19][cH:20][cH:21][c:22]1[F:23])[N:10]1[CH2:9][CH2:8][CH:7]([SH:6])[CH2:25][CH2:24]1.[CH3:1][CH2:2][CH2:3][C:4](=[O:5])Cl>>[CH3:1][CH2:2][CH2:3][C:4](=[O:5])[S:6][CH:7]1[CH2:8][CH2:9][N:10]([CH:11]([C:12](=[O:13])[CH:14]2[CH2:15][CH2:16]2)[c:17]2[cH:18][cH:19][cH:20][cH:21][c:22]2[F:23])[CH2:24][CH2:25]1.
\end{lstlisting}
This representation encodes atom-level changes between reactants and products.

\subsubsection{Rule-derived Information}
From the atom-mapped SMILES, additional information is extracted using rule-based algorithms.
A reaction template is obtained by identifying the minimum bond changes between reactants and products.
In the example above, sulfur forms a new bond and chlorine is removed from the reactant.
This change is represented as \texttt{([S:6]).([C:4](=O)-Cl)>>([C:4](=O)-[S:6])}.
This instance-specific template is then canonicalized as \texttt{[S:2].[C:1](=O)-Cl>>[C:1](=O)-[S:2]}.
Functional group names and positions, as well as substructures, are extracted using SMARTS~\cite{smarts} patterns.
Functional groups defined in RDKit~\cite{landrum2013rdkit} and substructures used in MACCS keys~\cite{durant2002reoptimization} fingerprints are employed.
The newly formed bond is identified by comparing bond sets between reactants and products and is treated as the strategic bond disconnection.
Synthons are then generated by disconnecting this bond in the product, resulting in two molecular fragments.
For example, when the disconnected bond is a single bond between atoms 4 and 6, the resulting synthons are \texttt{[6*]C(=O)CCC} and \texttt{[4*]SC1CCN(C(C(=O)C2CC2)c2ccccc2F)CC1}.
Here, \texttt{[n*]} denotes a placeholder corresponding to the original atom index.
Finally, each synthon is mapped to its synthetic equivalent using atom mapping information. In this example, the synthon \texttt{[6*]C(=O)CCC} corresponds to the reactant \texttt{CCCC(=O)Cl}, while the synthon \texttt{[4*]SC1CCN(C(C(=O)C2CC2)c2ccccc2F)CC1} corresponds to the reactant \texttt{O=C(C1CC1)C(c1ccc\\cc1F)N1CCC(S)CC1}.

\subsection{Asynchronous Data Generation}
\label{appendix:subsec:asynchronous_data_generation}
\begin{figure*}[!t]
  \vskip 0.2in
  \begin{center}
    % \centerline{
    \includegraphics[width=0.90\linewidth]{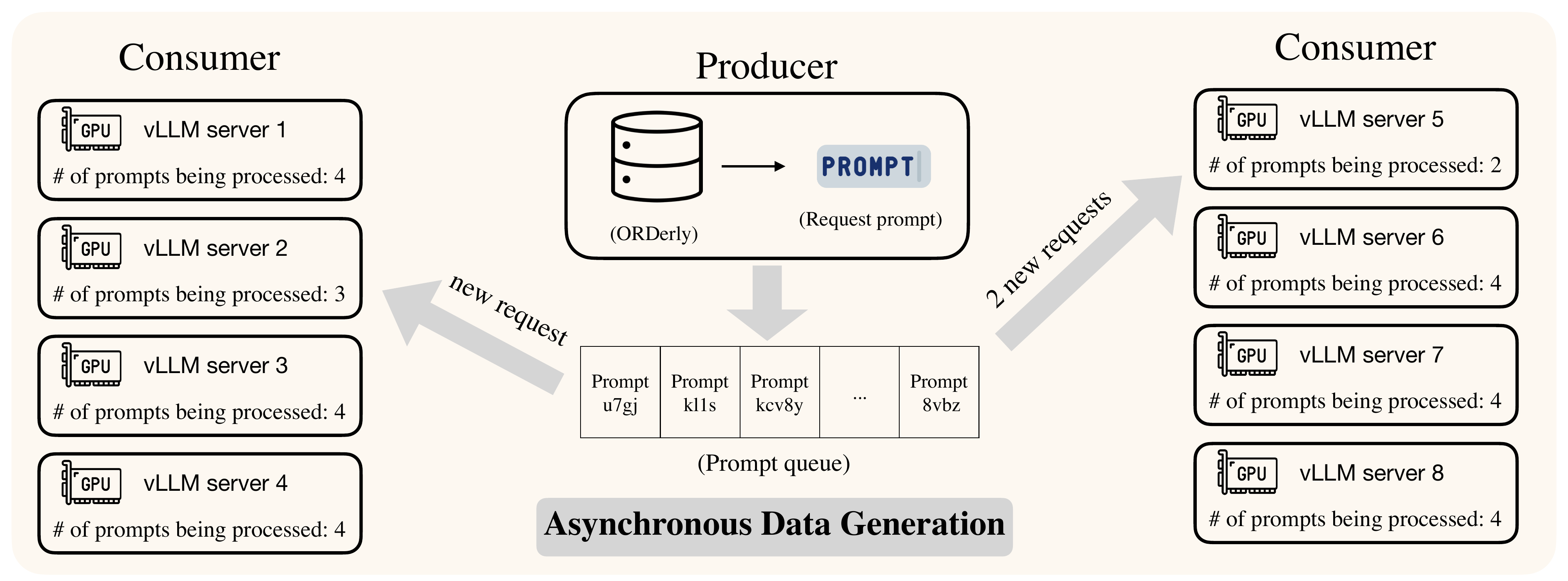}
    \caption{
      Schematic of the asynchronous data generation system in SyntheticRetro. A producer constructs prompts for linking text generation from ORDerly RXN SMILES and pushes them into a prompt queue, which is then processed asynchronously by multiple vLLM servers. Each vLLM server has a fixed maximum number of concurrent requests, and generation proceeds by filling these slots asynchronously.
    }
    \label{fig:asynchronous_data_generation}
  \end{center}
\end{figure*}

\begin{table*}[!t]
% \centering
\begin{minipage}{0.30\linewidth}
  \begin{small}
    \begin{tabular}{@{}ll@{}}
    \toprule
    Parameter         & Value       \\ \midrule
    Model Name        & GPT-oss-20B \\
    Temperature       & 0.8         \\
    Max Tokens        & 500         \\
    Presence Penalty  & 0.0         \\
    Frequency Penalty & 0.3         \\
    Reasoning Effort  & Low         \\
    Max Producers     & 20          \\
    Max Consumers     & 31*20       \\
    Prompt Queue Size & 5000        \\ \bottomrule
    \end{tabular}
  \end{small}
  \caption{Hyperparameters used in SyntheticRetro data generation.}
  \label{tab:hyperparameters_SyntheticRetro}
\end{minipage} \hspace{0.02\linewidth}
\begin{minipage}{0.65\linewidth}
  \begin{small}
    \begin{tabular}{@{}lll@{}}
    \toprule
    Experiment Name & Number of Instances & Number of Tokens        \\ \midrule
    Prediction-Only (SFT)                                                         & 522,630             & 59,439,861 $\times$ 15  \\
    RetroReasoner (SFT)                                                           & 522,630             & 8,050,108,240           \\
    RetroReasoner (SFT, 1 sample)                                                 & 522,630             & 536,799,367 $\times$ 15 \\
    RetroReasoner (SFT, 1.7B)                                                     & 522,630             & 8,050,108,240           \\
    RetroReasoner (SFT, w/o $\mathcal{L}_{12},\mathcal{L}_{23},\mathcal{L}_{34}$) & 522,630             & 319,138,383 $\times$ 15 \\
    RetroReasoner (SFT, w/ only $\mathcal{R}_1$)                                     & 522,630             & 403,749,140 $\times$ 15 \\
    Round-trip Forward Model (0.6B)                                               & 939,001             & 107,580,021 $\times$ 15 \\
    Round-trip Forward Model (8B)                                                 & 939,001             & 107,580,021 $\times$ 15 \\ \bottomrule
    \end{tabular}
  \end{small}
  \caption{Number of instances used to train the model used in each experiment in supervised fine-tuning and total number of tokens based on 15 epochs.}
  \label{tab:dataset_statistics}
\end{minipage}
\vskip -0.1in
\end{table*}
\begin{figure*}[!t]
  \vskip 0.2in
  \begin{center}
    % \centerline{
    \includegraphics[width=0.97\linewidth]{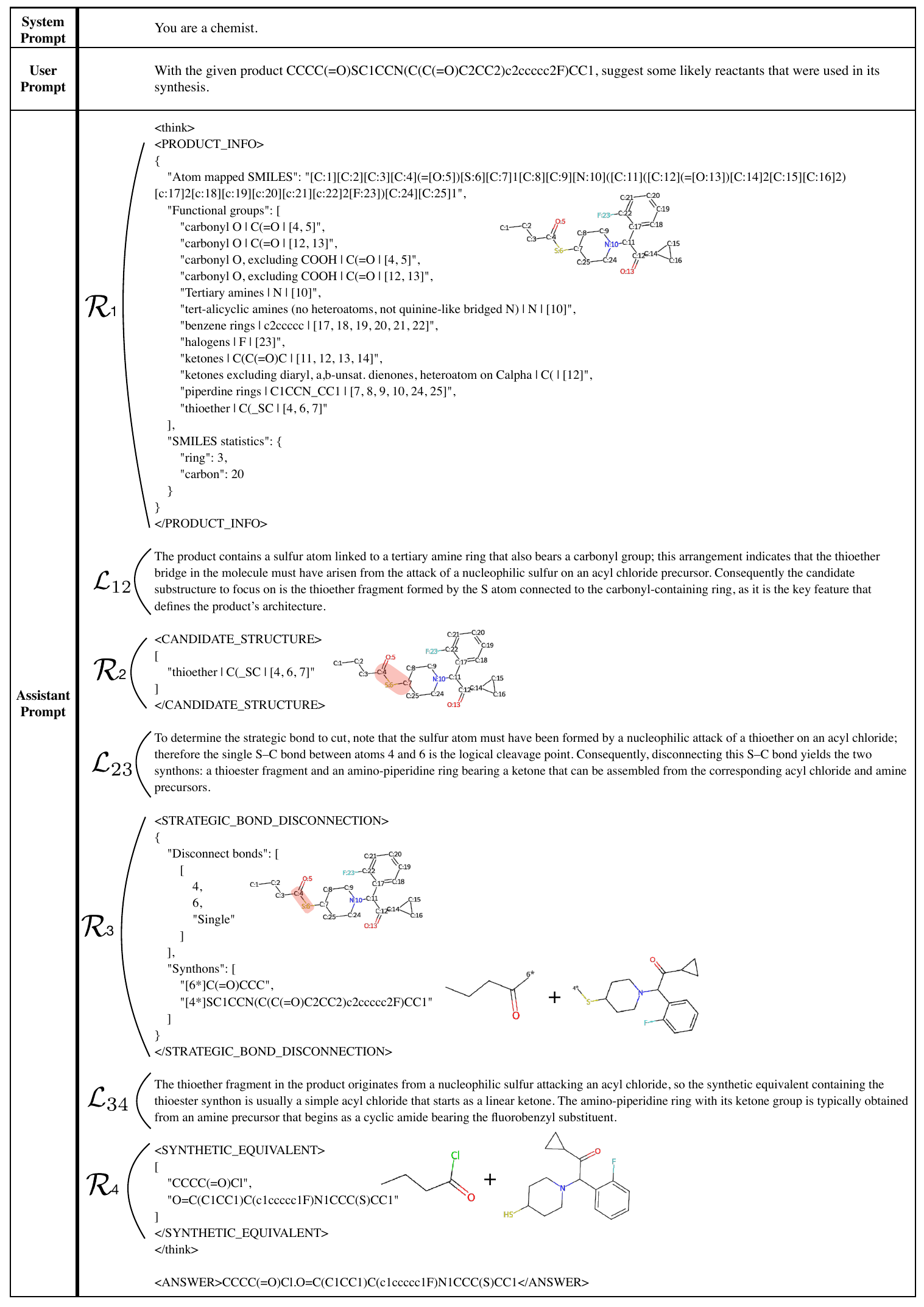}
    \caption{
      Example of a reasoning process generated by SyntheticRetro.
    }
    \label{fig:rationale_example}
  \end{center}
\end{figure*}

SyntheticRetro uses vLLM~\cite{kwon2023efficient} for efficient generation of linking text.
However, the total number of requests is the product of the number of instances, the number of linking texts per instance, and the number of samples per instance, which results in a very large request volume.
To handle this efficiently, SyntheticRetro adopts a producer--consumer pattern for asynchronous data generation.
\Cref{fig:asynchronous_data_generation} illustrates the overall generation system, which uses eight GPUs, each running a single vLLM serving instance.
The producer constructs prompts for linking text generation and pushes them into a global prompt queue.
The consumers, corresponding to the vLLM servers, asynchronously retrieve prompts from the queue and process multiple requests in parallel.
Each consumer is configured to handle up to four concurrent requests, and the maximum number of active requests per consumer is controlled using a semaphore mechanism.

\subsection{Data Filtering}
Instances are excluded when LocalMapper fails to produce an atom-mapped output and returns \texttt{None}.
Instances are also excluded when generation exceeds the maximum allowed token length, which is indicated by a finish reason of length.

\subsection{Generation Hyperparameters}
\label{appendix:subsec:generation_hyperparameters}
Hyperparameters for data generation in SyntheticRetro are shown in~\Cref{tab:hyperparameters_SyntheticRetro}.

\subsection{Rationale Example}
\label{appendix:subsec:rationale_example}
A representative example of the generated data is shown in~\Cref{fig:rationale_example}.

\section{Details of RetroReasoner and Prediction-Only}
\label{appendix:sec:details_of_RetroReasoner}

\subsection{Training Details}
\begin{figure*}[!t]
  \vskip 0.2in
  \begin{center}
    % \centerline{
    \includegraphics[width=0.98\linewidth]{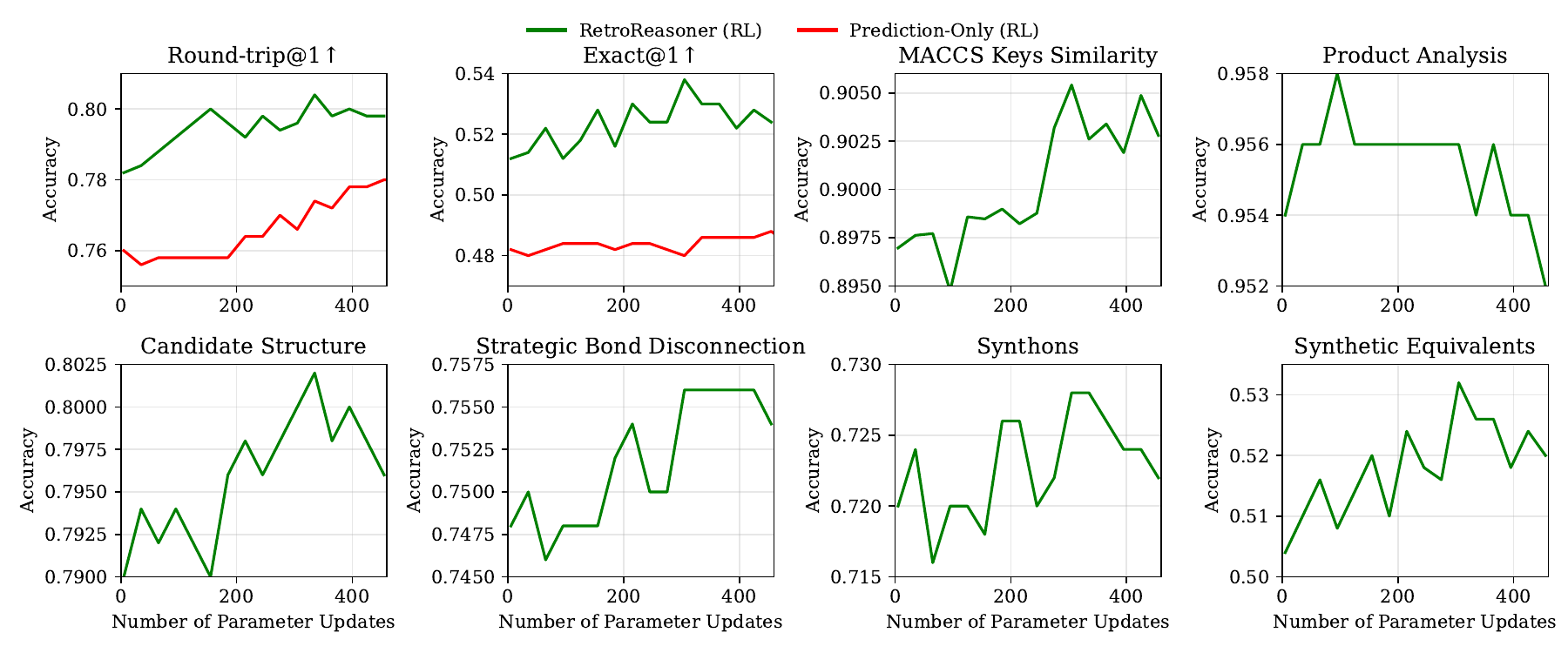}
    \caption{Learning curves for RetroReasoner (RL) and Prediction-Only (RL). The figure shows learning curves for \textsf{Round-trip@1↑} and \textsf{Exact@1↑}, as well as MACCS Keys fingerprint similarity to the reactants and the accuracy of the structured reasoning steps.}
    \label{fig:learning_curve_rl}
  \end{center}
\end{figure*}

Supervised fine-tuning is conducted using distributed data parallelism (DDP) on eight NVIDIA H100 GPUs, with full parameter tuning in BFloat16 precision.
Ten different user prompt templates are used, with only the product SMILES modified for each instance.
A simple system prompt of \texttt{You are a chemist.} is used.
Cross-entropy loss is masked for both system and user prompts, and loss is computed only on the assistant prompt tokens.
The final training loss is calculated as the average loss over assistant prompt tokens in each batch.
Reinforcement learning is also performed using eight NVIDIA H100 GPUs with full parameter tuning in BFloat16 precision.
Three rollout workers are used, with one worker assigned per GPU, and training is conducted with FSDP using a shard size of four across four GPUs.
Reward computation is handled by running a round-trip worker on the remaining GPU.
During reward calculation, the caching system described below is used to improve efficiency.
The learning curve of reinforcement learning is shown in~\Cref{fig:learning_curve_rl}.

\subsection{Hyperparameters}
\begin{table*}[!t]
  \begin{center}
    \begin{small}
        \resizebox{\linewidth}{!}{%
\begin{tabular}{@{}lcccccccc@{}}
\toprule
 &
  \begin{tabular}[c]{@{}c@{}}Base\\ Model\end{tabular} &
  Optimizer &
  \begin{tabular}[c]{@{}c@{}}Learning\\ Rate\end{tabular} &
  \begin{tabular}[c]{@{}c@{}}Gradient\\ Clip\end{tabular} &
  \begin{tabular}[c]{@{}c@{}}Weight\\ Decay\end{tabular} &
  \begin{tabular}[c]{@{}c@{}}Warmup\\ Steps\end{tabular} &
  \begin{tabular}[c]{@{}c@{}}Batch\\ Size\end{tabular} &
  \begin{tabular}[c]{@{}c@{}}Backend\\ Framework\end{tabular} \\ \midrule
Prediction-Only (SFT) &
  Qwen3-8B &
  \begin{tabular}[c]{@{}c@{}}AdamW\\ \cite{loshchilov2017decoupled}\end{tabular} &
  $10^{-5}$ &
  0.5 &
  0.1 &
  100 &
  64 &
  \begin{tabular}[c]{@{}c@{}}PyTorch Lightning\\ \cite{pytorch_lightning},\\ DeepSpeed (ZeRO stage1)\\ \cite{rasley2020deepspeed}\end{tabular} \\
RetroReasoner (SFT) &
  Qwen3-8B &
  AdamW &
  $10^{-5}$ &
  0.5 &
  0.1 &
  100 &
  64 &
  \begin{tabular}[c]{@{}c@{}}PyTorch Lightning,\\ DeepSpeed (ZeRO stage1)\end{tabular} \\
RetroReasoner (SFT, 1 sample) &
  Qwen3-8B &
  AdamW &
  $10^{-5}$ &
  0.5 &
  0.1 &
  100 &
  64 &
  \begin{tabular}[c]{@{}c@{}}PyTorch Lightning,\\ DeepSpeed (ZeRO stage1)\end{tabular} \\
RetroReasoner (SFT, 1.7B) &
  Qwen3-1.7B &
  AdamW &
  $10^{-5}$ &
  0.5 &
  0.1 &
  100 &
  64 &
  \begin{tabular}[c]{@{}c@{}}PyTorch Lightning,\\ DeepSpeed (ZeRO stage1)\end{tabular} \\
RetroReasoner (SFT, w/o $\mathcal{L}_{12},\mathcal{L}_{23},\mathcal{L}_{34}$) &
  Qwen3-8B &
  AdamW &
  $10^{-5}$ &
  0.5 &
  0.1 &
  100 &
  64 &
  \begin{tabular}[c]{@{}c@{}}PyTorch Lightning,\\ DeepSpeed (ZeRO stage1)\end{tabular} \\
RetroReasoner (SFT, w/ only $\mathcal{R}_1$)) &
  Qwen3-8B &
  AdamW &
  $10^{-5}$ &
  0.5 &
  0.1 &
  100 &
  64 &
  \begin{tabular}[c]{@{}c@{}}PyTorch Lightning,\\ DeepSpeed (ZeRO stage1)\end{tabular} \\
Round-trip Forward Model (0.6B) &
  Qwen3-0.6B &
  AdamW &
  $10^{-5}$ &
  0.5 &
  0.1 &
  100 &
  64 &
  \begin{tabular}[c]{@{}c@{}}PyTorch Lightning,\\ DeepSpeed (ZeRO stage1)\end{tabular} \\
Round-trip Forward Model (8B) &
  Qwen3-8B &
  AdamW &
  $10^{-5}$ &
  0.5 &
  0.1 &
  100 &
  64 &
  \begin{tabular}[c]{@{}c@{}}PyTorch Lightning,\\ DeepSpeed (ZeRO stage1)\end{tabular} \\ \bottomrule
\end{tabular}%
        }
    \end{small}
  \end{center}
  \caption{Hyperparameters of the models used in the SFT training experiments.}
  \label{tab:hyperparameters_SFT}
  \vskip -0.1in
\end{table*}
\begin{table*}[!t]
  \begin{center}
    \begin{small}
        \resizebox{\linewidth}{!}{%
\begin{tabular}{@{}lcccccccc@{}}
\toprule
 &
  \begin{tabular}[c]{@{}c@{}}Base\\ Model\end{tabular} &
  Optimizer &
  \begin{tabular}[c]{@{}c@{}}Learning\\ Rate\end{tabular} &
  \begin{tabular}[c]{@{}c@{}}Gradient\\ Clip\end{tabular} &
  \begin{tabular}[c]{@{}c@{}}Weight\\ Decay\end{tabular} &
  \begin{tabular}[c]{@{}c@{}}Warmup\\ Steps\end{tabular} &
  \begin{tabular}[c]{@{}c@{}}KL loss\\ coefficient\end{tabular} &
  \begin{tabular}[c]{@{}c@{}}Entropy\\ Coefficient\end{tabular} \\ \midrule
Prediction-Only (RL) &
  Qwen3-8B &
  AdamW &
  $3 \times 10^{-7}$ &
  1.0 &
  0.1 &
  10 &
  0.05 &
  0.05 \\
RetroReasoner (RL) &
  Qwen3-8B &
  AdamW &
  $3 \times 10^{-7}$ &
  1.0 &
  0.1 &
  10 &
  0.05 &
  0.05 \\
RetroReasoner (RL, w/ $R^{\text{exact}}$) &
  Qwen3-8B &
  AdamW &
  $3 \times 10^{-7}$ &
  1.0 &
  0.1 &
  10 &
  0.05 &
  0.05 \\ \midrule
 &
  \begin{tabular}[c]{@{}c@{}}Clip Ratio\\ (low, high)\end{tabular} &
  \begin{tabular}[c]{@{}c@{}}Group\\ Size\end{tabular} &
  \begin{tabular}[c]{@{}c@{}}Parameter\\ Sync Step\end{tabular} &
  \begin{tabular}[c]{@{}c@{}}Batch\\ Size\end{tabular} &
  \begin{tabular}[c]{@{}c@{}}Backend\\ Framework\end{tabular} &
  \begin{tabular}[c]{@{}c@{}}Rollout\\ Workers\end{tabular} &
  \begin{tabular}[c]{@{}c@{}}Trainer\\ Workers\end{tabular} &
  \begin{tabular}[c]{@{}c@{}}Round-trip\\ Workers\end{tabular} \\ \midrule
Prediction-Only (RL) &
  0.2 &
  16 &
  5 &
  48 &
  \begin{tabular}[c]{@{}c@{}}verl\\ \cite{sheng2024hybridflow},\\ FSDP2\\ \cite{zhao2023pytorch}\end{tabular} &
  3 &
  4 &
  1 \\
RetroReasoner (RL) &
  0.2 &
  16 &
  5 &
  48 &
  \begin{tabular}[c]{@{}c@{}}verl,\\ FSDP2\end{tabular} &
  3 &
  4 &
  1 \\
RetroReasoner (RL, w/ $R^{\text{exact}}$) &
  0.2 &
  16 &
  5 &
  48 &
  \begin{tabular}[c]{@{}c@{}}verl,\\ FSDP2\end{tabular} &
  3 &
  4 &
  1 \\ \bottomrule
\end{tabular}%
        }
    \end{small}
  \end{center}
  \caption{Hyperparameters of the models used in the RL training experiments.}
  \label{tab:hyperparameters_RL}
  \vskip -0.1in
\end{table*}
All hyperparameters of SFT and RL are shown in~\Cref{tab:hyperparameters_SFT,tab:hyperparameters_RL}.

\subsection{Cached Round-trip Verifier}
Computing the round-trip reward requires running the round-trip model $f_{\phi}$ to predict a product for each sampled reactant set $\hat{\mathbf{y}}^{\text{reactant}}$.
Since RL training generates many samples and often revisits the same reactant SMILES strings across workers, repeatedly evaluating $f_{\phi}$ on identical inputs becomes a major computational bottleneck.
To reduce redundant computation, a shared cache is used for round-trip verification.
The cache is a key value store that maps a canonicalized reactant SMILES to the corresponding product prediction from $f_{\phi}$.
During training, each rollout worker first queries the cache with $\hat{\mathbf{y}}^{\text{reactant}}$.
If a cached entry exists, the stored product prediction is reused to compute $R^{\text{round-trip}}$.
Otherwise, $f_{\phi}(\hat{\mathbf{y}}^{\text{reactant}})$ is computed once, stored in the cache, and then used for reward computation.

\section{Details of Round-trip Prediction Model}
\label{appendix:sec:details_of_roundtrip}

\subsection{Training Details}
The round-trip model $f_\phi$ follows the same learning process as the Prediction-Only model, with the only difference being the input and output SMILES.
Unlike Prediction-Only, the input consists of reactants and the output is the product.
During training, the round-trip model is trained in the same manner as RetroReasoner and Prediction-Only, using cross-entropy for next token prediction.
% Since the round-trip model is used for evaluation and reward calculation, there is no strict need to divide the data into train, validation, and test sets.
% Therefore, the train and test data split from ORDerly are combined, and 300 random samples are selected as the validation set.
For evaluation, the 8B model is trained and used for higher accuracy, while the 0.6B model is trained for reward calculation to ensure faster computation.
The instances used for evaluation are excluded from the round-trip model training.

\subsection{Hyperparameters}
The hyperparameters for the 8B and 0.6B models in round-trip training are shown in~\Cref{tab:hyperparameters_SFT}.

\subsection{Learning Curve}
\begin{figure*}[!t]
  \vskip 0.2in
  \begin{center}
    \includegraphics[width=0.75\linewidth]{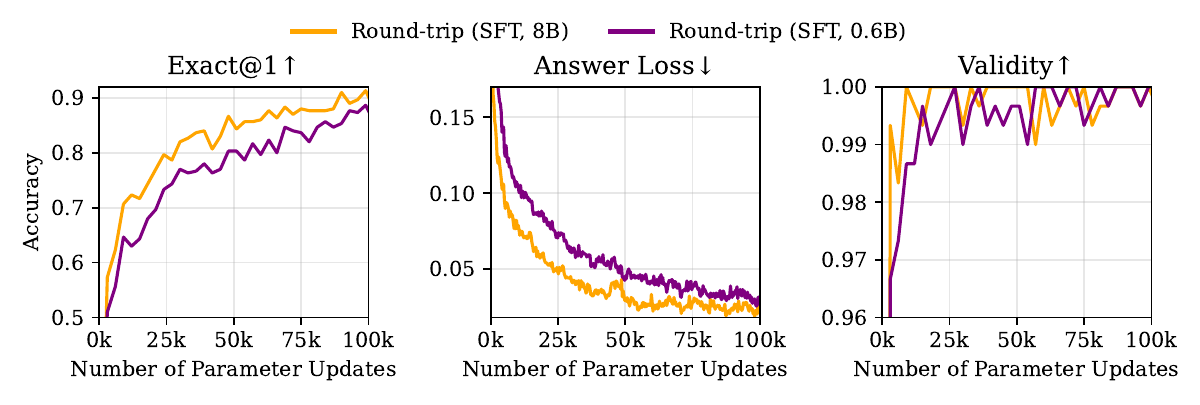}
    \caption{Learning curves for training the round-trip model. The figure shows \textsf{Exact@1↑}, the teacher forcing loss computed on the product SMILES tokens, which is reported as \textsf{Answer Loss↓}, and the validity of the generated product SMILES (\textsf{Validity↑}).}
    \label{fig:learning_curve_roundtrip}
  \end{center}
\end{figure*}

The learning curves for the 8B and 0.6B models in round-trip training are shown in~\Cref{fig:learning_curve_roundtrip}. In~\Cref{fig:learning_curve_roundtrip}, the Exact@1 metric for the round-trip model indicates product prediction accuracy, in contrast to reactant prediction accuracy for RetroReasoner or Prediction-Only.
The answer loss represents the average teacher forcing loss computed for the answer tokens, i.e., the product SMILES tokens.
Validity indicates whether the predicted product molecule is chemically valid.
From the learning curve, it is evident that the 8B model, used for evaluation, shows higher accuracy, achieving a validation accuracy close to 0.91, confirming its suitability for evaluation tasks.
\section{Details of Monte Carlo Tree Search for Retrosynthesis}
\label{appendix:sec:details_of_mcts}

\subsection{Problem Formulation}
Given a target molecule $m_{\mathrm{target}}$, the objective is to find a synthesis route that decomposes $m_{\mathrm{target}}$ into commercially available stock molecules in $\mathcal{B}$.
At each step, the template-free single-step retrosynthesis proposal model $\pi_\theta$ takes a product molecule $m$ as input and samples $k$ candidate reactant sequences.
Each valid sequence is parsed into a reactant set $R_j = \{m_{j,1}, m_{j,2}, \ldots, m_{j,|R_j|}\}$, which corresponds to one retrosynthetic reaction.
A synthesis route is considered solved when all of its leaf molecules belong to $\mathcal{B}$.

\subsection{Search Tree Structure}
The retrosynthesis search tree is represented as an AND-OR tree.
A molecule is represented as an OR node, and a retrosynthetic reaction is represented as an AND node.
A molecule node is denoted by $m$, and a reaction node is denoted by $r$.
The function $\mathrm{child}(\cdot)$ denotes the child nodes of a given node.
Thus, $\mathrm{child}(m)$ contains reaction nodes, while $\mathrm{child}(r)$ contains molecule nodes.
A molecule node is solved if it is already included in the stock set or if at least one of its child reaction nodes is solved.
A reaction node is solved only when all of its child molecule nodes are solved.
If a non-stock molecule node has no child reaction node, it is treated as unsolved.

\subsection{Synthesis Cost Estimation}
To estimate the synthesis difficulty of each molecule, the value model proposed in Retro*~\cite{chen2020retro} is adopted.
To avoid notational conflict with the MCTS node value $V$, the synthesis cost model is denoted by $C$.
The model takes a 2048-bit Morgan fingerprint with radius 2 as input and outputs a non-negative synthesis cost through a two-layer MLP with a softplus output activation.
A larger $C(m)$ indicates that molecule $m$ is more difficult to synthesize, and $C(m)=0$ is assigned for stock molecules.

\subsection{MCTS Algorithm for Retrosynthesis}
The AND-OR tree is explored using Monte Carlo Tree Search (MCTS).
Each node maintains a visit count $N$ and a cumulative search value $V$.
For a reaction node $r$, these statistics are denoted by $N_r$ and $V_r$.
Each MCTS iteration consists of four stages: selection, expansion, evaluation, and backpropagation.
The procedure is repeated until the root molecule node is solved or the maximum number of iterations $T$ is reached.

\paragraph{Selection}
Starting from the root molecule node, the tree is traversed until an unexpanded or terminal molecule node is reached.
For the transition from a molecule node to one of its child reaction nodes, the reaction node is selected according to the Upper Confidence Bound (UCB) score.
In the WU-UCT setting, the score is computed as
\begin{equation}
\text{UCB}(r)_\text{WU} =
\frac{V_r}{N_r}
+
c \cdot p_r \cdot
\frac{\sqrt{N_{\text{parent}} + O_{\text{parent}}}}
{1 + N_r + O_r}.
\end{equation}
Here, $V_r$ and $N_r$ denote the cumulative value and visit count of reaction node $r$, $p_r$ denotes its prior probability, and $c=1.4$ denotes the exploration constant.
The term $N_{\text{parent}}$ is the visit count of the parent molecule node.
The unobserved counts $O_r$ and $O_{\text{parent}}$ track in-flight simulations assigned to the reaction node and its parent molecule node, respectively.
For unvisited reaction nodes with $N_r=0$, the node is selected before applying the UCB score to avoid division by zero.
After a reaction node is selected, the transition from the reaction node to a molecule node is made by selecting an unsolved child molecule with the smallest visit count.
This encourages balanced exploration among the unsolved reactants of the selected reaction.

\paragraph{Expansion}
For a leaf molecule node $m_\ell$ reached during selection, the proposal model $\pi_\theta$ is queried to sample $k$ candidate reactant sequences.
Each valid sequence is parsed into a reactant set $R_j$, and a corresponding reaction node $r_j$ is added as a child of $m_\ell$.
The reactant molecules in $R_j$ are then added as child molecule nodes of $r_j$.
The prior probability $p_{r_j}$ of each reaction node is computed from the fraction of its reactants that already belong to the stock set, with Laplace smoothing:
\begin{equation}
p_{r_j}
=
\frac{
n_{\mathrm{stock}}(R_j) + 1
}{
|R_j| + 1
},
\end{equation}
where $n_{\mathrm{stock}}(R_j)$ is the number of reactants in $R_j$ that belong to $\mathcal{B}$.
This prior assigns a higher score to reactions with more immediately purchasable reactants while ensuring that every valid reaction has a positive prior.

\paragraph{Evaluation}
A value $v(m_\ell) \in [0,1]$ is estimated for the expanded molecule node $m_\ell$.
The evaluation combines two signals: the highest reaction prior among the newly expanded reactions and the predicted synthesis cost of the remaining non-stock molecules.
The non-stock child molecules generated by the newly expanded reactions are collected, and their average synthesis cost is computed using $C$.
If all generated child molecules are stock molecules, the average synthesis cost is set to zero.
The evaluation value is then defined as the larger value between the maximum reaction prior and the inverse-cost score $1/(1+\bar{C}(m_\ell))$, where $\bar{C}(m_\ell)$ denotes the average synthesis cost of the non-stock child molecules.
This optimistic estimate gives a high value when either the immediate stock coverage is high or the predicted synthesis cost of the remaining molecules is low.

\paragraph{Backpropagation}
After evaluation, all nodes along the selected path are updated in reverse order.
For each visited node, the visit count is increased by one.
The cumulative value is increased by $1$ if the node is solved, and otherwise increased by the evaluation value $v(m_\ell)$.
Assigning a value of $1$ to solved nodes prioritizes confirmed synthesis routes in subsequent iterations.

\subsection{Parallelization via WU-UCT}
Calls to the proposal model $\pi_\theta$ are expensive and constitute the main bottleneck of sequential MCTS.
To improve efficiency, parallel MCTS based on WU-UCT is used.
Each node additionally maintains an unobserved count $O$, which records the number of in-flight simulations currently assigned to the node but not yet completed.
The unobserved count is used during selection to discourage multiple workers from choosing the same path.
When a path has already been assigned to a worker, its unobserved count increases, which lowers its exploration priority in subsequent selections.
This reduces redundant expansion while preserving the value estimate based only on completed simulations.
The master thread performs selection and backpropagation sequentially, while worker processes perform expansion calls to $\pi_\theta$ in parallel.
When a selected path is dispatched to a worker, the unobserved count of every node on the path is increased.
When the worker returns the expansion result, the corresponding unobserved counts are decreased, and the visit counts and cumulative values are updated using the evaluated value.
This enables multiple proposal-model calls to run in parallel without repeatedly expanding the same high-priority path.
\section{Experiment Details}
\label{appendix:sec:experiment_details}

\subsection{Metric Calculation Details}
\label{appendix:subsec:metric_calculation_details}
In this section, the calculation methods for the metrics used in the main text are described in detail, along with the corresponding mathematical formulas.
Let the input product be denoted as $\mathbf{x}$, the ground-truth reactant as $\mathbf{y}$, the top-1 greedy predicted reactant as $\hat{\mathbf{y}}^{(GR)}$, and the $K$ sampling predicted reactants as $\{\hat{\mathbf{y}}^{(1)}, \hat{\mathbf{y}}^{(2)}, \dots, \hat{\mathbf{y}}^{(K)}\}$.
Throughout this section, \(\mathbb{I}(a,b)\) denotes molecular identity based on standard InChI matching, as defined in~\Cref{equation:round-trip}.
For an evaluation dataset consisting of $N$ instances $\{(\mathbf{x}_i, \mathbf{y}_i)\}_{i=1}^{N}$, each metric is calculated as follows.

\subsubsection{Greedy Metrics}
The \textsf{Exact@1↑} metric is computed by averaging the indicator function across all instances to check if the top-1 greedy predicted reactant exactly matches the ground-truth reactant:
\begin{equation}
\text{\textsf{Exact@1↑}} = \frac{1}{N} \sum_{i=1}^{N} \mathbb{I}\left( \hat{\mathbf{y}}^{(GR)}_i, \mathbf{y}_i \right).
\end{equation}

The \textsf{Round-trip@1↑} metric is calculated by averaging the indicator function to check if the top-1 greedy predicted reactant, when passed through the round-trip model $f_\phi$, reconstructs the original product:
\begin{equation}
\text{\textsf{Round-trip@1↑}} = \frac{1}{N} \sum_{i=1}^{N} \mathbb{I}\left( \mathbf{x}_i, f_\phi(\hat{\mathbf{y}}^{(GR)}_i) \right).
\end{equation}

\subsubsection{Sampling Metrics}
The \textsf{Exact@100↑} metric is calculated by averaging across all instances, where at least one of the $K$ sampled predictions matches the ground-truth reactant:
\begin{equation}
\text{\textsf{Exact@100↑}} = \frac{1}{N} \sum_{i=1}^{N} \max_{1 \leq k \leq K} \mathbb{I}\left( \hat{\mathbf{y}}_i^{(k)}, \mathbf{y}_i \right).
\end{equation}

The \textsf{Round-trip@100↑} metric is computed by averaging across all instances, where at least one of the $K$ sampled predictions, when passed through the round-trip model $f_\phi$, successfully reconstructs the original product:
\begin{equation}
\begin{aligned}
&\text{\textsf{Round-trip@100↑}} = \\
&\frac{1}{N} \sum_{i=1}^{N} \max_{1 \leq k \leq K} \mathbb{I}\left( \mathbf{x}_i, f_\phi(\hat{\mathbf{y}}_i^{(k)}) \right).
\end{aligned}
\end{equation}

The \textsf{Feasible Ratio↑} metric measures the proportion of the $K$ sampled predictions that pass the round-trip validation, averaged across all instances:
\begin{equation}
\begin{aligned}
&\text{\textsf{Feasible Ratio↑}} \\
&= \frac{1}{N} \sum_{i=1}^{N} \left( \frac{1}{K} \sum_{k=1}^{K} \mathbb{I}\left( \mathbf{x}_i, f_\phi(\hat{\mathbf{y}}_i^{(k)}) \right) \right).
\end{aligned}
\end{equation}

The \textsf{Template Diversity↑} metric is calculated by counting the number of distinct reaction templates for the $K$ sampled predictions that successfully pass the round-trip validation, and then averaging this count across all instances:
\begin{equation}
\text{\textsf{Template Diversity↑}} = \frac{1}{N} \sum_{i=1}^{N} |S_i|,
\end{equation}

where 
\begin{equation}
\begin{aligned}
S_i = \biggl\{ T\left(\hat{\mathbf{y}}_i^{(k)}, \mathbf{x}_i\right) : &\mathbb{I}\left( \mathbf{x}_i, f_\phi(\hat{\mathbf{y}}_i^{(k)}) \right) = 1, \\ &1 \leq k \leq K \biggl\},
\end{aligned}
\end{equation}
Here, $T$ extracts a canonical reaction template from a predicted-reactant--target-product pair.

\subsubsection{Multi-step Retrosynthesis Metrics}

Three complementary metrics introduced in PaRoutes~\cite{genheden2022paroutes} are adopted to evaluate the performance of retrosynthesis search.
Let $\mathcal{B}$ denote the purchasable building-block stock, and let $\mathcal{R}_i$ denote the set of complete synthesis routes extracted for the $i$-th target molecule $m^{(i)}_{\text{target}}$ after search.
A route is considered complete if all of its leaf molecules belong to $\mathcal{B}$.
The \textsf{Solved Targets↑} metric is calculated by counting the number of target molecules for which the search finds at least one complete synthesis route:
\begin{equation}
\begin{aligned}
&\text{\textsf{Solved Targets↑}} \\
&= \sum_{i=1}^{N}
\mathbb{I}\left(
\mathrm{solved}(m^{(i)}_{\text{target}}), \mathrm{True}
\right),
\end{aligned}
\end{equation}
where $\mathrm{solved}(m^{(i)}_{\text{target}})$ follows the recursive solved-status definition in~\Cref{appendix:sec:details_of_mcts}, and evaluates to true when the target molecule can be recursively decomposed into purchasable molecules in $\mathcal{B}$.

The \textsf{Routes Extracted↑} metric measures the total number of complete synthesis routes discovered by the search across all targets.
For each target, complete routes are enumerated from the constructed AND-OR tree after search termination:

\begin{equation}
\text{\textsf{Routes Extracted↑}} =
\sum_{i=1}^{N}
|\mathcal{R}_i|,
\end{equation}
where $\mathcal{R}_i$ is the set of complete routes extracted for the $i$-th target molecule.
If no complete route is found for a target, then $|\mathcal{R}_i|=0$.
This metric reflects the overall number of complete synthesis routes discovered by the search.

The \textsf{Number of Clusters↑} metric measures the total number of structurally distinct route clusters discovered across all targets.
For each target molecule with at least three extracted routes ($N_i \geq 3$), routes are clustered based on their pairwise structural distances, which are computed using the LSTM-based route distance model from PaRoutes~\cite{genheden2022paroutes}.
Targets with fewer than three routes are excluded from the cluster-count summation.

The clustering procedure proceeds as follows. 
First, the maximum number of candidate clusters is set to $\lceil N_i / d_{\min} \rceil$, where $d_{\min}=2$ is the minimum cluster density hyperparameter.
Then, the optimal number of clusters $k_i^*$ is selected from the range $[2, \min(K_i^{\max}, N_i - 1)]$ by maximizing the mean Silhouette score over all routes of the target.
The Silhouette coefficient of each route is defined in the standard way as $(b - a) / \max(a, b)$, where $a$ is the mean distance from the route to other routes in the same cluster, and $b$ is the smallest mean distance from the route to routes in any other cluster.
Once the optimal partition is obtained, the number of clusters $n_i$ for the $i$-th target is given by the number of distinct cluster labels.

The final \textsf{Number of Clusters↑} metric is computed by summing the number of clusters over all targets for which clustering is performed:

\begin{equation}
\text{\textsf{Number of Clusters↑}} =
\sum_{i: N_i \geq 3} n_i.
\end{equation}
A higher value indicates that the search discovers a larger number of structurally distinct synthesis strategies across the benchmark.

\subsection{Evaluation Dataset Generation}
\label{appendix:subsec:evaluation_dataset_generation}
\subsubsection{In-distribution Evaluation Dataset}
The in-distribution dataset consists of 500 instances, which are specifically derived from the ORDerly retrosynthesis benchmark test dataset\footnote{https://figshare.com/articles/dataset/ORDerly\_chemical\\\_reactions\_condition\_benchmarks/23298467}.
Instances that overlap with the training datasets from LlaSMol and Mol-Instructions are excluded from the ORDerly benchmark test dataset.
Overlap is determined by checking if both the reactant and product molecules are the same.
Instances are also excluded if LocalMapper fails to make a prediction or if there are multiple reactants for a single product in the entire ORDerly training dataset.
After exclusions, 500 instances are randomly selected from the remaining ones to be used as the in-distribution evaluation dataset.

\subsubsection{Rare Template and Rare Atom/Token Evaluation Dataset}
\begin{figure*}[!t]
  % \vskip -0.2in
  \begin{center}
    % \centerline{
    \includegraphics[width=0.95\linewidth]{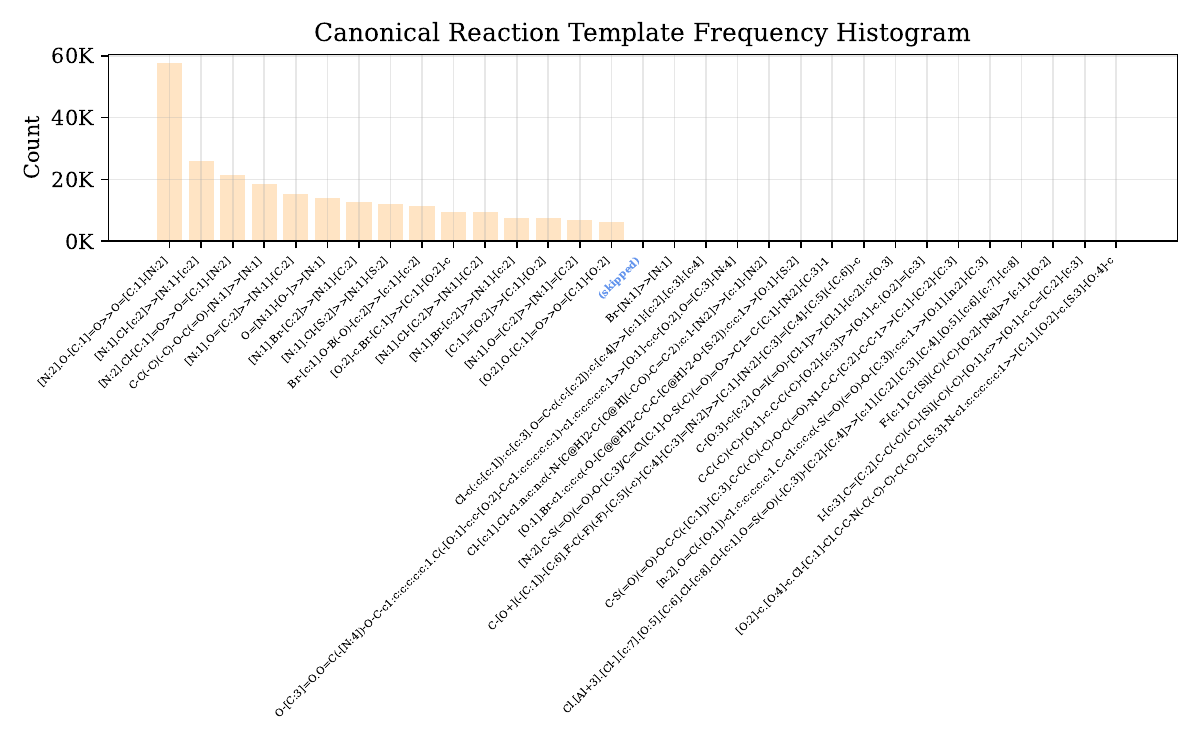}
    \caption{Histogram of the canonical reaction template distribution extracted from all instances in ORDerly. The x-axis lists canonical reaction templates. The top 15 and bottom 15 templates are shown, and templates with intermediate frequencies are skipped.
}
    \label{fig:histogram_template}
  \end{center}
\end{figure*}

\begin{figure*}[!t]
  % \vskip -0.2in
  \begin{center}
    % \centerline{
    \includegraphics[width=0.80\linewidth]{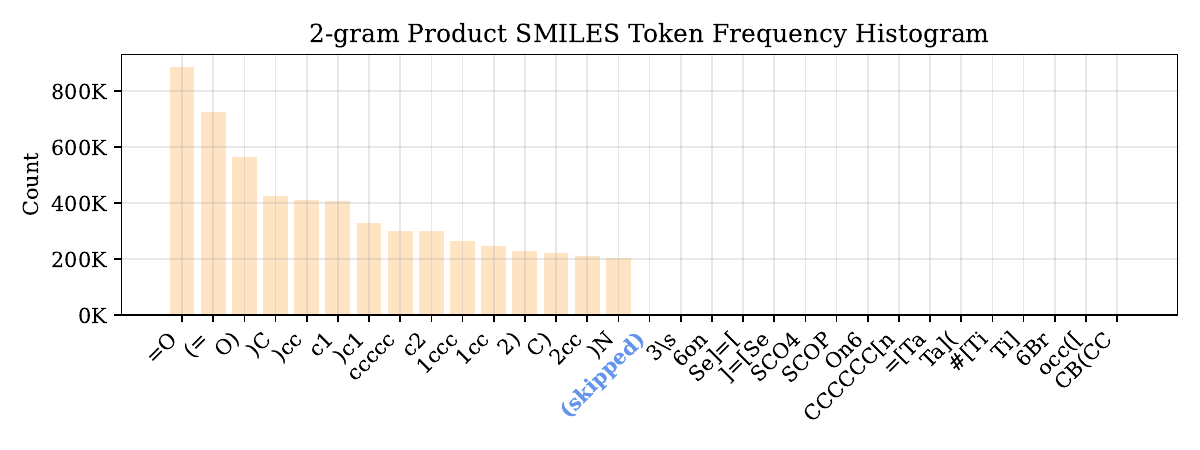}
    \caption{Histogram of the distribution of 2-gram tokens in product SMILES. The top 15 and bottom 15 2-gram tokens are shown, and tokens with intermediate frequencies are skipped.}
    \label{fig:histogram_2gram}
  \end{center}
\end{figure*}

\begin{figure*}[!t]
  % \vskip -0.2in
  \begin{center}
    % \centerline{
    \includegraphics[width=0.80\linewidth]{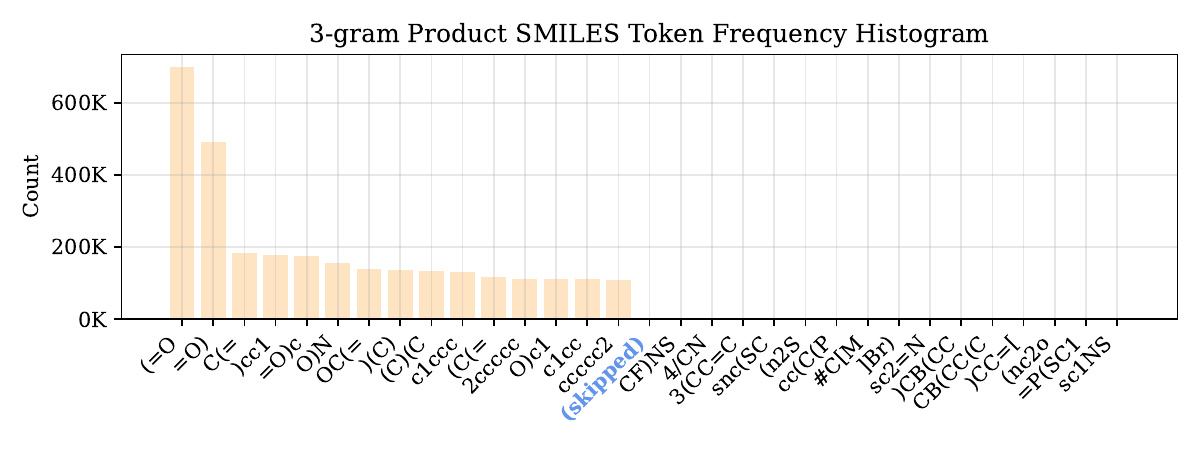}
    \caption{Histogram of the distribution of 3-gram tokens in product SMILES. The top 15 and bottom 15 3-gram tokens are shown, and tokens with intermediate frequencies are skipped.}
    \label{fig:histogram_3gram}
  \end{center}
\end{figure*}

Hard instances are created by constructing a dataset that includes rare reaction templates and rare tokens.
First, canonical reaction templates are extracted from the entire ORDerly train and test sets, and the count of each reaction template is recorded.
Additionally, 2-grams and 3-grams are extracted from the tokenized product SMILES, and the count of each 2-gram and 3-gram is stored.
The distribution of canonical reaction templates is shown in~\Cref{fig:histogram_template}, while the distributions of 2-grams and 3-grams in product SMILES are shown in~\Cref{fig:histogram_2gram,fig:histogram_3gram}, respectively.
Each distribution displays the top 15 and bottom 15 most frequent entries.
The most common reaction template is \texttt{[N:2].O-[C:1]=O>>O=[C:1]-[N:2]}, which represents the condensation reaction between amines and carboxylic acids, a common reaction in organic chemistry.
On the other hand, examining the bottom reaction templates reveals longer and more complex reactions.
When examining the top distributions of 2-grams and 3-grams, it is observed that frequently occurring atoms such as carbon (C) and oxygen (O) in organic molecules are common, while the bottom distributions include elements that are relatively rare in organic molecules, such as selenium (Se) and titanium (Ti).

\paragraph{Rare Reaction Template Instances}
The rare reaction template evaluation dataset is based on this distribution.
From the ORDerly test benchmark, 50 instances are randomly selected whose reaction-template frequency is between 1 and 3, and 50 instances are randomly selected whose reaction-template frequency is between 4 and 6, resulting in a total of 100 instances.

\paragraph{Rare Atom/Token Instances}
Rare atom/token instances are selected by scoring each instance based on the presence of rare tokens in the product SMILES.
The top 50 instances based on 2-grams and the top 50 instances based on 3-grams are combined to form a total of 100 instances.
Specifically, let the entire $n$-gram token set be $V$, and let the count set of each token be $\{c(t)\}_{t \in V}$.
The total token count is $N = \sum_{v \in V} c(v)$, and the occurrence ratio for each token $t$ is $p_a(t) = \frac{c(t)}{N}$.
The score for each token is defined based on the occurrence ratio as $s(t) = -\log p_a(t)$.
If a product SMILES is composed of $\mathbf{x} = (t_1, \dots, t_L)$, the instance score $S(x)$ is calculated as $S(x) = \frac{1}{L} \sum_{i=1}^{L} s(t_i)$, and instances are selected based on this score.

\subsection{Sampling Hyperparameters}
\begin{table*}[!t]
  \begin{center}
    \begin{small}
        % \resizebox{\linewidth}{!}{%
\begin{tabular}{@{}lccccc@{}}
\toprule
                                                                              & Temperature & Top-p & Max new tokens & Reasoning Effort & Enable Thinking \\ \midrule
\emph{Molecular Prediction LLMs} &     &      &      &        &   \\
LlaSMol                          & 1.0 & 1.0  & 256  & -      & - \\
Mol-Instructions                 & 1.0 & 1.0  & 256  & -      & - \\
BioT5+                           & 1.0 & 1.0  & -    & -      & - \\
PRESTO                           & 1.0 & 1.0  & -    & -      & - \\
Mol-LLM                          & 1.0 & 1.0  & -    & -      & - \\ \midrule
\emph{Molecular Reasoning LLMs}  &     &      &      &        &   \\
Chem-R                           & 1.0 & 1.0  & -    & -      & - \\
ChemDFM                          & 1.0 & 1.0  & -    & -      & - \\
ether0                           & 1.0 & 1.0  & -    & -      & - \\ \midrule
\emph{General Purpose LLMs}      &     &      &      &        &   \\
OpenAI-o3                        & -   & 1.0  & 2048 & low    & - \\
GPT-5-mini                       & -   & 1.0  & 2048 & medium & - \\
GPT-oss-120B                     & 1.0 & 0.95 & -    & high   & - \\
Qwen3-8B                         & 1.2 & 0.95 & -    & -      & O \\
Qwen3-235B-A22B                  & 1.2 & 0.95 & -    & -      & O \\ \midrule
\emph{Expert Models}             &     &      &      &        &   \\
RetroSynFlow    & N/A & N/A  & N/A  & N/A    & N/A \\
G2S-HCVAE       & 1.0 & N/A  & N/A  & N/A    & N/A \\ \midrule
Prediction-Only (SFT)            & 1.2 & 1.0  & 500  & -      & - \\
Prediction-Only (RL)             & 1.2 & 1.0  & 500  & -      & - \\
RetroReasoner (SFT)              & 1.2 & 1.0  & 3000 & -      & - \\
RetroReasoner (SFT, w/o $\mathcal{L}_{12},\mathcal{L}_{23},\mathcal{L}_{34}$) & 1.2         & 1.0   & 3000           & -                & -               \\
RetroReasoner (SFT, w/ only $\mathcal{R}_3, \mathcal{R}_4$)                                     & 1.2         & 1.0   & 3000           & -                & -               \\
RetroReasoner (SFT, w/ only $\mathcal{R}_1$)                                     & 1.2         & 1.0   & 3000           & -                & -               \\
RetroReasoner (RL)               & 1.2 & 1.0  & 3000 & -      & - \\
RetroReasoner (RL, w/ $R^{\text{exact}}$)                               & 1.2         & 1.0   & 3000           & -                & -               \\ \bottomrule
\end{tabular}%
        % }
    \end{small}
  \end{center}
  \caption{Summary table of the sampling parameters used to compute the sampling metrics.}
  \label{tab:sampling_hyperparameters}
  \vskip -0.1in
\end{table*}
The hyperparameters used for reactant sampling in the experiment are summarized in~\Cref{tab:sampling_hyperparameters}.

\subsection{Training Details of Expert Models}
\paragraph{RetroSynFlow}
Training RetroSynFlow consists of two stages: reaction center prediction and synthon completion.
For the reaction center prediction model, the publicly available G2G~\cite{shi2020graph} architecture of RetroSynFlow is used without modification and trained with AdamW using a learning rate of $1\times10^{-4}$.
The model is trained for 100 epochs with a total batch size of 128, and the resulting checkpoint is used for inference.
For the synthon completion model, the flow matching model is trained with distributed data parallelism on 8 NVIDIA H100 GPUs.
The base model follows the 5-layer Graph Transformer architecture of RetroSynFlow.
AdamW is used with a constant learning rate schedule, where the learning rate is $8\times10^{-4}$, $\beta_1=0.9$, $\beta_2=0.999$, weight decay is $10^{-12}$, and AMSGrad is enabled.
The model is trained for 500 epochs with a total batch size of 256 across 8 GPUs.

During inference, retrosynthesis is performed through a two-stage pipeline.
First, the trained reaction center identification model predicts the bonds to be disconnected in the product molecular graph and generates the top-2 synthon candidates.
Then, for each synthon candidate, the synthon completion model restores the leaving groups by performing 100-step Euler integration with a linear time schedule.
A total of 100 reactant candidates are generated for each product, with 70 candidates sampled from the top-1 synthon and 30 candidates sampled from the top-2 synthon.
At each integration step, the model takes the current noisy graph, the synthon, and the product context as input and predicts the final reactant.
The vector field computed from this prediction is used to update the probability distribution, and the discrete graph state is transitioned through categorical sampling.
Atoms that are already fixed in the synthon are kept unchanged throughout the entire generation process.

\paragraph{G2S-HCVAE}
Training G2S-HCVAE is performed end-to-end as a single Hierarchical Conditional VAE plugged into a Graph2SMILES~\cite{tu2022permutation} backbone, with a 256-dimensional continuous latent variable $z$ and a discrete latent variable $c$ over $K=10$ reaction classes.
The decoder is trained with teacher forcing to reconstruct reactant SMILES, and the loss is the ELBO combining a token-level cross-entropy reconstruction term (label smoothing $0.1$) and the KL terms for $z$ and $c$, with a 10-epoch KL warmup.
Optimization uses AdamW with the original Transformer inverse-square-root learning rate schedule (peak $\mathrm{lr}=1.0$, 8000 warmup steps) and token-level batching (batch size $4096$, token limit $80{,}000$, gradient accumulation of $2$).
Checkpointing maintains a queue of the last 10 saved models, and weight-averaging over the top 5 is applied automatically every 10 saves.
Training was run for 545 epochs on 8 NVIDIA B200 GPUs under a fixed random seed of $17$.

During inference, only the product is available, so the Recognition Network is discarded and the Prior Network alone seeds the decoder through a latent-seeded beam search.
For each beam, an independent continuous latent is sampled and mapped to a discrete reaction-class token, yielding distinct class tokens that are injected as the initial token of each beam.
Decoding then proceeds as standard autoregressive Transformer beam search, with candidates ranked by accumulated log-probability.
This seeding fragments the beam across different reaction modes before decoding begins, mitigating the mode-collapse of deterministic Transformer decoding.
The beam size is 10 for single-step retrosynthesis, 5 for Retro$^{*}$-based multistep search, and 50 for DirectMultiStep on PaRoutes returning the top-5 routes.

% \subsection{Multi-step Baseline Details}
% XX.
\section{Additional Experimental Results}
\label{appendix:sec:additional_experimental_results}

\subsection{Ablation Studies}

% The ablation study examines the effectiveness of the reasoning strategy that follows the chemist's approach, the impact of linking text, the effect of linking text diversity, and the influence of the round-trip reward.
% Detailed experimental settings, ablation studies on model size, analysis on various sampling parameters, qualitative comparisons, and the accuracy of the structured reasoning steps can be found in~\Cref{appendix:sec:experiment_details}.

\subsubsection{Reasoning Strategy}
\begin{table*}[!t]
  \begin{center}
    \begin{small}
        \resizebox{\linewidth}{!}{%
\begin{tabular}{@{}lllllll@{}}
\toprule
                                                                              & \textsf{Exact@1↑} & \textsf{Round-trip@1↑} & \textsf{Exact@100↑} & \textsf{Round-trip@100↑} & \textsf{Feasible Ratio↑} & \textsf{Template Diversity↑} \\ \midrule
RetroReasoner (SFT)                                                           & \textbf{0.512}    & \textbf{0.812}         & \textbf{0.734}      & 0.944                    & 0.765                    & \textbf{3.898}               \\
RetroReasoner (SFT, w/o $\mathcal{L}_{12},\mathcal{L}_{23},\mathcal{L}_{34}$) & 0.494             & 0.806                  & 0.712               & 0.944                    & 0.779                    & 2.656                        \\
RetroReasoner (SFT, w/ only $\mathcal{R}_3, \mathcal{R}_4$)                   & 0.498             & 0.810                  & 0.706               & 0.926                    & 0.784                    & 2.402                        \\
RetroReasoner (SFT, w/ only $\mathcal{R}_1$)                                  & 0.498             & 0.806                  & 0.708               & \textbf{0.948}           & \textbf{0.777}           & 2.862                        \\ \bottomrule
\end{tabular}%
        }
    \end{small}
  \end{center}
  \caption{Performance comparison of SFT training when linking texts $\mathcal{L}_{12}, \mathcal{L}_{23}, \mathcal{L}_{34}$ are not used and when only the product analysis step $\mathcal{R}_1$ is used as structured reasoning. The best performance is highlighted in \textbf{bold}.
}
  \label{tab:ablation_reasoning_strategy}
  \vskip -0.1in
\end{table*}
\paragraph{Retrosynthesis prediction needs strategic reasoning.}
\Cref{tab:ablation_reasoning_strategy} reports SFT training results for RetroReasoner and an ablation model trained using only $\mathcal{R}_1$, which performs product analysis as its only reasoning step.
Using only $\mathcal{R}_1$ can be regarded as representative of molecular reasoning LLMs.
Comparing the first and fourth rows, \textsf{Round-trip@1↑}, \textsf{Round-trip@100↑}, and \textsf{Feasible Ratio↑} are similar, whereas the full RetroReasoner achieves clearly higher \textsf{Exact@1↑}, \textsf{Exact@100↑}, and \textsf{Template Diversity↑}.
This suggests that complex tasks such as retrosynthesis prediction require strategic reasoning similar to that used by chemists.

\subsubsection{Effect of Linking Text}
\paragraph{Linking text helps improve prediction accuracy and produce more diverse reactant suggestions.}
\Cref{tab:ablation_reasoning_strategy} also compares the SFT training performance of RetroReasoner when only structured reasoning steps $\mathcal{R}_1,\mathcal{R}_2,\mathcal{R}_3,\mathcal{R}_4$ are used, excluding linking texts $\mathcal{L}_{12},\mathcal{L}_{23},\mathcal{L}_{34}$.
The results show similar \textsf{Round-trip@1↑} and \textsf{Round-trip@100↑}, while the full model achieves higher \textsf{Exact@1↑}, \textsf{Exact@100↑}, and \textsf{Template Diversity↑} than the variant without linking texts.
This suggests that linking texts that connect the structured reasoning steps is essential and enables the model to handle a broader feasible reactant space through more diverse reasoning paths.

\subsubsection{Diversity of Linking Text}
\begin{figure}[!t]
  % \vskip 0.2in
  \begin{center}
    % \centerline{
    \includegraphics[width=0.95\columnwidth]{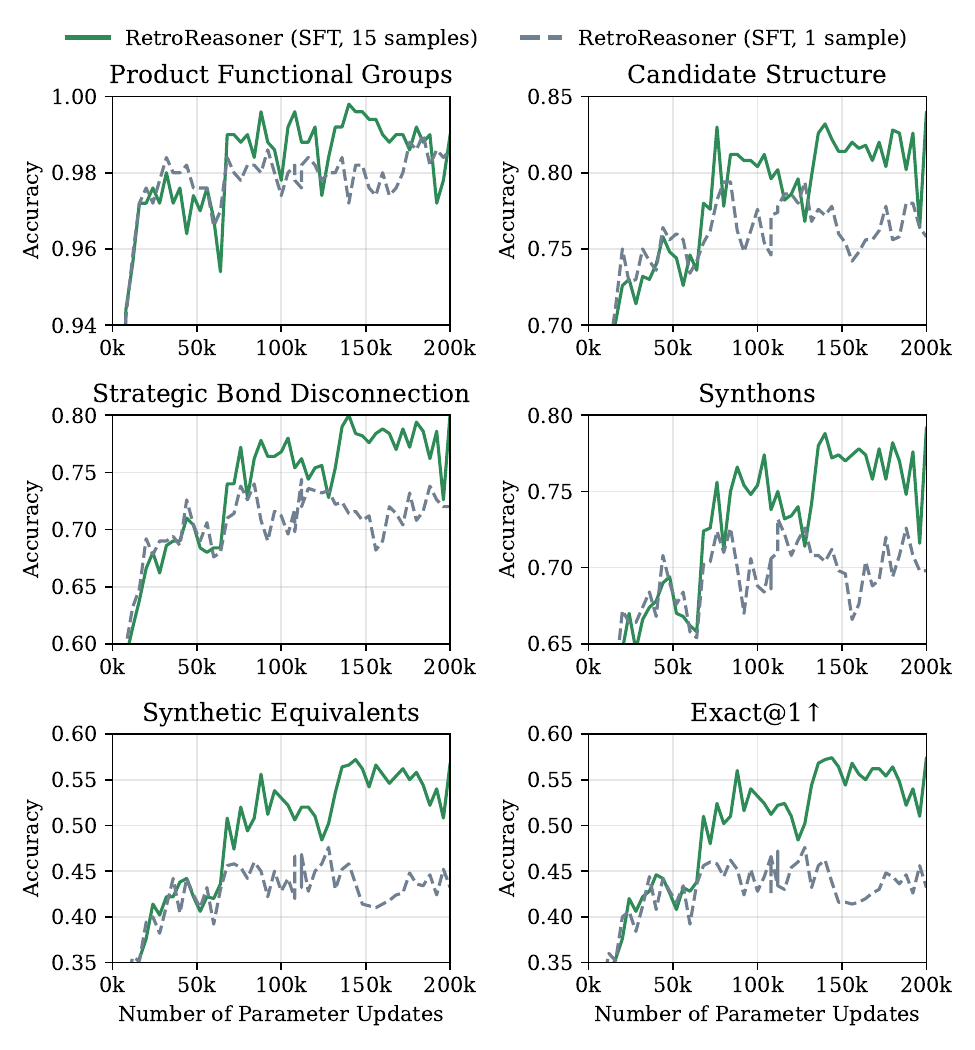}
    \caption{
      Learning curve comparing the effect of the number of linking texts used during SFT. In the default setting, 15 linking texts are generated per instance, with a different linking text used in each epoch. This is compared to the case where only one linking text $(n=1)$ is used consistently across all epochs. The accuracy of each structured reasoning step and the \textsf{Exact@1↑} metric for reactant prediction are shown. The x-axis represents the number of parameter updates.}
    \label{fig:reasoning_diversity_ablation}
  \end{center}
\end{figure}

\paragraph{Training diverse reasoning paths improves train-test generalization.}
\Cref{fig:reasoning_diversity_ablation} presents the learning curves comparing the use of $n=1$ linking text per instance, where the same linking text is used in each epoch during SFT training, with $n = 15$ different linking texts per instance, where a different linking text is used in each epoch.
The changes in \textsf{Exact@1↑} and accuracy for structured reasoning steps are reflected in the corresponding learning curve.
The results clearly show a performance difference in the middle of the learning process.
This indicates that increasing the diversity of reasoning paths, by using different linking texts each epoch, is essential and improves the generalization from train to test.
% The more varied the reasoning paths, the better the model's ability to generalize to new and unseen examples.
% \paragraph{Diversity of Linking Text.}

\subsubsection{Model Size}
\begin{figure*}[!t]
  \vskip 0.2in
  \begin{center}
    \begin{minipage}{0.72\textwidth}
      \centering
      \includegraphics[width=\textwidth]{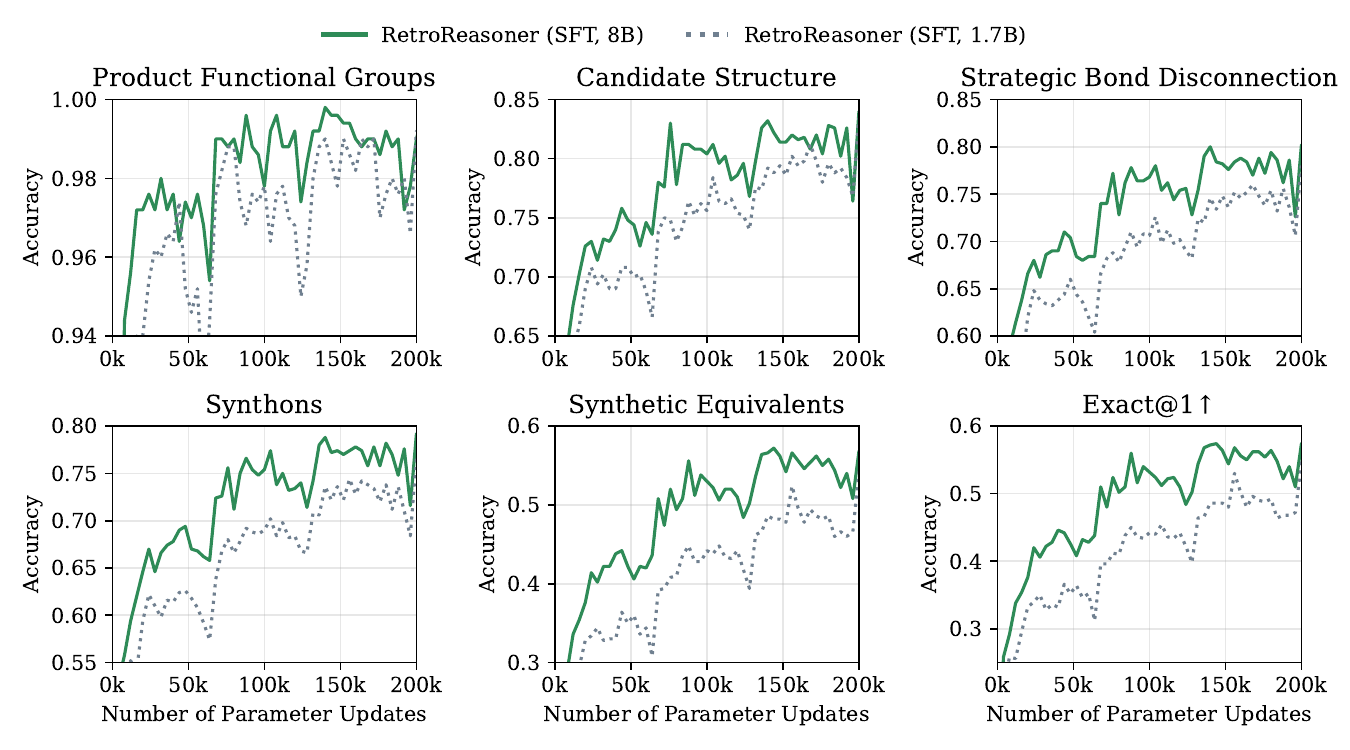}
      \caption{
        Learning curve comparing different model sizes (1.7B, 8B) of the start model, Qwen3. The accuracy of each structured reasoning step and the Exact@1 metric for reactant prediction are shown.}
      \label{fig:model_size_ablation}
    \end{minipage}\hfill
    \begin{minipage}{0.27\textwidth}
      \centering
      \includegraphics[width=\textwidth]{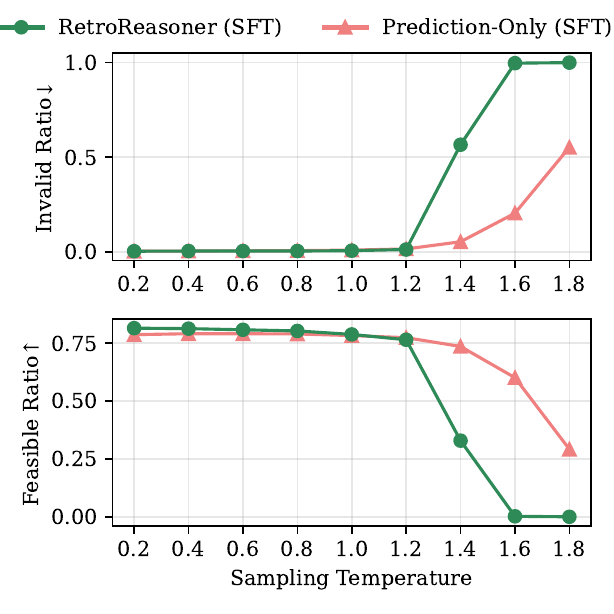}
      \caption{
        Changes in \textsf{Feasible Ratio↑} and \textsf{Invalid Ratio↓} across sampling temperatures. Invalid Ratio is the average, over all instances, of the fraction of sampled reactants that do not satisfy SMILES format.}
      \label{fig:sampling_invalid_ratio}
    \end{minipage}
  \end{center}
\end{figure*}

\paragraph{Larger models are essential for learning the chemist's reasoning strategy.}
\Cref{fig:model_size_ablation} compares the learning curves of two different model sizes: Qwen3-8B, the base model for RetroReasoner, and Qwen3-1.7B.
While both models show improvement during training, a clear performance gap emerges toward the end, indicating that a sufficiently large model size is crucial for effectively learning a reasoning model that mimics the chemist's strategy.

\subsubsection{Effect of Sampling Hyperparameters}
\label{appendix:subsubsec:effect_of_sampling_hyperparameters}
\begin{figure*}[!t]
  \vskip 0.2in
  \begin{center}
    % \centerline{
    \includegraphics[width=0.95\linewidth]{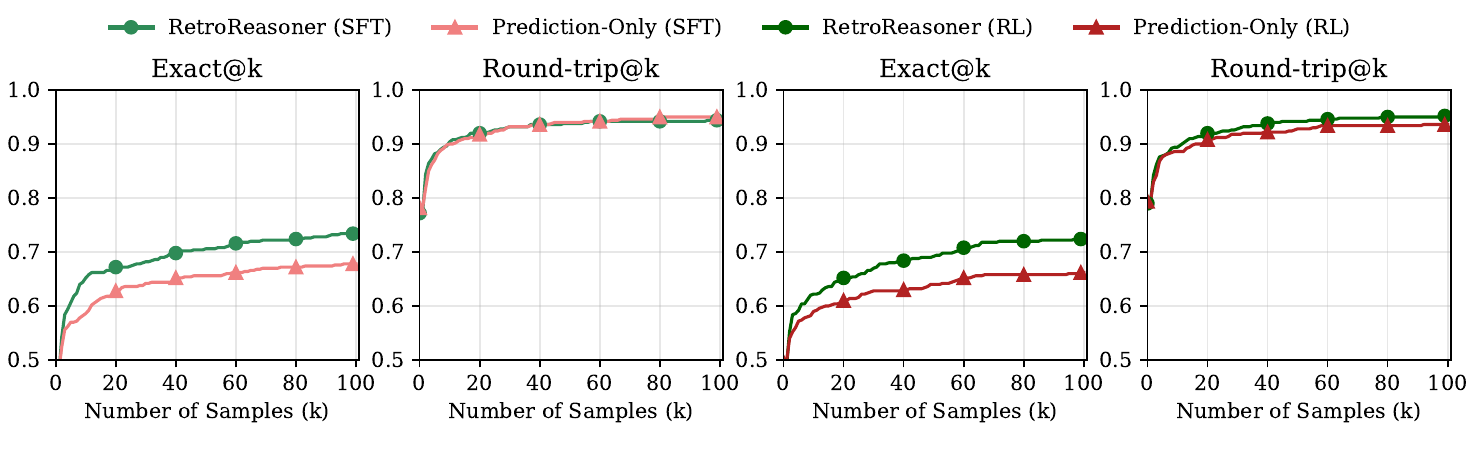}
    \caption{
      Changes in \textsf{Exact@k↑} and \textsf{Round-trip@k↑} performance of the SFT and RL trained models for RetroReasoner and Prediction-Only across different numbers of samples $k$.
    }
    \label{fig:at_k_curve}
  \end{center}
\end{figure*}

\begin{table}[!t]
        \resizebox{\linewidth}{!}{%
\begin{tabular}{@{}lll@{}}
\toprule
\textbf{Reasoning step}                 & \textbf{RetroReasoner (SFT)} & \textbf{RetroReasoner (RL)} \\ \midrule
$\mathcal{R}_1$ - product info          & <0.001                       & <0.001                      \\
$\mathcal{R}_1$ - atom-mapped smiles    & <0.001                       & <0.001                      \\
$\mathcal{R}_1$ - functional groups     & <0.001                       & <0.001                      \\
$\mathcal{L}_\text{12}$                 & 0.822                        & 0.857                       \\
$\mathcal{R}_2$ - candidate structure   & 0.016                        & 0.009                       \\
$\mathcal{L}_\text{23}$                 & 0.619                        & 0.672                       \\
$\mathcal{R}_3$ - bond disconnection    & 0.001                        & <0.001                      \\
$\mathcal{R}_3$ - synthons              & 0.001                        & 0.001                       \\
$\mathcal{L}_\text{34}$                 & 0.714                        & 0.759                       \\
$\mathcal{R}_4$ - synthetic equivalents & 0.010                        & 0.005                       \\ \bottomrule
\end{tabular}%
        }
  \caption{Average token entropy of each reasoning component for RetroReasoner before and after RL. Structured reasoning components exhibit substantially lower entropy than the free-form linking texts.}
  \label{tab:reasoning_token_entropy}
  \vskip -0.1in
\end{table}
Sampling parameters control the diversity and feasibility of reactant proposals.
Increasing the sampling temperature allows lower-probability tokens to be selected more frequently, which can increase output diversity but also destabilize generation.
\Cref{fig:sampling_invalid_ratio} shows that the \textsf{Invalid Ratio↓} increases sharply at high temperatures, accompanied by a decrease in \textsf{Feasible Ratio↑}.
Based on this trade-off, a temperature of 1.2 is used as the default setting for the sampling-based evaluation.

\paragraph{The diversity reduction after RL is associated with increased uncertainty in the linking texts.}
\Cref{tab:reasoning_token_entropy} reports the average token entropy of each reasoning component for RetroReasoner before and after RL.
The structured reasoning components remain highly concentrated, with near-zero entropy for most fields.
In contrast, the free-form linking texts $\mathcal{L}_{12}$, $\mathcal{L}_{23}$, and $\mathcal{L}_{34}$ exhibit substantially higher entropy, which further increases after RL from 0.822 to 0.857, 0.619 to 0.672, and 0.714 to 0.759, respectively.
At higher temperatures, this high-entropy tail increases the probability of sampling low-probability out-of-distribution tokens, including unrelated foreign-language characters and Unicode symbols.
Such tokens can disrupt subsequent reasoning and sharply reduce the feasibility of the generated reactants.

To control this decoding instability, min-$p$ decoding with $p=0.05$ is applied.
At each decoding step, tokens with probabilities below 0.05 times the maximum token probability are filtered before sampling.
Sampling temperature is then varied from 1.2 to 3.0 in increments of 0.1 to explore different feasibility–diversity operating points. As shown in~\Cref{fig:feasibility_diversity_tradeoff}, RetroReasoner (RL) consistently achieves higher \textsf{Template Diversity↑} than RetroReasoner (SFT) at comparable \textsf{Feasible Ratio↑} levels, forming a more favorable empirical Pareto frontier.
This result indicates that the lower \textsf{Template Diversity↑} observed under the default decoding configuration is sensitive to the decoding operating point rather than representing a uniform reduction in diversity after RL.

% \Cref{fig:at_k_curve}

Finally,~\Cref{fig:at_k_curve} reports \textsf{Exact@k↑} and \textsf{Round-trip@k↑} as the number of sampled candidates increases.
Both metrics improve with additional samples and gradually saturate, showing the benefit of sampling multiple reactant candidates for covering alternative retrosynthetic solutions.

\subsection{Effect of Reasoning Components}
\begin{table}[!t]
        \resizebox{\linewidth}{!}{%
\begin{tabular}{@{}lll@{}}
\toprule
                              & \textsf{Exact@1↑} & \textsf{Round-trip@1↑} \\ \midrule
Original - RetroReasoner (RL) & 0.526             & 0.826                  \\
$\mathcal{R}_1$-corrupt       & 0.018             & 0.028                  \\
$\mathcal{R}_2$-corrupt       & 0.390             & 0.660                  \\
$\mathcal{R}_3$-corrupt       & 0.014             & 0.074                  \\
$\mathcal{R}_4$-corrupt       & 0.000             & 0.000                  \\
$\mathcal{L}_{12}$-corrupt    & 0.212             & 0.532                  \\
$\mathcal{L}_{23}$-corrupt    & 0.464             & 0.794                  \\
$\mathcal{L}_{34}$-corrupt    & 0.468             & 0.774                  \\ \bottomrule
\end{tabular}%
        }
  \caption{Comparison of performance changes when each component of RetroReasoner's reasoning is corrupted.}
  \label{tab:reasoning_corrupt}
  \vskip -0.1in
\end{table}
\paragraph{All components that make up RetroReasoner's reasoning play an important role in reactant prediction.}
To analyze whether each component of the structured reasoning process contributes to the final reactant prediction, a step-corruption experiment is conducted.
RetroReasoner follows the reasoning structure $\mathcal{R}_1 \rightarrow \mathcal{L}_{12} \rightarrow \mathcal{R}_2 \rightarrow \mathcal{L}_{23} \rightarrow \mathcal{R}_3 \rightarrow \mathcal{L}_{34} \rightarrow \mathcal{R}_4 \rightarrow \text{reactant prediction}$.
In each corruption setting, one intermediate reasoning component is replaced with the corresponding component generated for another instance.
Starting from the corrupted component, RetroReasoner then generates the remaining reasoning sequence and the final reactant prediction.
\Cref{tab:reasoning_corrupt} reports the results.
Original denotes the standard inference procedure without any perturbation, i.e., RetroReasoner (RL), whereas \{X\}-corrupt denotes the setting in which component \{X\} is randomly replaced by the corresponding component from another RetroReasoner-generated instance.
As shown in~\Cref{tab:reasoning_corrupt}, corrupting any reasoning component consistently leads to a substantial decrease in final prediction performance compared with the original inference procedure.
This result indicates that each component in the structured reasoning chain plays a functional role in deriving the final reactant prediction.
In other words, the generated reasoning is not merely an explanatory text produced after the fact; rather, it serves as an intermediate computational process that guides subsequent reasoning steps and ultimately affects the predicted reactants.

\subsection{Effect of Forward-model Error on RL}
\begin{table*}[!t]
        \resizebox{\linewidth}{!}{%
\begin{tabular}{@{}lcccccc@{}}
\toprule
                                      & \textsf{Exact@1↑} & \textsf{Round-trip@1↑} & \textsf{Exact@100↑} & \textsf{Round-trip@100↑} & \textsf{Feasible Ratio↑} & \textsf{Template Diversity↑} \\ \midrule
RetroReasoner (RL w/ 87\% round-trip) & 0.526             & 0.826                  & 0.724               & 0.952                    & 0.786                    & 3.186                        \\
RetroReasoner (RL w/ 91\% round-trip) & 0.538             & 0.834                  & 0.728               & 0.942                    & 0.793                    & 3.172                        \\ \bottomrule
\end{tabular}%
        }
  \caption{Comparison of RL training performance based on the accuracy of the round-trip model used for RetroReasoner training. The 87\% round-trip model is 0.6B, and the 91\% round-trip model is 8B.}
  \label{tab:roundtrip_acc_effect}
  % \vskip -0.1in
\end{table*}
% tab:roundtrip_acc_effect
\paragraph{Although the error of the forward model affects performance after RL training to some extent, it is not significant.}
To analyze how forward-model accuracy affects RL training, RetroReasoner is trained with forward models of different scales.
The main experiments use a 0.6B forward model during RL training for inference efficiency, while evaluation is performed with a more accurate 8B forward model.
The 0.6B and 8B forward models achieve approximately 87\% and 91\% accuracy on the forward-model test set, respectively.
\Cref{tab:roundtrip_acc_effect} reports the results when the 8B forward model is also used during RL training.
As shown in~\Cref{tab:roundtrip_acc_effect}, replacing the 0.6B forward model with the more accurate 8B model during RL training produces only slight changes in the final performance.
This suggests that forward-model errors introduce some noise into the RL reward signal, but their effect on the learned policy is limited.
Therefore, the 0.6B forward model offers a practical trade-off between reward reliability and training efficiency.

\subsection{Effect of Removing Early Reasoning Steps}
\paragraph{All reasoning components must be used together to enable more feasible and diverse reactant predictions.}
To analyze the contribution of the full structured reasoning process, a variant using only $\mathcal{R}_3$ and $\mathcal{R}_4$ is evaluated.
This setting is the closest among the variants to conventional semi-template retrosynthesis pipelines, as it primarily relies on reaction-center-level reasoning and reactant prediction without the earlier reasoning components or linking texts.
\Cref{tab:ablation_reasoning_strategy} reports the results.
As shown in~\Cref{tab:ablation_reasoning_strategy}, the variant using only $\mathcal{R}_3$ and $\mathcal{R}_4$ performs worse than the full model, particularly in sampling performance and diversity-related metrics.
This indicates that the earlier reasoning components, $\mathcal{R}_1$ and $\mathcal{R}_2$, as well as the linking texts between reasoning steps, provide meaningful guidance for the overall prediction process.
The results suggest that RetroReasoner benefits from the complete structured reasoning trajectory, rather than relying only on the later reaction-centered reasoning steps.

% \subsection{Generality of an LLM}
% XX

\subsection{Effect of Batch Inference}
\begin{table*}[!t]
        \resizebox{\linewidth}{!}{%
\begin{tabular}{@{}lcccccc@{}}
\toprule
\textbf{} & \multicolumn{1}{l}{\textbf{}} & \multicolumn{1}{l}{\textbf{}} & \multicolumn{2}{c}{\textbf{Batch 1}}         & \multicolumn{2}{c}{\textbf{Batch 100}}       \\ \midrule
          & Avg input tokens              & Avg output tokens             & Total wall time (s) & Throughput (samples/s) & Total wall time (s) & Throughput (samples/s) \\ \midrule
RetroReasoner (SFT) & 67.4 & 953.8 & 733 & 0.14 & 17 & 5.91 \\
RetroReasoner (RL)  & 67.4 & 955.0 & 735 & 0.14 & 17 & 5.94 \\ \bottomrule
\end{tabular}%
        }
  \caption{Comparison of inference speed when serving RetroReasoner as vLLM and predicting one instance 100 times consecutively versus processing 100 requests simultaneously.}
  \label{tab:inference_speed}
  % \vskip -0.1in
\end{table*}
\paragraph{Throughput can be significantly increased when performing predictive processing on a batch basis.}
Inference cost is measured using 100 test examples under both sequential serving and batched serving with vLLM.
\Cref{tab:inference_speed} reports the inference cost comparison.
The results show that $\mathcal{R}_1$ and $\mathcal{R}_2$ account for a large portion of the generated output length, indicating that the early reasoning stages contribute substantially to the overall inference cost.
As shown in~\Cref{tab:inference_speed}, LLM-based structured reasoning introduces a nontrivial computational cost compared with more compact prediction-only inference.
However, batched serving with vLLM substantially improves throughput, reducing the practical inference overhead.
These results indicate that although RetroReasoner requires longer generation due to its explicit reasoning process, efficient serving strategies can mitigate the computational cost in practical deployment.

\subsection{Confidence Interval Measurement}
\begin{table*}[!t]
  \begin{center}
    \begin{small}
        \resizebox{\linewidth}{!}{%
\begin{tabular}{@{}lcccccc@{}}
\toprule
                         & \textsf{Exact@1↑} & \textsf{Round-trip@1↑} & \textsf{Exact@100↑} & \textsf{Round-trip@100↑} & \textsf{Feasible Ratio↑} & \textsf{Template Diversity↑} \\ \midrule
\textit{Main Evaluation}            &                      &                      &                      &                        &                        &                            \\
Prediction-Only (SFT)    & 0.475±0.008          & 0.790±0.007          & 0.669±0.007          & 0.940±0.007            & 0.774±0.005            & 2.477±0.044                \\
Prediction-Only (RL)     & 0.480±0.008          & 0.805±0.008          & 0.655±0.008          & 0.935±0.006            & 0.788±0.006            & 2.213±0.033                \\
RetroReasoner (SFT)      & 0.510±0.009          & 0.819±0.009          & \textbf{0.746±0.008} & \textbf{0.957±0.007}   & 0.770±0.006            & \textbf{3.986±0.090}       \\
RetroReasoner (RL)       & \textbf{0.519±0.008} & \textbf{0.834±0.008} & 0.725±0.009          & 0.951±0.006            & \textbf{0.792±0.006}   & 3.153±0.069                \\ \midrule
\textit{Rare Template}   &                      &                      &                      &                        &                        &                            \\
Prediction-Only (SFT)    & 0.494±0.059          & 0.766±0.034          & 0.696±0.056          & 0.926±0.022            & 0.738±0.029            & 2.282±0.189                \\
Prediction-Only (RL)     & 0.494±0.055          & 0.774±0.025          & 0.687±0.059          & 0.927±0.021            & 0.751±0.028            & 2.096±0.176                \\
RetroReasoner (SFT)      & 0.527±0.068          & 0.808±0.031          & \textbf{0.783±0.039} & \textbf{0.962±0.021}   & 0.742±0.028            & \textbf{4.010±0.355}       \\
RetroReasoner (RL)       & \textbf{0.529±0.071} & \textbf{0.813±0.025} & 0.764±0.048          & 0.955±0.021            & \textbf{0.766±0.029}   & 3.375±0.320                \\ \midrule
\textit{Rare Atom/Token} &                      &                      &                      &                        &                        &                            \\
Prediction-Only (SFT)    & 0.147±0.030          & 0.713±0.050          & 0.310±0.030          & 0.912±0.042            & 0.639±0.041            & 3.111±0.298                \\
Prediction-Only (RL)     & 0.147±0.030          & \textbf{0.714±0.049} & 0.279±0.041          & 0.910±0.027            & \textbf{0.652±0.040}   & 2.803±0.276                \\
RetroReasoner (SFT)      & \textbf{0.155±0.021} & 0.682±0.035          & \textbf{0.432±0.030} & \textbf{0.934±0.021}   & 0.621±0.032            & \textbf{5.400±0.577}       \\
RetroReasoner (RL)       & 0.137±0.022          & 0.694±0.043          & 0.403±0.040          & 0.931±0.022            & 0.646±0.029            & 4.470±0.364                \\ \bottomrule
\end{tabular}%
        }
    \end{small}
  \end{center}
  \caption{Performance comparison including confidence intervals for each evaluation dataset of RetroReasoner and Prediction-Only. The mean and standard deviation of each metric are shown.}
  \label{tab:main_confidence_intervals}
  % \vskip -0.1in
\end{table*}
\paragraph{RetroReasoner demonstrates statistically significantly more feasible and diverse predictive performance than Prediction-Only.}
To assess whether the observed performance gains are robust to the choice of evaluation samples, additional experiments are conducted with repeatedly resampled test sets. Specifically, ten main in-distribution test sets are constructed with 500 instances each, along with ten rare-template test sets and ten rare atom/token test sets with 100 instances each.
\Cref{tab:main_confidence_intervals} reports the mean and standard deviation across these resampled test sets.
As shown in~\Cref{tab:main_confidence_intervals}, the performance gains remain consistent across different resampled evaluation sets, particularly in the main in-distribution setting.
The relatively small standard deviations indicate that the improvements are not driven by a specific test-set sample, but are stable across multiple subsets drawn from the eligible test data.
These results further support the reliability of the reported performance improvements.

\subsection{Effects of Using an External Forward Model}
\label{appendix:subsec:effects_of_using_an_external_forward_model}
\begin{table*}[!t]
        \begin{center}
        \resizebox{0.75\linewidth}{!}{%
\begin{tabular}{@{}lcccc@{}}
\toprule
                              & N (data count) & \textsf{In USPTO↑} & \textsf{USPTO Ratio↑} & \textsf{Unique USPTO Template↑} \\ \midrule
\textbf{Our round-trip model} &                &           &              &                        \\
Prediction-Only (RL)          & 690            & 458       & 66.4\%        & 150                    \\
RetroReasoner (RL)            & 805            & \textbf{542} & \textbf{67.3\%} & \textbf{176}     \\ \midrule
\textbf{MolecularTransformer} &                &           &              &                        \\
Prediction-Only (RL)          & 454            & 313       & 68.9\%        & 112                    \\
RetroReasoner (RL)            & 550            & \textbf{387} & \textbf{70.4\%} & \textbf{143}     \\ \bottomrule
\end{tabular}%
        }
        \end{center}
  \caption{Comparison of the degree to which reaction templates extracted from reactant-product pairs match the reaction template set in the actual reaction dataset, the USPTO dataset, when predicting the product using a round-trip model.}
  \label{tab:roundtrip_MT_comparison_USPTO}
  % \vskip -0.1in
\end{table*}
\paragraph{Evaluations using an external round-trip model are similar, and the use of a round-trip model provides a learning signal based on actual reactions.}
To assess whether the observed gains are specific to the internally trained round-trip model, Molecular Transformer (MT)~\cite{schwaller2019molecular} is used as an independent validator that is not trained on ORDerly.
\Cref{tab:roundtrip_MT_comparison} reports \textsf{Round-trip@1↑} and \textsf{Round-trip@100↑} under both validators.
A similar trend is observed when using either the internal round-trip model or MT, indicating that the improvements are not specific to a single validator.
In addition to round-trip validation, the chemical feasibility of the predicted reactions is examined using reaction templates extracted from an external USPTO reaction dataset.
For each prediction, a reaction of the form \{\textit{predicted reactants}\} \(\rightarrow\) \{\textit{target product}\} is constructed, and its reaction template is compared against a template set extracted from USPTO.
This provides an additional criterion for evaluating whether the predicted reactants correspond to known chemically feasible forward reaction patterns.
The analysis is conducted on the full 500-instance main in-distribution test set.
For each target product, 10 predicted reactant candidates are sampled, and duplicate predictions within each instance are removed.
Among these candidates, only the predictions that pass round-trip validation are retained, i.e., cases where the predicted reactants reconstruct the original target product through the round-trip model.
For the resulting reactions,~\Cref{tab:roundtrip_MT_comparison_USPTO} reports three template-based feasibility metrics: the number of predicted reaction templates included in the USPTO template set (\textsf{In USPTO↑}), the ratio of such templates among round-trip-valid predictions (\textsf{USPTO Ratio↑}), and the number of distinct USPTO-covered templates (\textsf{Unique USPTO Template↑}).
The comparison is performed between RetroReasoner (RL) and Prediction-Only (RL), using both the internal round-trip model and MT as validators.
As shown in~\Cref{tab:roundtrip_MT_comparison_USPTO}, RetroReasoner achieves higher \textsf{In USPTO↑} and \textsf{USPTO Ratio↑} than Prediction-Only, indicating that its round-trip-valid predictions more frequently correspond to reaction patterns observed in the external USPTO-derived template set.
RetroReasoner also obtains a higher \textsf{Unique USPTO Template↑}, suggesting that it covers a broader range of chemically feasible reaction templates.
Notably, among the predicted reactants judged as valid by round-trip validation, nearly 70\% match USPTO-covered reaction patterns capable of producing the target product.
The same overall trend is observed when MT is used as the independent round-trip validator.
These results suggest that the RL signal provided by round-trip validation is meaningfully aligned with chemically feasible reaction patterns, rather than merely exploiting specificity of the internally trained forward model.
The agreement between the internal validator, MT, and the USPTO template-based analysis provides additional evidence that RetroReasoner improves both predictive performance and chemical feasibility.

\subsection{Performance on Unfiltered Multi-label Test Set}
\begin{table*}[!t]
  \begin{center}
    \begin{small}
        \resizebox{\linewidth}{!}{%
\begin{tabular}{@{}lcccccc@{}}
\toprule
\textbf{}             & \textsf{Exact@1↑} & \textsf{Round-trip@1↑} & \textsf{Exact@100↑} & \textsf{Round-trip@100↑} & \textsf{Feasible Ratio↑} & \textsf{Template Diversity↑} \\ \midrule
\textit{Filtered}     &                   &                        &                     &                          &                          &                              \\
Prediction-Only (SFT) & 0.482             & 0.784                  & 0.678               & 0.950                    & 0.774                    & 2.562                        \\
Prediction-Only (RL)  & 0.486             & 0.802                  & 0.662               & 0.936                    & 0.785                    & 2.324                        \\
RetroReasoner (SFT)   & 0.512             & 0.812                  & 0.734               & 0.944                    & 0.765                    & 3.898                        \\
RetroReasoner (RL)    & 0.526             & 0.826                  & 0.724               & 0.952                    & 0.786                    & 3.186                        \\ \midrule
\textit{Unfiltered}   &                   &                        &                     &                          &                          &                              \\
Prediction-Only (SFT) & 0.462             & 0.772                  & 0.666               & 0.922                    & 0.761                    & 2.454                        \\
Prediction-Only (RL)  & 0.464             & 0.790                  & 0.652               & 0.918                    & 0.773                    & 2.166                        \\
RetroReasoner (SFT)   & 0.482             & 0.786                  & 0.716               & 0.946                    & 0.744                    & 4.004                        \\
RetroReasoner (RL)    & 0.506             & 0.814                  & 0.700               & 0.942                    & 0.763                    & 3.192                        \\ \bottomrule
\end{tabular}%
        }
    \end{small}
  \end{center}
  \caption{Performance comparison when duplicate reactant data instances are not filtered from the main evaluation dataset.}
  \label{tab:main_hard_unfiltered}
  \vskip -0.1in
\end{table*}
The main evaluation excludes instances with multiple valid reactant sets from the ORDerly test set to reduce potential bias in exact-match metrics such as \textsf{Exact@1↑} and \textsf{Exact@100↑}.
Since this filtering choice can affect the interpretation of exact-match performance, an additional evaluation is conducted on an unfiltered 500-instance test set that includes such cases.
\Cref{tab:main_hard_unfiltered} reports the comparison between the filtered and unfiltered settings.
As shown in~\Cref{tab:main_hard_unfiltered}, the relative performance gap between Prediction-Only and RetroReasoner remains substantial under both evaluation settings.
This indicates that the improvements of RetroReasoner are not an artifact of excluding instances with multiple valid reactant sets.
The main conclusion therefore remains unchanged: RetroReasoner consistently improves retrosynthesis prediction performance over the prediction-only baseline.

\subsection{Failure Analysis}
\begin{table*}[!t]
  \begin{center}
    % \begin{small}
        \resizebox{\linewidth}{!}{%
\begin{tabular}{@{}ll@{}}
\toprule
\textbf{First deviation point}                                             & \textbf{Fraction of \textsf{Exact@1↑} errors} \\ \midrule
Product perception error (PRODUCT\_INFO)                                    & 7.6\%                                \\
Different (but existing) bond chosen for disconnection                     & 49.4\%                               \\
Incorrect intermolecular versus intramolecular grouping (synthon grouping) & 4.2\%                                \\
Different synthetic equivalent choice                                      & 38.8\%                               \\ \bottomrule
\end{tabular}%
        }
    % \end{small}
  \end{center}
  \caption{First deviation point of RetroReasoner (RL) \textsf{Exact@1↑} errors (n=237).}
  \label{tab:failure_analysis}
\end{table*}
\paragraph{Most exact-match errors arise from alternative retrosynthetic decisions rather than internally inconsistent reasoning.}
Complete reasoning trajectories from RetroReasoner (RL) are analyzed on the 500-instance test set under greedy decoding. Internal consistency checks verify the selected disconnection, resulting synthons, synthetic equivalents, and final answer, while comparison with the SyntheticRetro reference identifies the first deviating step.
As shown in~\Cref{tab:failure_analysis}, among 237 Exact@1 errors, 97.9\% pass all internal consistency checks, with most deviations occurring at alternative bond disconnections (49.4\%) or synthetic-equivalent choices (38.8\%).
Moreover, 69.2\% (164/237) are round-trip consistent.
In contrast, only 0.4\% of \textsf{Exact@1↑}-correct predictions fail an internal consistency check, and 97.0\% match the reference reasoning throughout.
These results indicate that many exact-match failures reflect alternative, internally consistent retrosynthetic decisions rather than malformed reasoning, highlighting the limitation of single-reference evaluation in retrosynthesis.

\subsection{Performance on Existing Benchmarks}
\begin{table*}[!t]
  \begin{center}
    \begin{small}
        \resizebox{\linewidth}{!}{%
\begin{tabular}{@{}lllllll@{}}
\toprule
                                    & \textsf{Exact@1↑}    & \textsf{Round-trip@1↑} & \textsf{Exact@100↑}  & \textsf{Round-trip@100↑} & \textsf{Feasible Ratio↑} & \textsf{Template Diversity↑} \\ \midrule
MHNreact~\cite{seidl2021modern}    &                      &                        &                      &                          &                          &                              \\
Prediction-Only (SFT)               & 0.131                & 0.845                  & 0.345                & 0.952                    & 0.810                    & 2.988                        \\
Prediction-Only (RL)                & 0.119                & 0.845                  & 0.333                & 0.964                    & \textbf{0.825}           & 2.667                        \\
RetroReasoner (SFT)                 & 0.143                & 0.845                  & \textbf{0.417}       & 0.976                    & 0.764                    & \textbf{5.310}               \\
RetroReasoner (RL)                  & \textbf{0.179}       & \textbf{0.857}         & 0.405                & \textbf{0.988}           & 0.793                    & 4.405                        \\ \midrule
TempRe~\cite{xuan2025tempre}     &                      &                        &                      &                          &                          &                              \\
Prediction-Only (SFT)               & 0.083                & 0.719                  & 0.146                & 0.905                    & 0.676                    & 2.542                        \\
Prediction-Only (RL)                & 0.087                & 0.731                  & \textbf{0.166}       & 0.909                    & 0.702                    & 2.245                        \\
RetroReasoner (SFT)                 & \textbf{0.103}       & \textbf{0.751}         & 0.154                & \textbf{0.964}           & 0.694                    & \textbf{3.708}               \\
RetroReasoner (RL)                  & \textbf{0.103}       & \textbf{0.751}         & 0.158                & 0.949                    & \textbf{0.726}           & 3.067                        \\ \bottomrule
\end{tabular}%
        }
    \end{small}
  \end{center}
  \caption{Performance comparison of RetroReasoner and Prediction-Only in MHNreact and TempRe, external rare reaction template benchmarks. For both datasets, the remaining datasets are used for evaluation after excluding the training instances used in RetroReasoner and Prediction-Only. (84 MHNreact instances, 253 TempRe instances)}
  \label{tab:main_hard_existing}
  % \vskip -0.1in
\end{table*}
In this study, the robustness of RetroReasoner is evaluated by constructing relatively challenging evaluation datasets from ORDerly, including rare template and rare atom/token subsets.
However, there are also publicly available benchmark datasets designed for such challenging evaluation, with representative examples including MHNreact~\cite{seidl2021modern} and TempRe~\cite{xuan2025tempre}.
These benchmarks consist of evaluation instances based on rare reaction templates.
However, they may partially overlap with ORDerly, which is used to train RetroReasoner.
Therefore, before evaluation, instances with overlapping reactants and products are removed through molecular comparison.
The results are shown in~\Cref{tab:main_hard_existing}.
Overall, RetroReasoner outperforms Prediction-Only on these benchmarks.
For RetroReasoner with RL, \textsf{Template Diversity↑} slightly decreases compared to the SFT model, while \textsf{Feasible Ratio↑} improves.
This trend is consistent with the results on the rare template and rare atom/token datasets constructed in this study, as well as with the main in-distribution results.

% 본 연구에서는 ORDerly 데이터셋을 사용해서 rare template, rare atom/token 데이터셋으로 상대적으로 어려운 평가 데이터셋을 만들어서 RetroReaoner의 강건성 평가를 실시했다.
% 그러나 이러한 어려운 평가 데이터셋을 벤치마크로 공개한 데이터셋이 있으며 대표적으로 MHNreact와 TempRe가 있다.
% 해당 벤치마크 데이터셋은 rare template들로 이루어진 평가 데이터셋이다.
% 그러나 해당 벤치마크는 RetroReasoner가 학습에 사용한 ORDerly와 학습 instance가 어느정도 겹칠 가능성이 있다.
% 따라서 해당 데이터들을 분자 비교를 통해 reactant와 product가 겹치는 instance들을 제외하고 평가를 실시하며 결과가 Table 19에 나타나 있다.
% 해당 결과를 보면 전반적으로 RetroReasoner가 Prediction-Only보다 더 좋은 성능을 보이며 RetroReasoner의 RL의 경우에는 SFT보다 Template Diversity가 조금 줄어들지만 Feasible Ratio가 더 높은것을 확인할 수 있다.
% 이는 본 연구에서 사용한 rare template, rare atom/token 데이터셋과 결과가 비슷하며, main in-distribution 결과와도 비슷함을 보여준다.

\subsection{Qualitative Comparison}
\label{appendix:subsec:qualitative_comparison}
\begin{sidewaysfigure*}[!t]
  \vskip 0.2in
  \begin{center}
    % \centerline{
    \includegraphics[width=\textwidth]{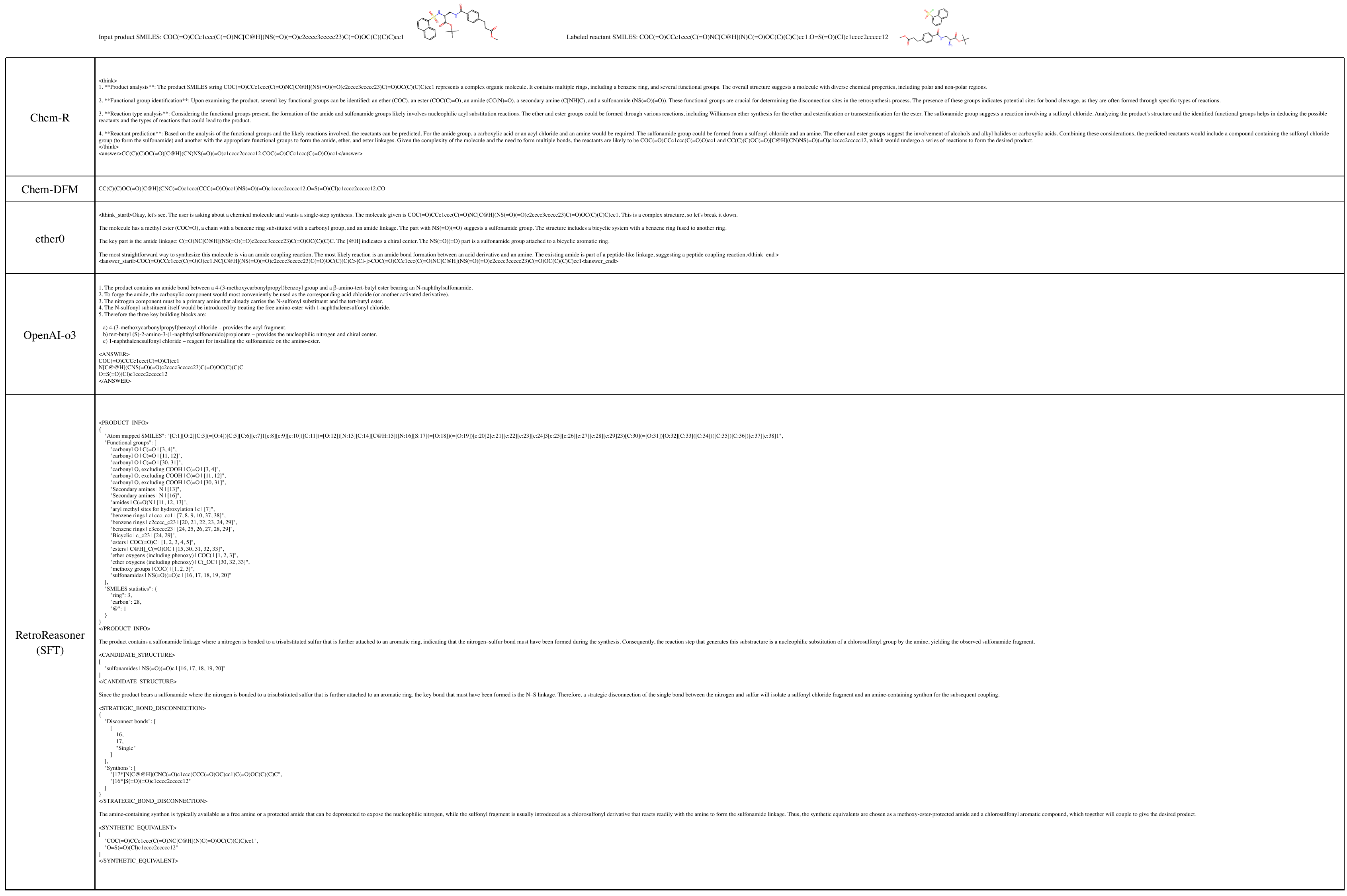}
    \caption{
      Qualitative comparison of the reasoning processes of \emph{Molecular Reasoning LLMs} and OpenAI-o3.
    }
    \label{fig:qualitative_comparison}
  \end{center}
\end{sidewaysfigure*}

\Cref{fig:qualitative_comparison} shows a qualitative comparison of the reasoning processes of Chem-R, ether0, OpenAI-o3, and RetroReasoner (RL).
First, Chem-R follows a reasoning sequence of product analysis → functional group identification → reaction type analysis → reactant prediction.
In the product analysis step, general reasoning is followed by identification of the relevant functional groups.
During this process, the corresponding SMILES of the functional groups are also provided in parentheses, which often differ from the product SMILES.
For example, the ester \texttt{COC(C)=O} in the reasoning is represented as \texttt{COC(=O)} in the actual product SMILES.
RetroReasoner, however, accurately represents functional groups' SMILES along with the position index numbers for each SMILES in \texttt{<PRODUCT\_INFO>}.
Next, after analyzing the reaction type, reactant prediction is made.
In contrast, RetroReasoner follows a chemist's step-by-step reasoning strategy: it finds candidate substructures, performs strategic bond disconnections, and then derives synthetic equivalents from the synthons.
ether0 performs functional group recognition and predicts the sulfonamide group as a key substructure but does not carry out strategic bond disconnections and proceeds directly to reactant prediction.
OpenAI-o3 mainly analyzes the product and describes the process of forming a single bond, but fails to generate accurate SMILES.
On the other hand, RetroReasoner follows a detailed strategy that proceeds from product analysis to identifying key substructures, performing strategic bond disconnections, and mapping synthetic equivalents, thereby gradually predicting the reactants in line with a chemist’s approach.

\subsection{Structured Reasoning Step Accuracy}
\begin{table*}[!t]
  \begin{center}
    \begin{small}
        \resizebox{\linewidth}{!}{%
\begin{tabular}{@{}lccccc@{}}
\toprule
Structured Reasoning Step &
   &
   &
  $\mathcal{R}_1$ &
  $\mathcal{R}_1$ &
  $\mathcal{R}_1$ \\ \midrule
 &
  \textsf{Exact@1↑} &
  \textsf{Round-trip@1↑} &
  Atom Mapping↑ &
  Functional Groups↑ &
  Product SMILES Statistics↑ \\ \midrule
RetroReasoner (SFT) &
  0.512 &
  0.812 &
  \textbf{0.980} &
  0.974 &
  0.992 \\
RetroReasoner (SFT, w/o $\mathcal{L}_{12},\mathcal{L}_{23},\mathcal{L}_{34}$) & 0.494 & 0.806 & 0.978 & \textbf{0.984} & \textbf{0.996} \\
RetroReasoner (SFT, w/ only $\mathcal{R}_1$) &
  0.498 &
  0.806 &
  0.978 &
  0.978 &
  0.988 \\
RetroReasoner (RL) &
  \textbf{0.526} &
  \textbf{0.826} &
  \textbf{0.980} &
  0.972 &
  0.990 \\
RetroReasoner (RL, w/ $R^{\text{exact}}$) &
  0.524 &
  0.824 &
  \textbf{0.980} &
  0.976 &
  0.992 \\ \midrule
Structured Reasoning Step &
  $\mathcal{R}_2$ &
  $\mathcal{R}_3$ &
  $\mathcal{R}_3$ &
  $\mathcal{R}_4$ &
   \\ \midrule
 &
  Candidate Structure↑ &
  Strategic Bond Disconnection↑ &
  Synthon↑ &
  Synthetic Equivalent↑ &
   \\ \midrule
RetroReasoner (SFT) &
  0.788 &
  0.746 &
  0.716 &
  0.512 &
   \\
RetroReasoner (SFT, w/o $\mathcal{L}_{12},\mathcal{L}_{23},\mathcal{L}_{34}$) &
  0.760 &
  0.712 &
  0.676 &
  0.494 &
   \\
RetroReasoner (SFT, w/ only $\mathcal{R}_1$) &
  N/A &
  N/A &
  N/A &
  N/A &
   \\
RetroReasoner (RL) &
  \textbf{0.802} &
  \textbf{0.752} &
  \textbf{0.722} &
  \textbf{0.526} &
   \\
RetroReasoner (RL, w/ $R^{\text{exact}}$) &
  0.788 &
  0.752 &
  0.716 &
  0.524 &
   \\ \bottomrule
\end{tabular}%
        }
    \end{small}
  \end{center}
  \caption{Table comparing the accuracy of structured reasoning steps. \textsf{Exact@1↑} and \textsf{Round-trip@1↑} are also reported for reference, and N/A indicates cases where a step is not generated by design, making the metric unavailable.}
  \label{tab:structured_reasoning_step_accuracy}
  \vskip -0.1in
\end{table*}
RetroReasoner uses structured reasoning steps that contain predefined elements such as functional groups in the product, candidate substructures, and synthons.
\Cref{tab:structured_reasoning_step_accuracy} reports results that measure whether each structured reasoning step produces these elements correctly.
The components in $\mathcal{R}_1$, including atom mapping, functional groups, and SMILES statistics, can be predicted almost perfectly after training.
However, as the steps progress, errors from earlier structured reasoning steps accumulate, and accuracy gradually decreases.
Interestingly, the highest accuracy in these later steps is achieved after RL with round-trip rewards.
This suggests that optimizing round-trip feasibility is partially aligned with improving the correctness of these intermediate reasoning paths.

\subsection{LLM Judge Evaluation}
\label{appendix:subsec:LLM_judge_evaluation}
\begin{figure*}[!t]
  \vskip 0.2in
  \begin{center}
    % \centerline{
    \includegraphics[width=0.55\linewidth]{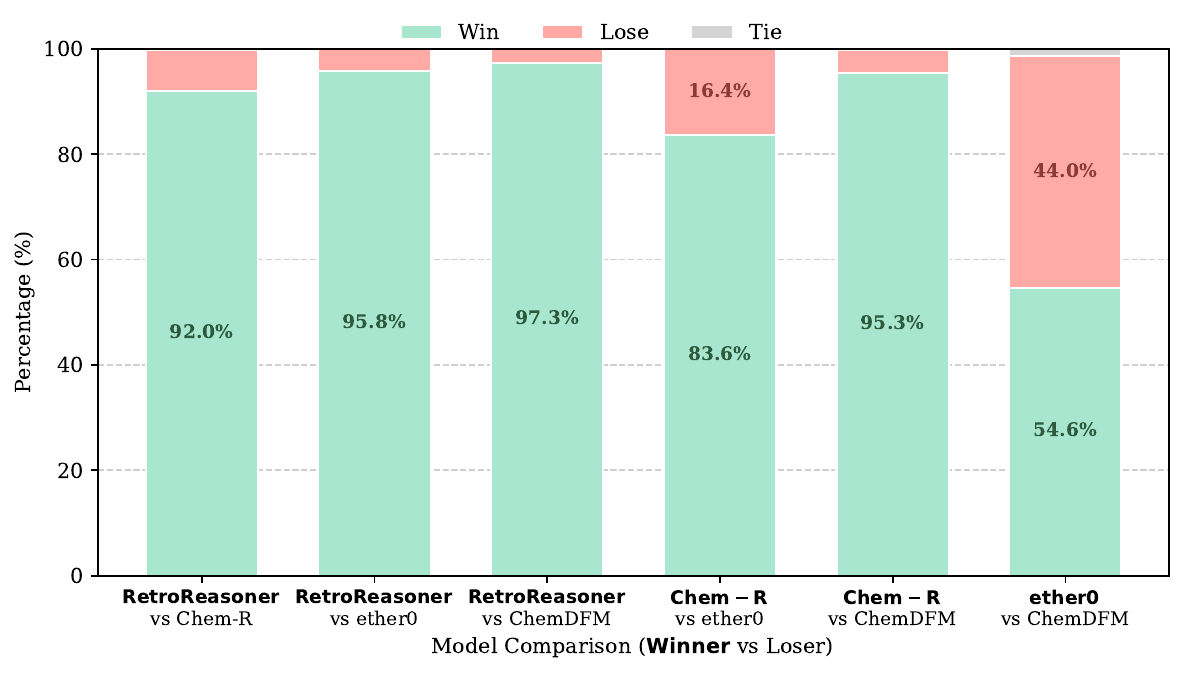}
    \caption{LLM judge evaluation results. Reasoning process generated by \emph{Molecular Reasoning LLMs} on the in-distribution evaluation set are compared in pairs. The judge model is GPT-oss-120B. Each pair is evaluated twice with the order of the two reasonings swapped, and the outcome is accepted only when the two evaluations agree. On the x-axis, the winner model name is shown in \textbf{bold}.
}
    \label{fig:LLM_judge}
  \end{center}
\end{figure*}

A comparison of reasoning strategies among \emph{Molecular Reasoning LLMs} is conducted using an LLM-based evaluation.
The evaluation uses the GPT-oss-120B model as the judge and compares RetroReasoner (SFT), Chem-R, ChemDFM, and ether0.
The criterion is how well the reasoning supports inference of the correct reactants.
The evaluated reasoning texts are those generated during in-distribution evaluation.
For each comparison, two reasoning outputs are provided as A and B, and the judge determines a winner or a tie.
To reduce bias caused by the ordering of A and B, the same comparison is evaluated again with the order swapped.
The result is accepted only when the two judgments are the same.
The judge prompt is shown in~\Cref{fig:prompt_step_evaluation}, and the results are shown in~\Cref{fig:LLM_judge}.
RetroReasoner shows a clear advantage over all other models, indicating stronger reasoning based on chemists' strategies, followed by Chem-R, ether0, and then ChemDFM.

\subsection{Details of Human Expert Evaluation}
\label{appendix:subsec:details_of_human_expert_evaluation}
We provide additional details of the human expert evaluation reported in~\Cref{subsubsec:human_expert_evaluation}.
The evaluation compares RetroReasoner (RL) with Chem-R, the strongest Molecular Reasoning LLM baseline in our main comparison.
As described in~\Cref{subsubsec:human_expert_evaluation}, the evaluation set consists of 50 instances: 33 instances for which both models predict the correct reactants, 9 instances for which only RetroReasoner predicts the correct reactants, and 8 instances for which only Chem-R predicts the correct reactants.

\paragraph{Evaluation Procedure.}
For each instance, evaluators are shown the complete reasoning process and predicted reactants from the two models.
The model identities are hidden, and the two responses are presented in randomized order as Response A and Response B.
Five chemistry experts independently evaluate the responses and select the one that demonstrates better chemical reasoning for retrosynthesis, or indicate a tie when neither response is clearly preferable.

\paragraph{Evaluation Criteria.}

The evaluators are instructed to assess the responses according to the
following criteria:
\begin{itemize}
    \item \textbf{Functional Group Analysis:}
    Does the response correctly identify key functional groups and their reactivity?
    \item \textbf{Reaction Mechanism Understanding:}
    Does the reasoning reflect an accurate understanding of the underlying reaction mechanisms?
    \item \textbf{Overall Logical Flow:}
    Is the retrosynthetic analysis coherent, well-structured, and chemically sound?
\end{itemize}

In addition, evaluators are asked to consider which response would be more helpful for planning an actual synthesis.
The complete instruction presented to the evaluators is as follows:

\begin{quote}
\small
\textit{Which response provides better chemical reasoning for
retrosynthesis?}
\textit{As an expert chemist, please evaluate based on the following
criteria:}

\textit{Functional Group Analysis: Does the response correctly identify key
functional groups and their reactivity?}

\textit{Reaction Mechanism Understanding: Does the reasoning reflect accurate
understanding of reaction mechanisms?}

\textit{Overall Logical Flow: Is the retrosynthetic analysis coherent,
well-structured, and chemically sound?}

\textit{Select the response that would be more helpful for planning an actual
synthesis.}
\end{quote}

As reported in~\Cref{tab:human_evaluation}, RetroReasoner is preferred in 55.2\% of the comparisons, while Chem-R is preferred in 26.8\%, with the remaining 18.0\% judged as ties.
These results show that experts more frequently prefer the overall single-step reasoning and reactant predictions generated by RetroReasoner.

\paragraph{Qualitative Expert Feedback.}

We additionally inspect representative evaluator comments to illustrate the considerations underlying their preferences.
Representative comments from cases in which RetroReasoner is preferred include:

\begin{quote}
\small
``The reasoning of A does not appear relevant to a Suzuki--Miyaura reaction.''
\end{quote}

\begin{quote}
\small
``Both responses assume a reaction involving a tertiary amine, but the product contains two tertiary amines. A explicitly identifies the position bearing the trifluoromethyl group and explains the corresponding reactant.''
\end{quote}

\begin{quote}
\small
``The term `tetraroylated' used by B does not appear to exist.''
\end{quote}

The expert evaluation also reveals cases in which Chem-R is preferred.
Representative comments include:

\begin{quote}
\small
``A contains an invalid SMILES.''
\end{quote}

\begin{quote}
\small
``Without additional constraints on which reactant is preferable, this would be a tie; otherwise, A appears more reasonable.''
\end{quote}

\begin{quote}
\small
``Both are similar, but judging from the final products, B appears more reasonable.''
\end{quote}

These examples illustrate that the evaluators consider not only the high-level logical flow of the reasoning, but also reaction-specific chemical interpretation and the plausibility of the resulting reactants.
They also highlight remaining failure modes of RetroReasoner, such as invalid molecular representations or ambiguous precursor choices.

\section{Prompts}
\label{appendix:sec:prompts}

\subsection{SyntheticRetro Generation Prompt}
\label{appendix:subsec:SyntheticRetro_generation_prompt}
\begin{figure*}[!t]
  \vskip 0.2in
  \begin{center}
    % \centerline{
    \includegraphics[width=0.95\linewidth]{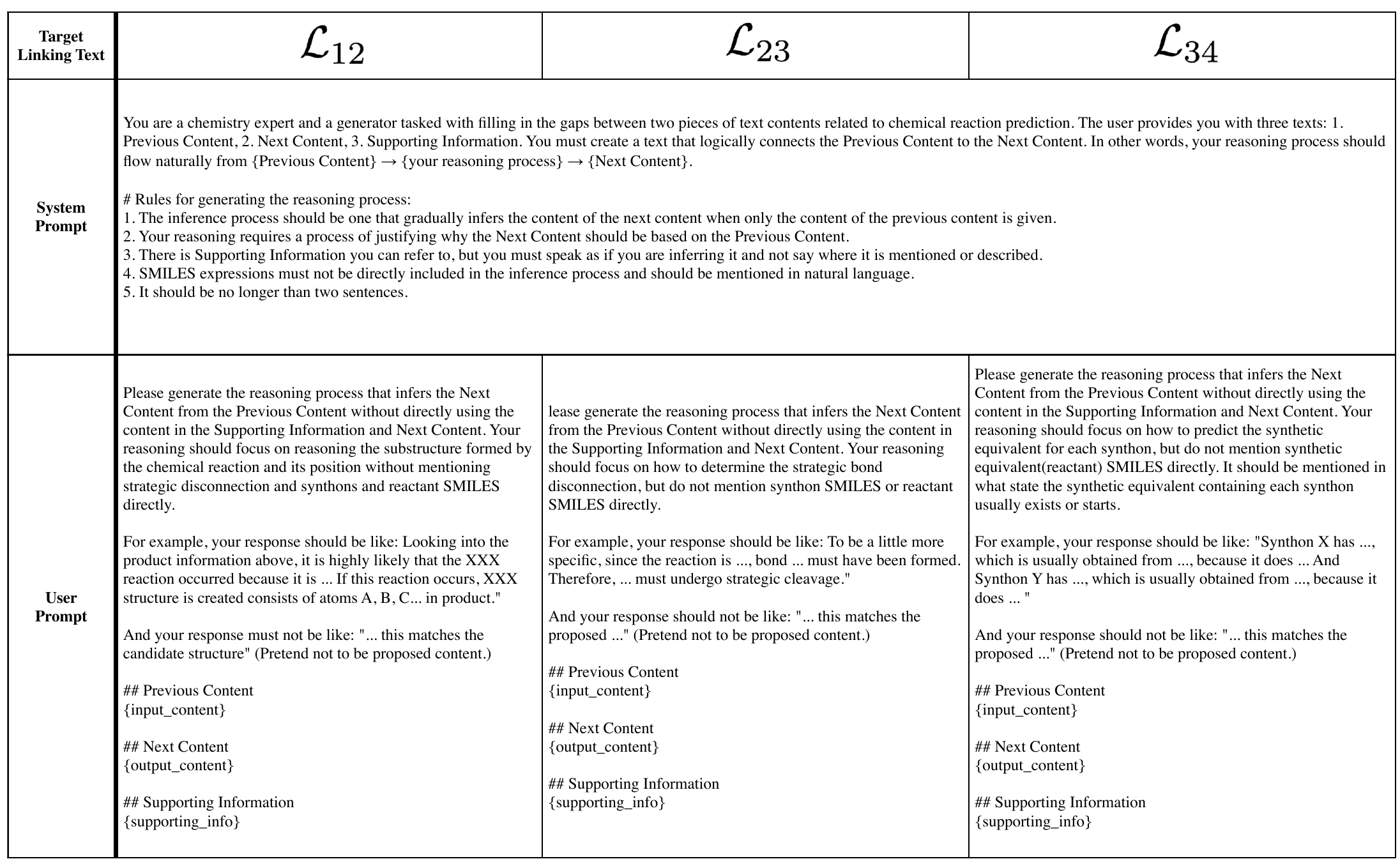}
    \caption{
      Prompt used by SyntheticRetro for linking text generation.
    }
    \label{fig:prompt_SyntheticRetro}
  \end{center}
\end{figure*}

SyntheticRetro generates each linking text $\mathcal{L}_{12}$, $\mathcal{L}_{23}$, and $\mathcal{L}_{34}$ in order, and the prompts for each stage are shown in~\Cref{fig:prompt_SyntheticRetro}.

\subsection{Evaluation}
\begin{figure*}[!t]
  \vskip 0.2in
  \begin{center}
    % \centerline{
    \includegraphics[width=0.95\linewidth]{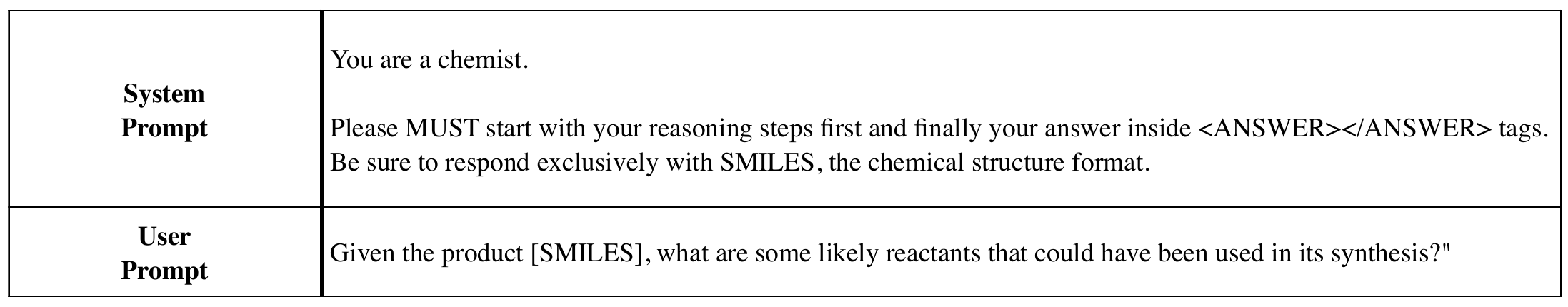}
    \caption{
      Prompt used to evaluate \emph{General Purpose LLMs}. It includes instructions intended to elicit SMILES generation.
    }
    \label{fig:prompt_evaluation}
  \end{center}
\end{figure*}

\emph{Molecular Prediction LLMs} and \emph{Molecular Reasoning LLMs} are evaluated using their original prompt formats when such prompts are provided in prior work.
\emph{General Purpose LLMs} that are not directly trained for the retrosynthesis task are evaluated using the prompt format in~\Cref{fig:prompt_evaluation}, which instructs the model to generate SMILES.

\subsection{LLM Judge Win Rate Evaluation}
\begin{figure*}[!t]
  \vskip 0.2in
  \begin{center}
    % \centerline{
    \includegraphics[width=0.95\linewidth]{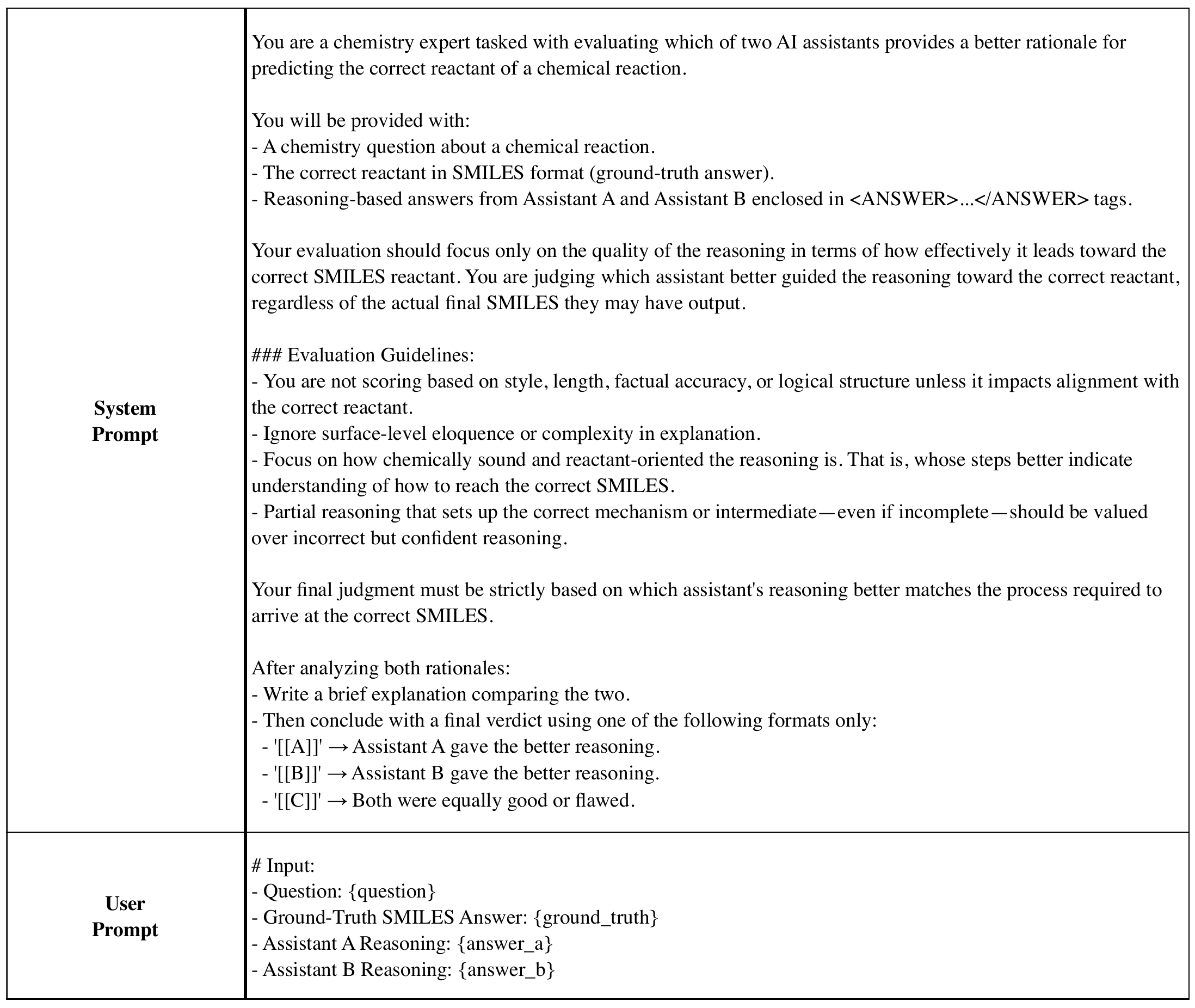}
    \caption{
      Prompt used for the LLM judge, GPT-oss-120B, in the evaluation of reasoning processes produced by \emph{Molecular Reasoning LLMs}.
    }
    \label{fig:prompt_step_evaluation}
  \end{center}
\end{figure*}

Reasoning text evaluation for \emph{Molecular Reasoning LLMs} that use an LLM as a judge is conducted using the prompt shown in~\Cref{fig:prompt_step_evaluation}.

\end{document}